\documentclass[10pt,twocolumn, latterpaper]{article}

\usepackage{iccv}

\usepackage{times}
\usepackage{appendix}
\usepackage{placeins}

\usepackage{authblk}

\usepackage[utf8]{inputenc}
\usepackage[T1]{fontenc}

\usepackage[dvipsnames]{xcolor}

\usepackage{subcaption}

\usepackage{booktabs} %
\usepackage{multirow}
\usepackage{multicol}

\usepackage{graphicx}
\usepackage{amsmath}
\usepackage{amssymb}

\newif\ifdraft

\newcommand{\Cityscapes}{Cordts16}
\newcommand{\LAF}{Pinggera16}
\newcommand{\BDDdset}{Yu18c}

\newcommand{\VGG}{Simonyan15}
\newcommand{\PSPnet}{Zhao17b}

\newcommand{\SegNet}{Badrinarayanan15}
\newcommand{\DiscriminativeFeatureNetwork}{Yu18b}

\newcommand{\PixToPixHD}{Wang18c}

\newcommand{\Dropout}{Srivastava14}
\newcommand{\BayesianSegNet}{Kendall15b}
\newcommand{\AleatoricVsEpistemic}{Kendall17}
\newcommand{\LightweightProb}{Gast18}
\newcommand{\DeepEnsembles}{Lakshminarayanan17}
\newcommand{\GrabCut}{Rother04}
\newcommand{\SELU}{Klambauer17}
\newcommand{\AdamOpt}{Kingma15}

\iccvfinalcopy

\usepackage[pagebackref=true,breaklinks=true,colorlinks,bookmarks=true]{hyperref}

\hypersetup{
	pdftitle={Detecting the Unexpected via Image Resynthesis},
	pdfauthor={Krzysztof Lis, Krishna Nakka, Pascal Fua, Mathieu Salzmann}
}

\begin{document}

\title{Detecting the Unexpected via Image Resynthesis}
\author{Krzysztof Lis}
\author{Krishna Nakka}
\author{Pascal Fua}
\author{Mathieu Salzmann}
\affil{Computer Vision Laboratory, EPFL}

\maketitle

\begin{abstract}
Classical semantic segmentation methods, including the recent deep learning ones, assume that all classes observed at test time have been seen during training. In this paper, we tackle the more realistic scenario where unexpected objects of unknown classes can appear at test time. The main trends in this area either leverage the notion of prediction uncertainty to flag the regions with low confidence as unknown, or rely on autoencoders and highlight poorly-decoded regions. Having observed that, in both cases, the detected regions typically do not correspond to unexpected objects, in this paper, we introduce a drastically different strategy: It relies on the intuition that the network will produce spurious labels in regions depicting unexpected objects. Therefore, resynthesizing the image from the resulting semantic map will yield significant appearance differences with respect to the input image. In other words, we translate the problem of detecting unknown classes to one of identifying poorly-resynthesized image regions. We show that this outperforms both uncertainty- and autoencoder-based methods.
\end{abstract}

\section{Introduction}\label{sec:intro}

\begin{figure}[t]
\centering
\begin{tabular}{cc}
	\includegraphics[width=0.49\linewidth]{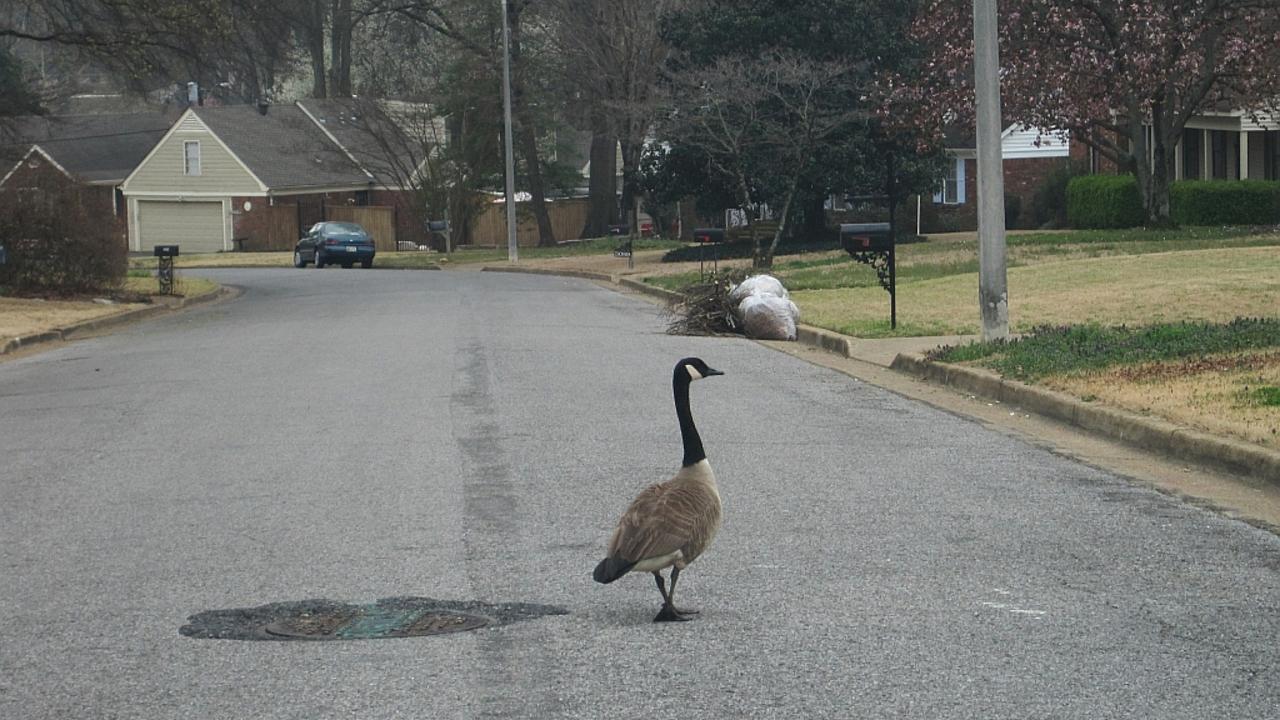}&
	\includegraphics[width=0.49\linewidth]{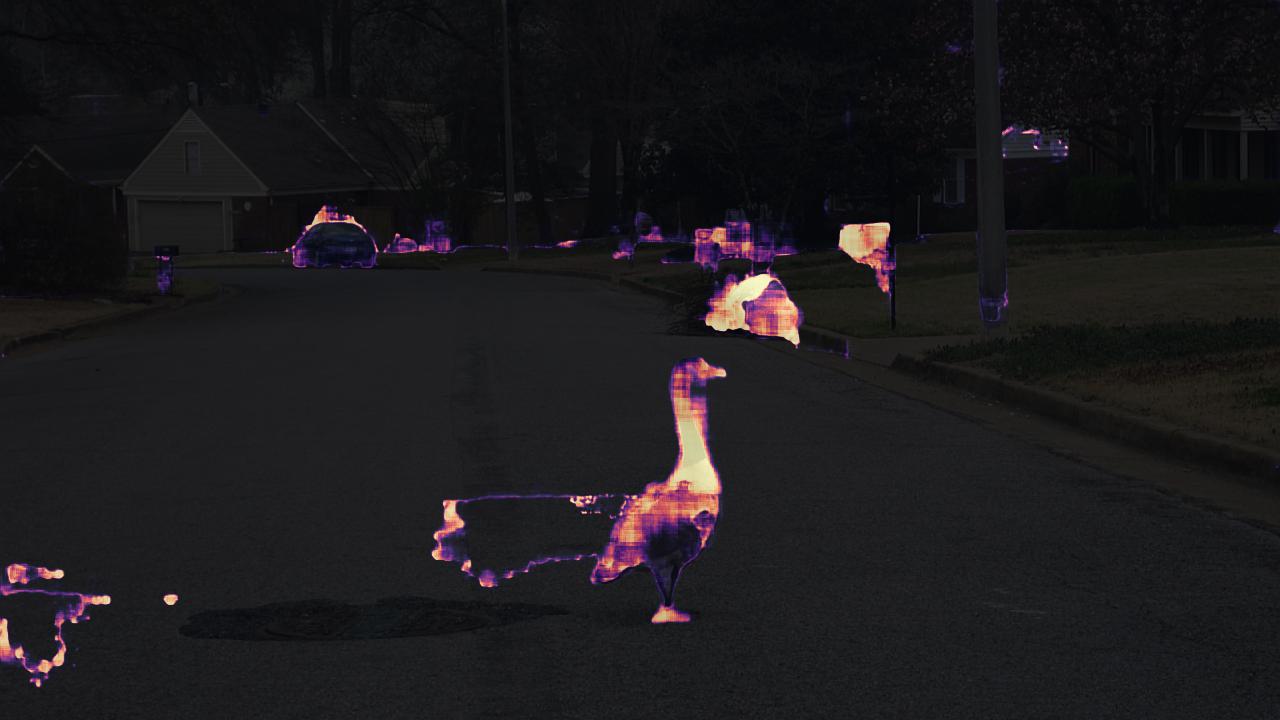}\\[-1mm]
	\small{Input}&
	\small{\bf Ours}\\
	\includegraphics[width=0.49\linewidth]{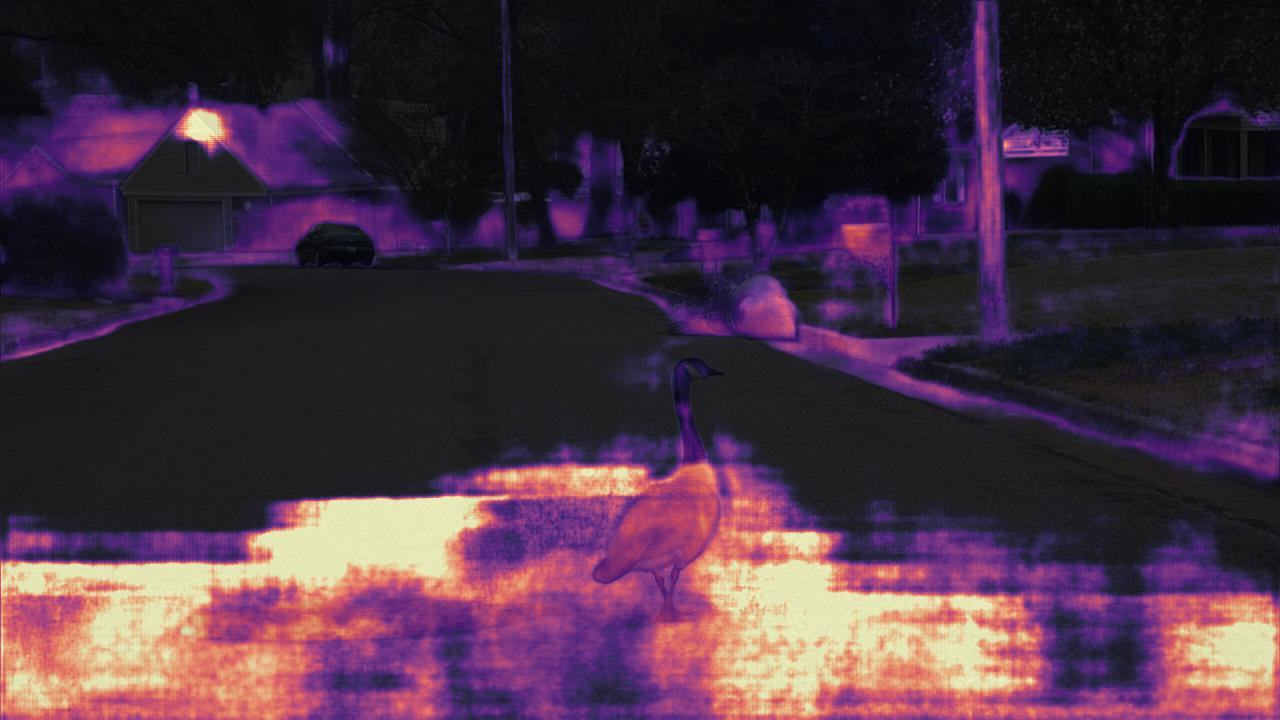}&
	\includegraphics[width=0.49\linewidth]{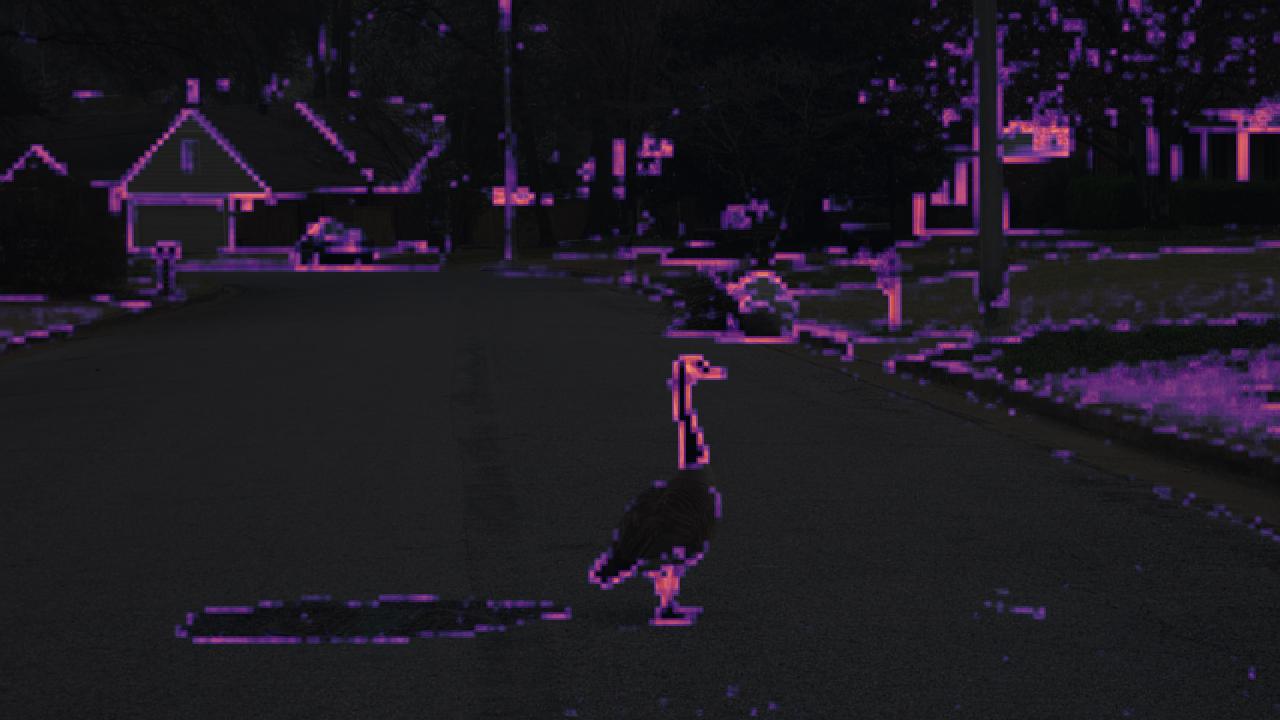}\\[-1mm]
	\small{Uncertainty (Dropout)}&
	\small{RBM autoencoder}\\
\end{tabular}
\vspace{-3mm}
\caption{%
{\bf Detecting the unexpected.}
	While uncertainty- and autoencoder-based methods tend to be distracted by the background, our approach focuses much more accurately on the unknown objects.}
\label{fig:teaser}
\end{figure}

Semantic segmentation has progressed tremendously in recent years and state-of-the-art methods rely on deep learning~\cite{Chen17a,Chen18c,\PSPnet,\DiscriminativeFeatureNetwork}. 
Therefore, they typically operate under the assumption that all classes encountered at test time have been seen at training time. In reality, however, guaranteeing that all classes that can ever be found are represented in the database is impossible when dealing with complex outdoors scenes. For instance, in an autonomous driving scenario, one should expect to occasionally find the unexpected, in the form of animals, snow heaps, or lost cargo on the road, as shown in  Fig.~\ref{fig:teaser}. %
Note that the corresponding labels are absent from standard segmentation training datasets~\cite{\Cityscapes,\BDDdset,Huang18a}. Nevertheless, a self-driving vehicle should at least be able to detect that some image regions cannot be labeled properly and warrant further attention.

Recent approaches to addressing this problem follow two trends. The first one involves reasoning about the prediction uncertainty of the deep networks used to perform the segmentation~\cite{\BayesianSegNet,\DeepEnsembles,\AleatoricVsEpistemic,\LightweightProb}. 
In the driving scenario, we have observed that the uncertain regions tend not to coincide with unknown objects,
and, as illustrated by Fig.~\ref{fig:teaser}, these methods therefore fail to detect the unexpected.
The second trend consists of leveraging autoencoders to detect anomalies~\cite{Creusot15,Munawar17,Akcay18}, assuming that never-seen-before objects will be decoded poorly. We found, however, that autoencoders tend to learn to simply generate a lower-quality version of the input image. As such, as shown in Fig.~\ref{fig:teaser}, they also fail to find the unexpected objects.

\begin{figure}[t]
	\centering
	\includegraphics[width=\linewidth]{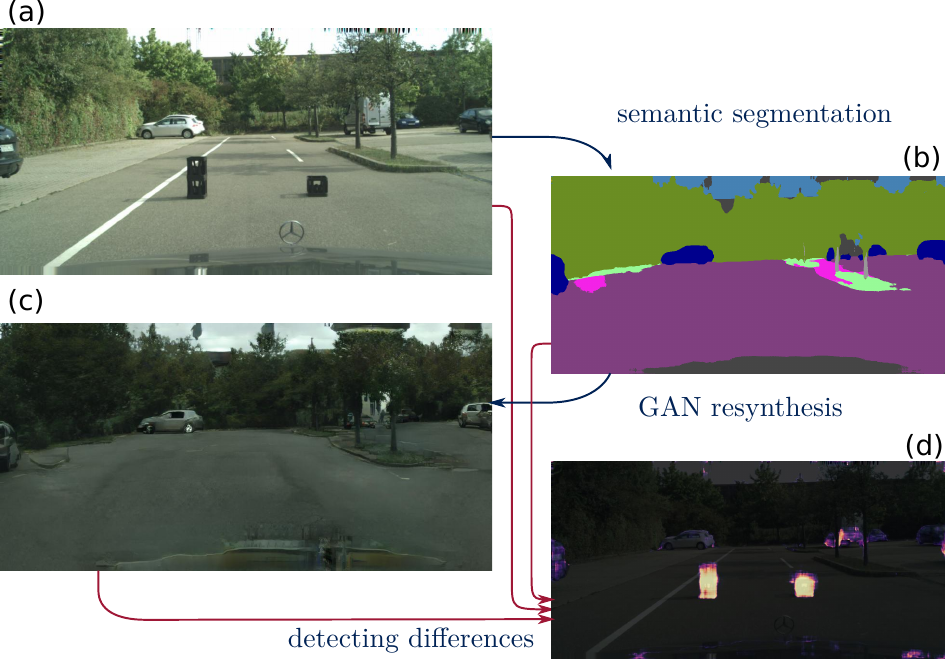}
	\caption{\small{\bf Our Approach.} \textbf{(a)} Input image from the  {\it Lost and Found}~\cite{\LAF} dataset containing objects of a class the segmentation algorithm has not been trained for.  \textbf{(b)} In the resulting semantic segmentation, these objects are lost. \textbf{(c)} In the  image resynthesized based on the segmentation labels, they are also lost.  \textbf{(d)} Using a specially trained {\it discrepancy network} to compare the original image and the resynthesized one highlights the unexpected objects. 
	}
	\label{fig:pipeline}
\end{figure}

In this paper, we therefore introduce a radically different approach to detecting the unexpected. Fig.~\ref{fig:pipeline} depicts our pipeline, built on the following intuition: In regions containing unknown classes, the segmentation network will make spurious predictions. Therefore, if one tries to resynthesize the input image from the semantic label map, the resynthesized unknown regions will look significantly different from the original ones. In other words, we reformulate the problem of segmenting unknown classes as one of identifying the differences between the original input image and the one resynthesized from the predicted semantic map. To this end, we leverage a generative network~\cite{\PixToPixHD}
to learn a mapping from semantic maps back to images. 
We then introduce a discrepancy network that, given as input the original image, the resynthesized one, and the predicted semantic map, produces a binary mask indicating unexpected objects.
To train this network \emph{without} ever observing unexpected objects, we simulate such objects by changing the semantic label of known object instances to other, randomly chosen classes.
This process, described in Section~\ref{sec:synthetic_training}, does {\it not} require seeing the unknown classes during training, which makes our approach applicable to detecting never-seen-before classes at test time.

Our contribution is therefore a radically new approach to identifying regions that have been misclassified by a given semantic segmentation method, based on comparing the original image with a resynthesized one. 
We demonstrate the ability of our approach to detect unexpected objects using the Lost and Found dataset~\cite{\LAF}. This dataset, however, only depicts a limited set of unexpected objects in a fairly constrained scenario. To palliate this lack of data, we create a new dataset depicting unexpected objects, such as animals, rocks, lost tires and construction equipment, on roads. Our method outperforms uncertainty-based baselines, as well as the state-of-the-art autoencoder-based method specifically designed to detect road obstacles~\cite{Creusot15}. 

Furthermore, our approach to detecting anomalies by comparing the original image with a resynthesized one is generic and applies to other tasks than unexpected object detection.   For example, deep learning segmentation algorithms are vulnerable to adversarial attacks~\cite{Xie17,cisse2017}, that is, maliciously crafted images that look normal to a human but cause the segmentation algorithm to fail catastrophically. As in the unexpected object detection case, re-synthesizing the image using the erroneous labels results in a synthetic image that looks nothing like the original one. Then, a simple non-differentiable detector, thus less prone to attacks, is sufficient to identify the attack. As shown by our experiments, our approach outperforms the state-of-the-art one of~\cite{Xiao18} for standard attacks, such as those introduced in~\cite{Xie17,cisse2017}.

\section{Related Work}\label{sec:related}

\subsection{Uncertainty in Semantic Segmentation}

Reasoning about uncertainty in neural networks can be traced back to the early 90s and Bayesian neural networks~\cite{Denker91,Mackay92,MacKay95}. Unfortunately, they are not easy to train and, in practice, {\it dropout}~\cite{\Dropout} has often been used to approximate Bayesian inference~\cite{Gal16}.  An approach relying on explicitly propagating activation uncertainties through the network was recently proposed~\cite{\LightweightProb}.  However, it has only been studied for a restricted set of distributions, such as the Gaussian one. Another alternative to modeling uncertainty is to replace a single network by an ensemble~\cite{\DeepEnsembles}.

For semantic segmentation specifically, the standard approach is to use dropout, as in the {\it Bayesian SegNet}~\cite{\BayesianSegNet}, a framework later extended in~\cite{\AleatoricVsEpistemic}. Leveraging such an approach to estimating label uncertainty then becomes an appealing way to detect unknown objects
because one would expect these objects to coincide with low confidence regions in the predicted semantic map. This approach was pursued in~\cite{Isobe17a, Isobe17b, Isobe17c}. These methods build upon the Bayesian SegNet and incorporate an uncertainty threshold to detect potentially mislabeled regions, including unknown objects. However, as shown in our experiments, uncertainty-based methods, such as the Bayesian SegNet~\cite{\BayesianSegNet} and network ensembles~\cite{\DeepEnsembles}, yield many false positives in irrelevant regions. %
By contrast, our resynthesis-based approach learns to focus on the regions depicting unexpected objects.

\subsection{Anomaly Detection via Resynthesis}

Image resynthesis and generation methods, such as autoencoder and GANs, have been used in the past for anomaly detection. The existing methods, however, mostly focus on finding behavioral anomalies in the temporal domain~\cite{Ravanbakhsh17,Kiran18}. For example,~\cite{Ravanbakhsh17} predicts the optical flow in a video, attempts to reconstruct the images from the flow, and treats significant differences from the original images as evidence for an anomaly. This method, however, was only demonstrated in scenes with a static background. Furthermore, as it relies on flow, it does not apply to single images.

To handle individual images, some algorithms compare the image to the output of a model trained to represent the distribution of the original images. For example, in~\cite{Akcay18}, the image is passed through an adversarial autoencoder and the feature loss between the output and input image is then measured. This can be used to classify whole images but not localize anomalies within the images.  Similarly, given a GAN trained to represent an original distribution, the algorithm of~\cite{Schlegl17} searches for the latent vector that yields the image most similar to the input, which is computationally expensive and does not localize anomalies either.

In the context of road scenes, image resynthesis has been employed to detect traffic obstacles. For example,~\cite{Munawar15} relies on the previous frame to predict the non-anomalous appearance of the road in the current one. In~\cite{Creusot15,Munawar17}, input patches are compared to the output of a shallow autoencoder trained on the road texture, which makes it possible to localize the obstacle. These methods, however, are very specific to roads and lack generality. Furthermore, as shown in our experiments, patch-based approaches such as the one of~\cite{Creusot15} yield many false positives and our approach outperforms it.

Note that the approaches described above typically rely on autoencoder for image resynthesis. We have observed that autoencoders tend to learn to perform image compression, simply synthesizing a lower-quality version of the input image, independently of its content. By contrast, we resynthesize the image from the semantic label map, and thus incorrect class predictions yield appearance variations between the input and resynthesized image.

\subsection{Adversarial Attacks in Semantic Segmentation}
As mentioned before, we can also use the comparison of an original image with a resynthesized one for adversarial attack detection. The main focus of the adversarial attack literature has been on image classification~\cite{goodfellow2014a,carlini2017,moosavi2016}, leading to several defense strategies~\cite{kurakin2016, tramer2017} and detection methods~\cite{metzen2017,lee2018adv,ma18b}. Nevertheless, in~\cite{Xie17,cisse2017}, classification attack schemes were extended to semantic segmentation networks. However, as far as defense schemes are concerned, only~\cite{Xiao18} has proposed an attack detection method in this scenario. This was achieved by analyzing the spatial consistency of the predictions of overlapping image patches. We will show that our approach outperforms this technique.

\section{Approach}
\label{sec:method}

\begin{figure}[t]
	\centering
	\includegraphics[width=\linewidth]{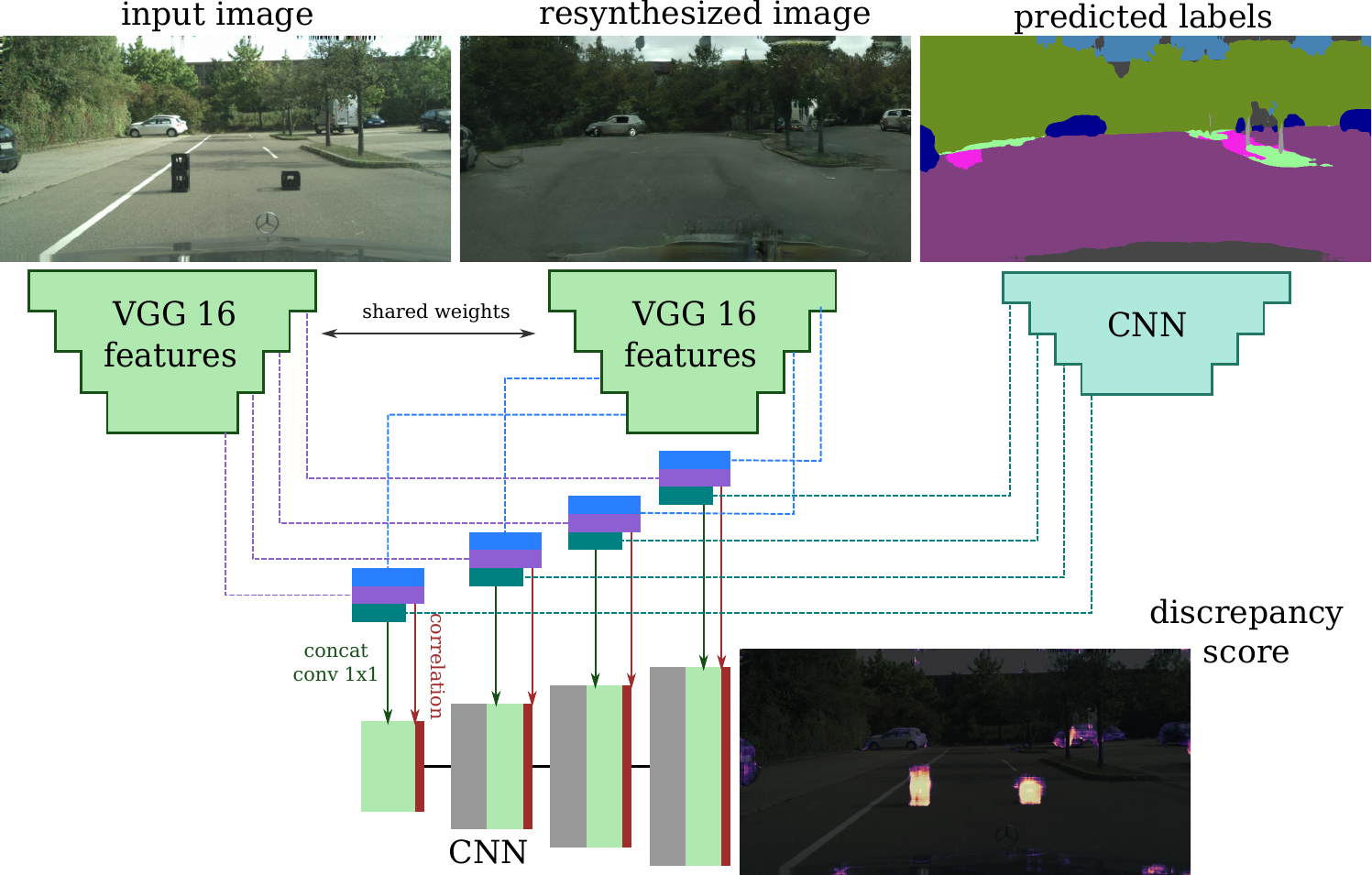}
	\caption{\small{\bf Discrepancy network.} 
		Given the original image, the predicted semantic labels and the resynthesized image as input,
		our discrepancy network detects meaningful differences caused by mislabeled objects.  
		The VGG~\cite{\VGG} network extracts features from both images, which are correlated at all levels of the pyramid.
		Image and label features are then fused using $1 \times 1$ convolutions.
		Both the features and their correlations are then fed to a decoder via skip connections to produce the final discrepancy map. }
	\label{fig:diff_net_arch}
	\vspace{-0.3cm}
\end{figure}

\providecommand{\localwidth}{}
\renewcommand{\localwidth}{0.45\linewidth}

\begin{figure*}[t]
	\centering

	\begin{tabular}{cc}
		\includegraphics[width=\localwidth]{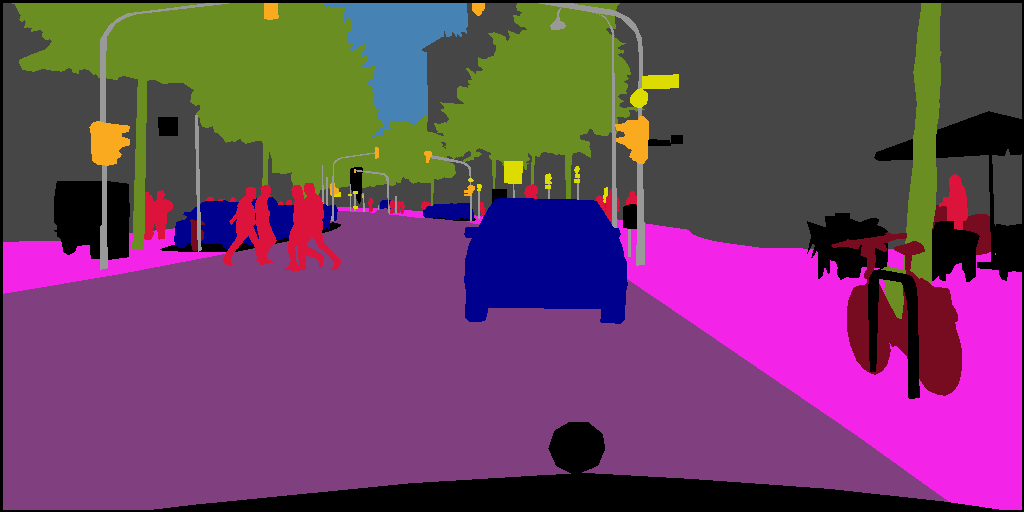}&
		\includegraphics[width=\localwidth]{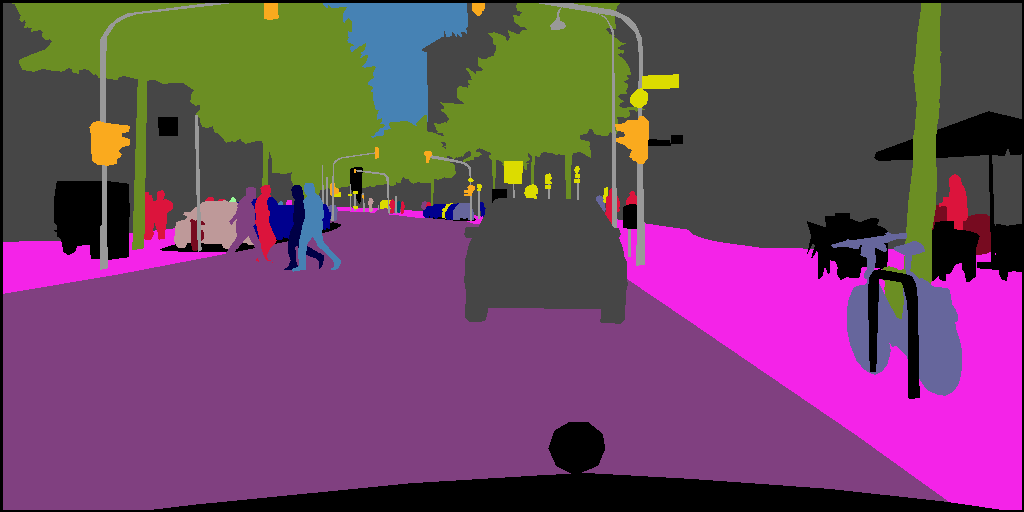}\\	
		(a)&(b)\\
		\includegraphics[width=\localwidth]{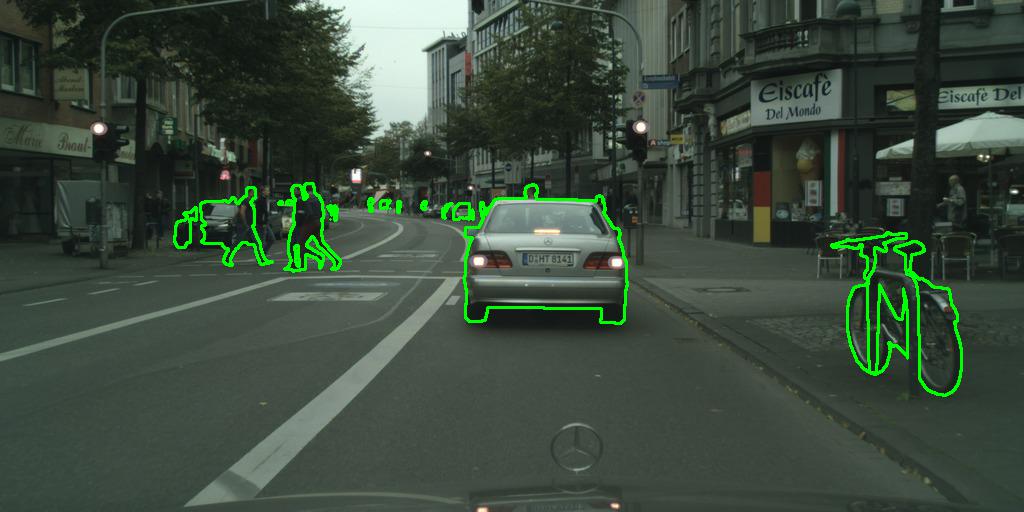}&
		\includegraphics[width=\localwidth]{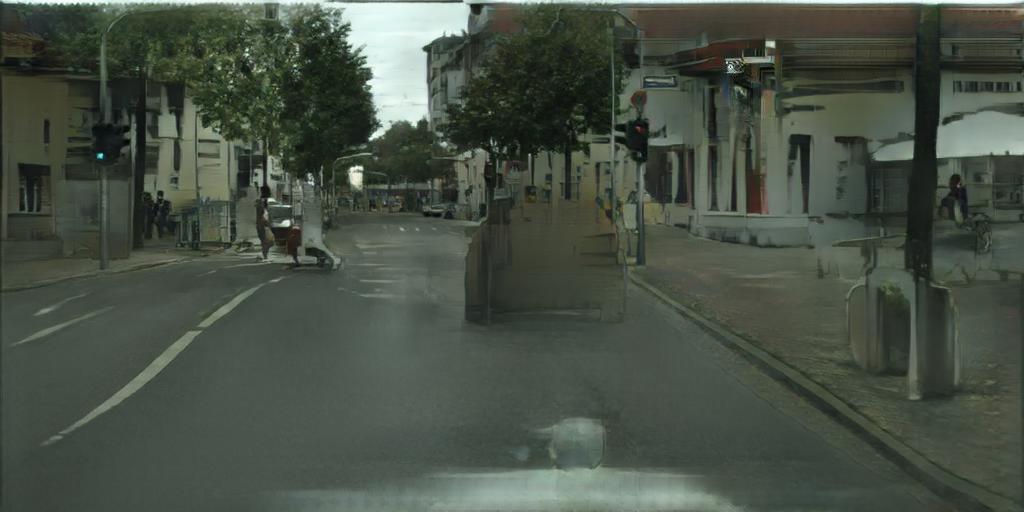}\\
		(c)&(d)\\	
	\end{tabular}

	\vspace{-3mm}
	\caption{{\bf Creating training examples for the discrepancy detector.} 
		{\bf (a)} Ground-truth semantic map. 
		{\bf (b)} We alter the map by replacing some object instances with randomly chosen labels. 
		{\bf (c)} Original image with the overlaid outlines of the altered objects. 
		{\bf (d)} Image re-synthesized using the altered map. 
		We train the discrepancy detector to find the pixels within the outlines of altered objects shown in {\bf (c)}.
	}
	\label{fig:synthetic_training}
\end{figure*}

Our goal is to handle unexpected objects at test time in semantic segmentation and to predict the probability that a pixel belongs to a never-seen-before class. This is in contrast to most of the semantic segmentation literature, which focuses
on assigning to each pixel a probability to belong to classes it has seen in training, without explicit provision for the unexpected. 

Fig.~\ref{fig:pipeline} summarizes our approach. We first use a given semantic segmentation algorithm, such as~\cite{\SegNet} and~\cite{\PSPnet}, to generate a semantic map. 
We then pass this map to a generative network~\cite{\PixToPixHD} that attempts to resynthesize the input image. If the image contains objects belonging to a class that the segmentation algorithm has not been trained for, the corresponding pixels will be mislabeled in the semantic map and therefore poorly resynthesized. We then identify these {\it unexpected} objects by detecting significant differences between the original image and the synthetic one. Below, we introduce our approach to detecting these discrepancies and assessing which differences are significant.

\subsection{Discrepancy Network}\label{sec:discrepancies}

Having synthesized a new image, we compare it to the original one to detect the meaningful differences that denote unexpected objects not captured by the semantic map. While the layout 
of the known objects is preserved in the synthetic image, precise information about the scene's appearance is lost and simply differencing the images would not yield meaningful results. Instead, we train a second network, which we refer to as the {\it discrepancy network}, to detect the image discrepancies that {\it are} significant. 

Fig.~\ref{fig:diff_net_arch} depicts the architecture of our discrepancy network. We drew our inspiration from the co-segmentation network of~\cite{Li18b} that uses feature correlations to detect objects co-occurring in two input images. Our network relies on a 
three-stream architecture that first extracts features from the inputs. We use 
a pre-trained VGG~\cite{\VGG} network for both the original and resynthesized image, and a custom CNN to process the one-hot representation of the predicted labels.
At each level of the feature pyramid, the features of all the streams are concatenated and passed through $1 \times 1$ convolution filters to reduce the number of channels.
In parallel, pointwise correlations between the features of the real image and the resynthesized one are computed and passed, along with the reduced concatenated features, 
to an upconvolution pyramid that returns the final discrepancy score. 
The details of this architecture are provided in the supplementary material.

\subsection{Training}\label{sec:synthetic_training}
When training our discrepancy network, we cannot observe the unknown classes. To address this, we therefore train it on synthetic data that mimics what happens in the presence of unexpected objects. In practice, the semantic segmentation network assigns incorrect class labels to the regions belonging to unknown classes. To simulate this, as illustrated in Fig.~\ref{fig:synthetic_training}, we therefore replace the label of randomly-chosen object instances with a different random one, sampled from the set of known classes. We then resynthesize the input image from this altered semantic map using the {\it pix2pixHD}~\cite{\PixToPixHD} generator trained on the dataset of interest. This creates pairs of real and synthesized images from which we can train our discrepancy network. Note that this strategy does {\it not} require seeing unexpected objects during training.

\subsection{Detecting Adversarial Attacks}
As mentioned above, comparing an input image to a resynthesized one also allows us to detect adversarial attacks.
To this end, we rely on the following strategy. As for unexpected object detection, we first compute a semantic map from the input image, adversarial or not, and resynthesize the scene from this map using the {\it pix2pixHD} generator. Here, unlike in the unexpected object case, the semantic map predicted for an adversarial example is completely wrong and the resynthesized image therefore completely distorted. This makes attack detection a simpler problem than unexpected object one. We can thus use a simple non-differentiable heuristic to compare the input image with the resynthesized one. Specifically, we use the $L^2$ distance between HOG~\cite{Dalal05} features computed on the input and resynthesized image. We then train a logisitic regressor on these distances to predict whether the input image is adversarial or not. Note that this simple heuristic is much harder to attack than a more sophisticated, deep learning based one.

\section{Experiments}\label{sec:experiments}

We first evaluate our approach on the task of detecting unexpected objects,
such as lost cargo, animals, and rocks, in traffic scenes, which constitute our target application domain and the central evaluation domain for semantic segmentation thanks to the availability of large datasets, such as Cityscapes~\cite{\Cityscapes} and BDD100K~\cite{\BDDdset}. For this application,
all tested methods output a per-pixel {\it anomaly score}, 
and we compare the resulting maps with the ground-truth anomaly annotations
using ROC curves and the area under the ROC curve (AUROC) metric. 
Then, we present our results on the task of adversarial attack detection.

We perform evaluations using the Bayesian SegNet~\cite{\BayesianSegNet} and the PSP Net~\cite{\PSPnet},
both trained using the BDD100K dataset~\cite{\BDDdset} (segmentation part)
chosen for its large number of diverse frames, allowing the networks to generalize to the anomaly datasets, whose images differ slightly and cannot be used during training.
To train the image synthesizer and discrepancy detector, we used the training set of Cityscapes~\cite{\Cityscapes}, downscaled to a resolution of $1024 \times 512$ because of GPU memory constraints.

\subsection{Baselines}

As a first baseline, we rely on an uncertainty-based semantic segmentation network. Specifically, we use 
the Bayesian SegNet~\cite{\BayesianSegNet}, which
samples the distribution of the network's results using random dropouts
--- the uncertainty measure is computed as the variance of the samples. We will refer to this method as {\it Uncertainty (Dropout)}.

It requires the semantic segmentation network to contain dropout layers,
which is not the case of most state-of-the-art networks, such as PSP~\cite{\PSPnet}, which is based on a ResNet backbone. 
To calculate the uncertainty of the PSP network, we therefore use the ensemble-based method of~\cite{\DeepEnsembles}:
We trained the PSP model four times, yielding different weights due to the random initialization.
We then use the variance of the outputs of these networks as a proxy for uncertainty.
We will refer to this method as {\it Uncertainty (Ensemble)}.

Finally, we also evaluate the road-specific approach of~\cite{Creusot15}, which relies on training
a shallow Restricted Boltzmann Machine autoencoder to resynthesize patches of road texture corrupted by Gaussian noise.
The regions whose appearance differs from the road are expected not to be reconstructed properly, and thus an
anomaly score for each patch can be obtained using the difference between the autoencoder's input and output.
The original implementation not being publicly available, we re-implemented it and will make our code publicly available for future comparisons.
As in the original article, we use $8\times 8$ patches with stride 6 and a hidden layer of size 20.
We extract the empty road patches required by this method for training from the Cityscapes images using the ground-truth labels to determine the road area.
We will refer to this approach as {\it RBM}. 

The full version of our discrepancy detector takes as input the original image, the resynthesized one and the predicted semantic labels. To study the importance of using both of these information sources as input, we also report the results of variants of our approach that have access to only one of them. We will refer to these variants as
{\it Ours (Resynthesis only)} and {\it Ours (Labels only)}.

\subsection{Anomaly Detection Results}
We evaluate our method's ability to detect unexpected objects using two separate datasets described below.
We did not use any portion of these datasets during training, because we tackle the task of finding never-seen-before objects.

\subsubsection{Lost and Found}

\begin{figure*}[t]
	\centering
	\begin{tabular}{ccc}
		\small{\bf Lost and Found} & \small{\bf Lost and Found} & \small{\bf Road Anomaly} \\
		\small{\bf ROI: all except ego-vehicle} & \small{\bf ROI: road only} & \\
		\includegraphics[width=0.32\linewidth]{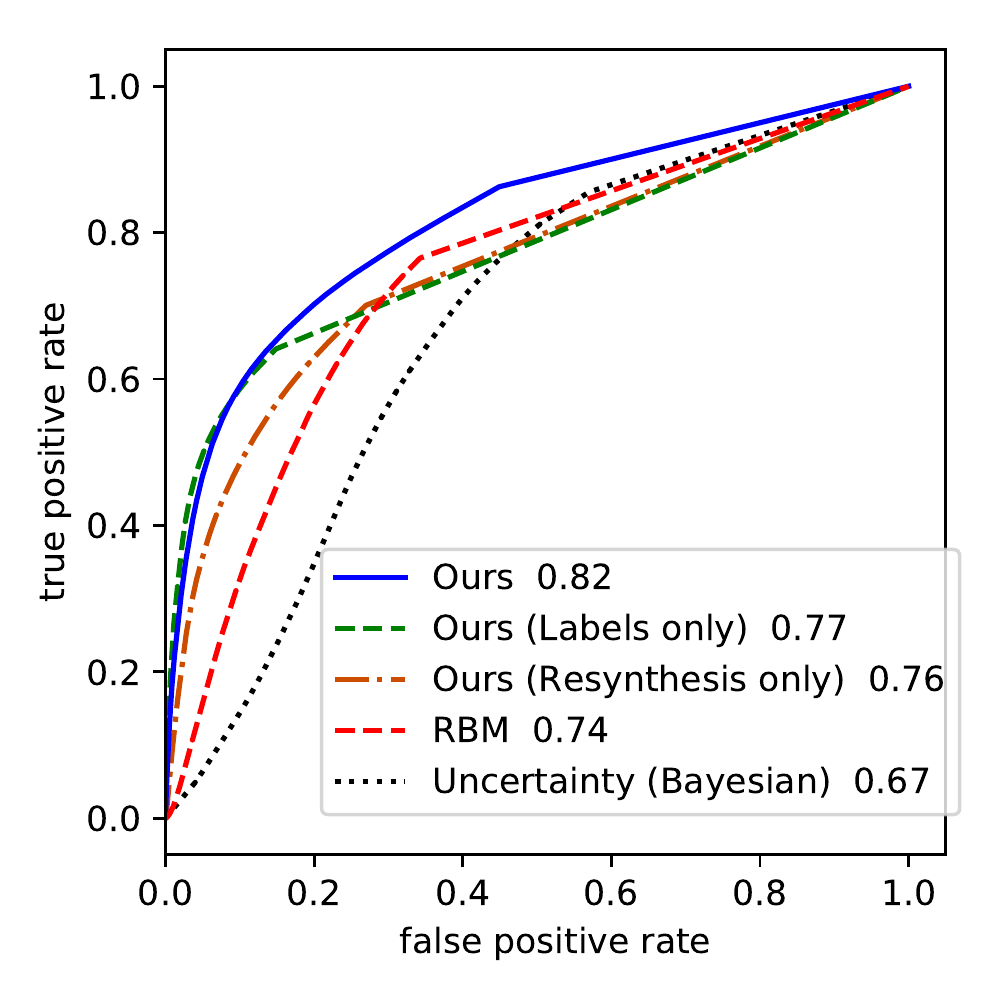} &
		\includegraphics[width=0.32\linewidth]{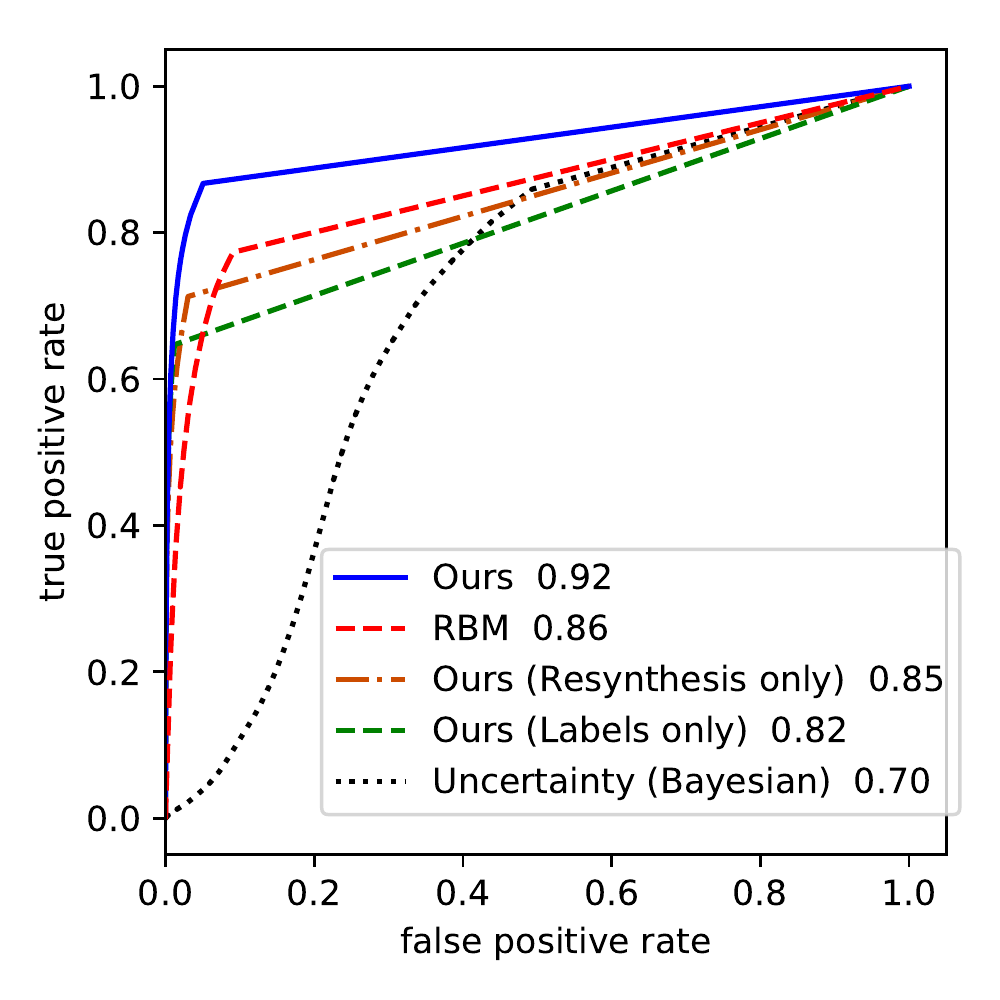} &
		\includegraphics[width=0.32\linewidth]{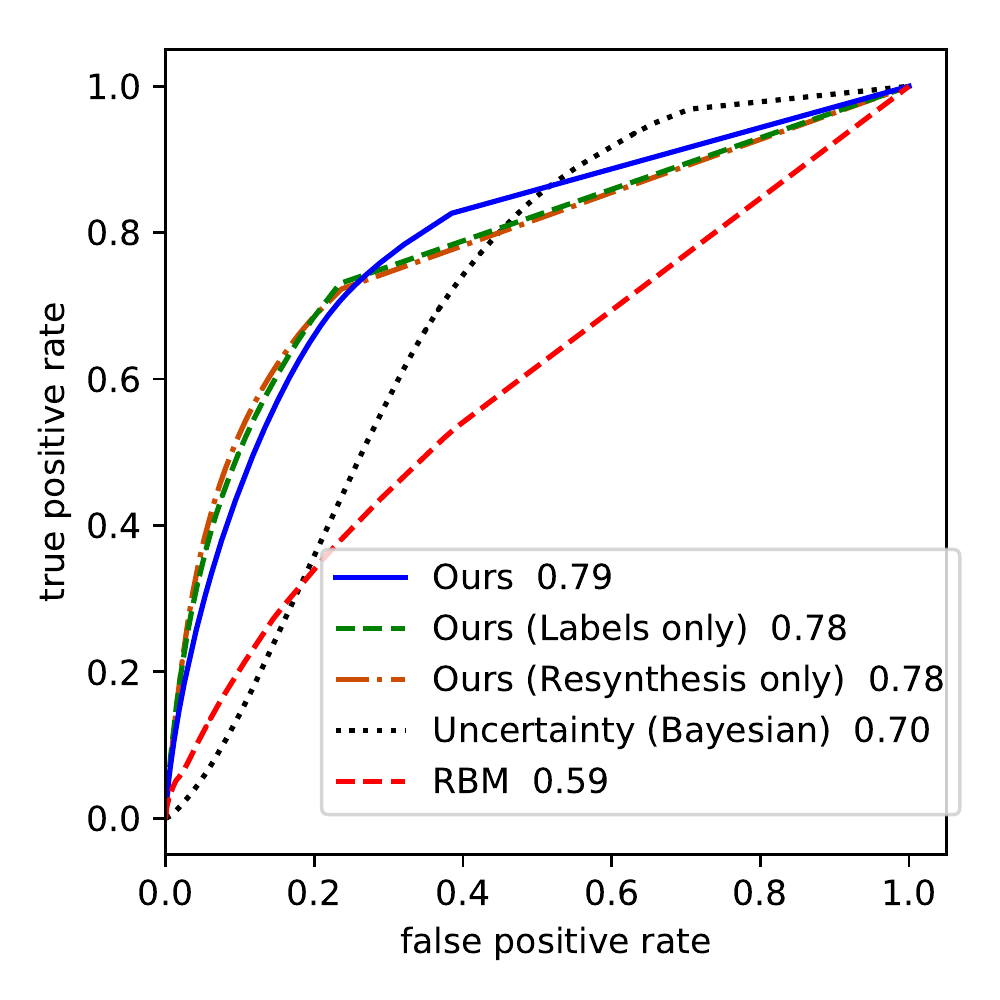} \\
		\small{Bayesian SegNet} & \small{Bayesian SegNet} & \small{Bayesian SegNet} \\
		\includegraphics[width=0.32\linewidth]{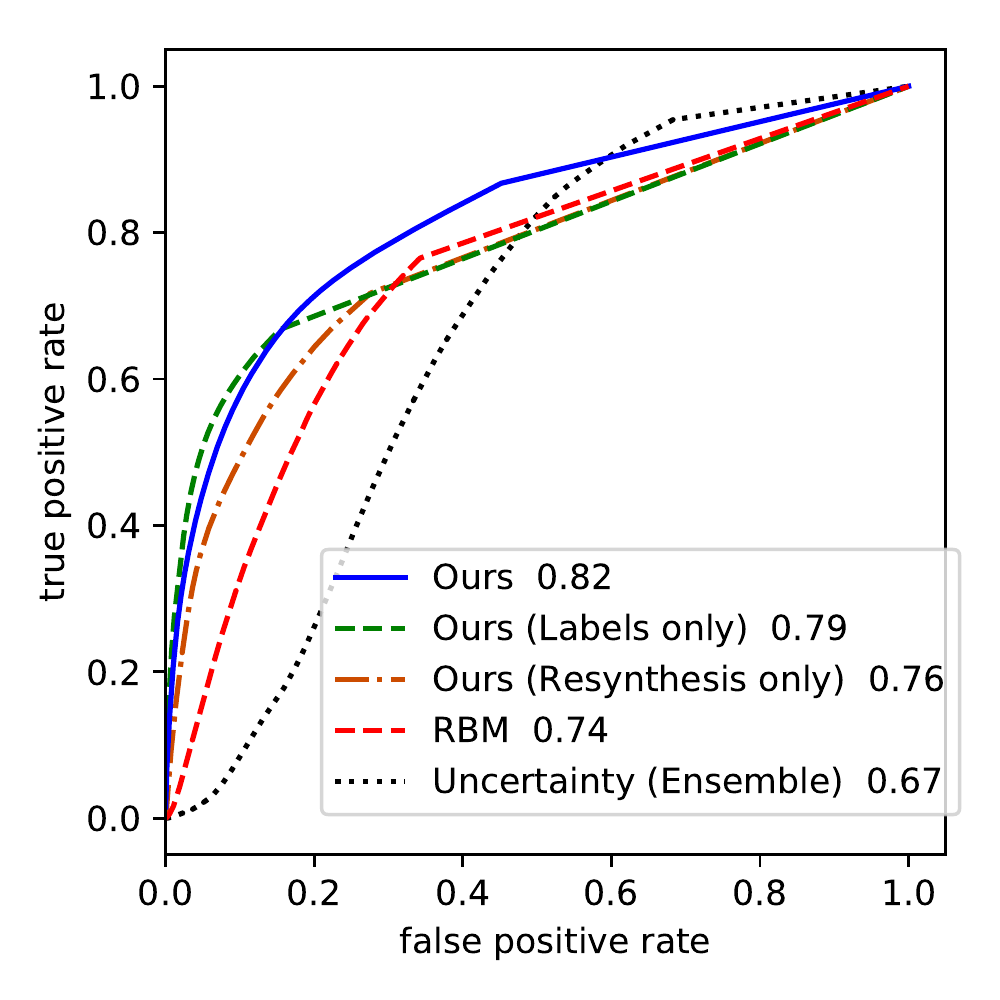} &
		\includegraphics[width=0.32\linewidth]{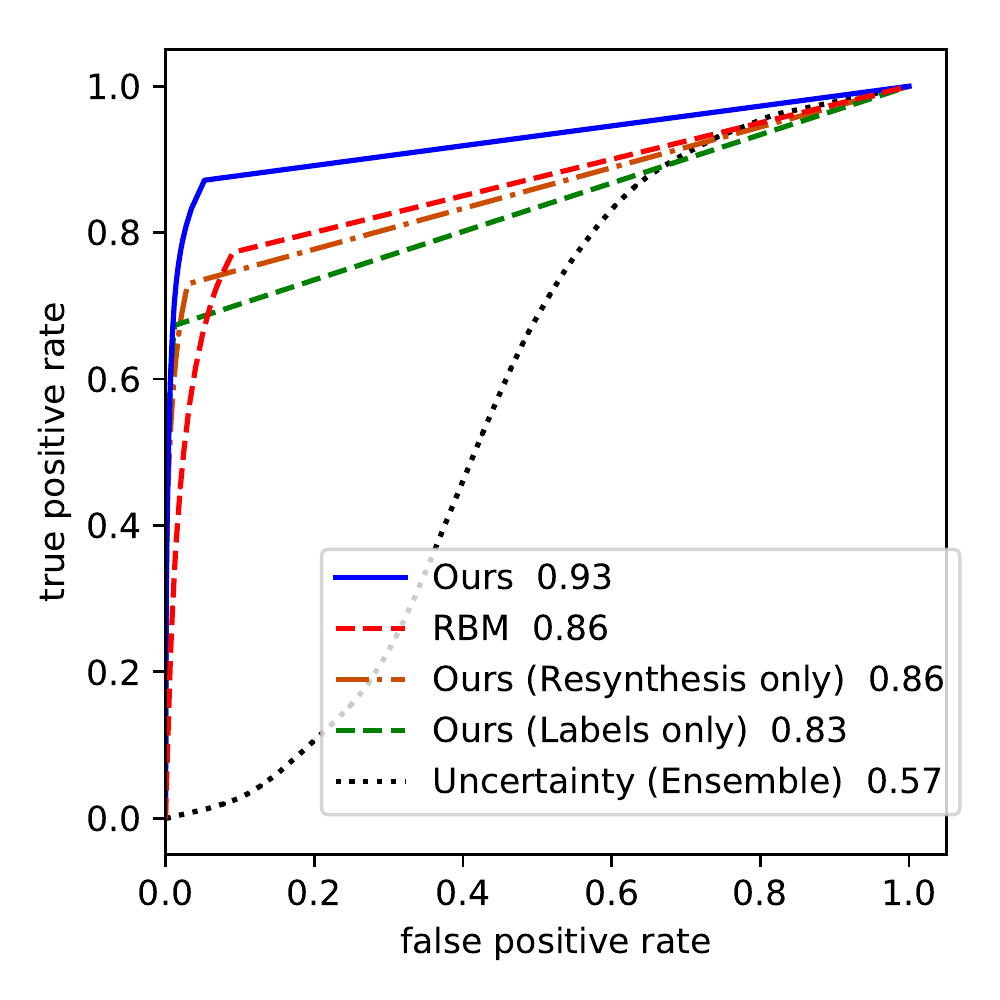} &
		\includegraphics[width=0.32\linewidth]{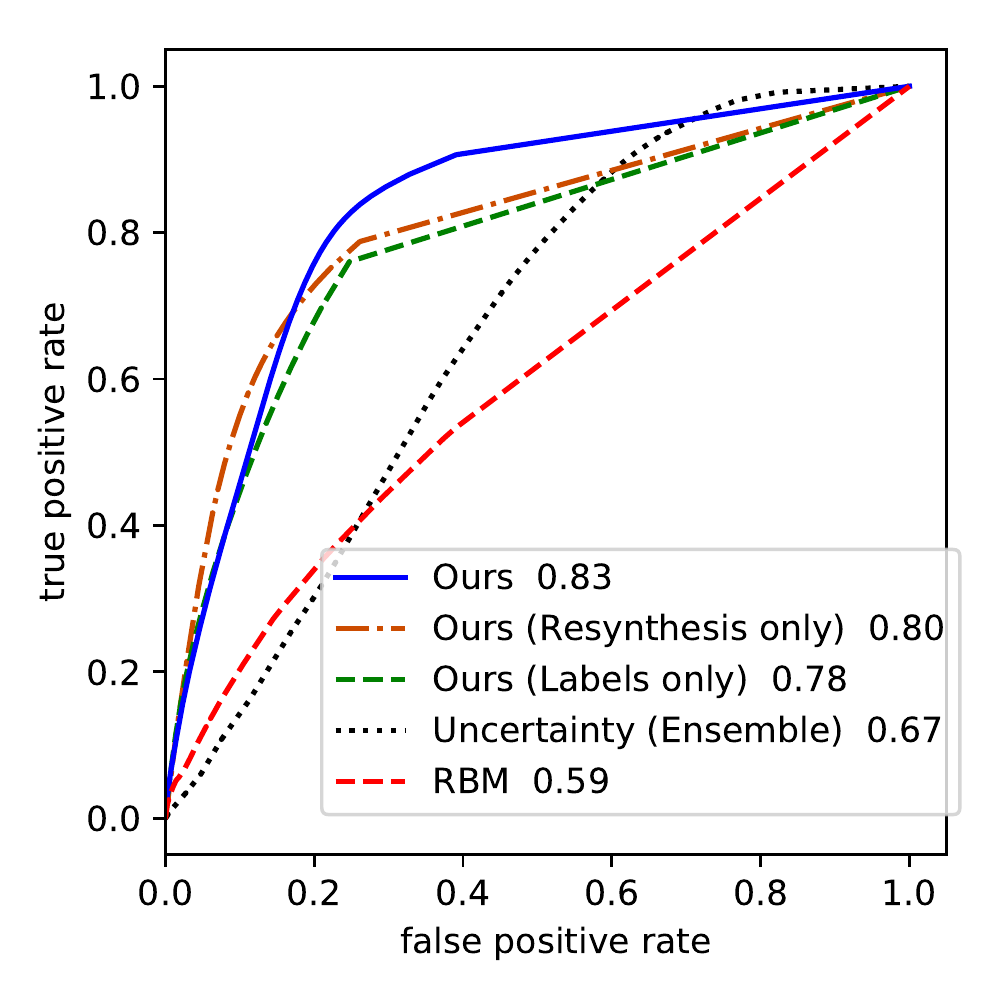} \\
		\small{PSP Net} & \small{PSP Net} & \small{PSP Net} \\
	\end{tabular}
	\vspace{-3mm}
	\caption{
		{\bf ROC curves for unexpected object detection.}
		The first two columns show results for the {\it Lost and Found}~\cite{\LAF} dataset:
		The curves on the left were computed over the entire images, excluding only the ego-vehicle. Those in the middle were obtained by restricting evaluation to the road, as defined by the ground-truth annotations.
		The right column depicts the results on our {\it Road Anomaly} dataset.
		The top and bottom rows depict the results of the {\it Bayesian SegNet} and the {\it PSP Net}, respectively.
		The methods are ordered according to their AUROC scores, provided on the right of the methods' name.	
	}
	\label{fig:rocs_all}
\end{figure*}

The {\it Lost And Found}~\cite{\LAF} dataset contains images of small items, such as cargo and toys, left on the street,
with per-pixel annotations of the obstacle and the free-space in front of the car. We perform our evaluation using the test set, excluding 17 frames for which the annotations are missing. 
We downscaled the images to $1024 \times 512$ to match the size of our training images and selected a region of interest which excludes the ego-vehicle and recording artifacts at the image boundaries.
We do not compare our results against the stereo-based ones introduced in~\cite{\LAF} because our study focuses on monocular approaches.

The ROC curves of our approach and of the baselines are shown in the left column of Fig.~\ref{fig:rocs_all}. %
Our method outperforms the baselines in both cases.  The Labels-only and Resynthesis-only variants of our approach show lower accuracy but remain competitive. By contrast, the uncertainty-based methods prove to be ill-suited for this task. Qualitative examples are provided in Fig.~\ref{fig:samples_laf}. 
Note that, while our method still produces false positives, albeit much fewer than the baselines, some of them are valid unexpected objects, such as the garbage bin in the first image. These objects, however, were not annotated as obstacles in the dataset.

Since the RBM method of~\cite{Creusot15} is specifically trained to reconstruct the road, 
we further restricted the evaluation to the road area. To this end, 
we defined the region of interest as the union of the {\it obstacle} and {\it freespace} annotations of {\it Lost And Found}.
The resulting ROC curves are shown in the middle column of Fig.~\ref{fig:rocs_all}. 
The globally-higher scores in this scenario show that distinguishing anomalies from only the road is easier than finding them in the entire scene. 
While the RBM approach significantly improves in this scenario, our method still outperforms it.

\providecommand{\localwidth}{}
\renewcommand{\localwidth}{0.30\linewidth}

\begin{figure*}[t]
	\centering
        \begin{tabular}{ccc}
		\includegraphics[width=\localwidth]{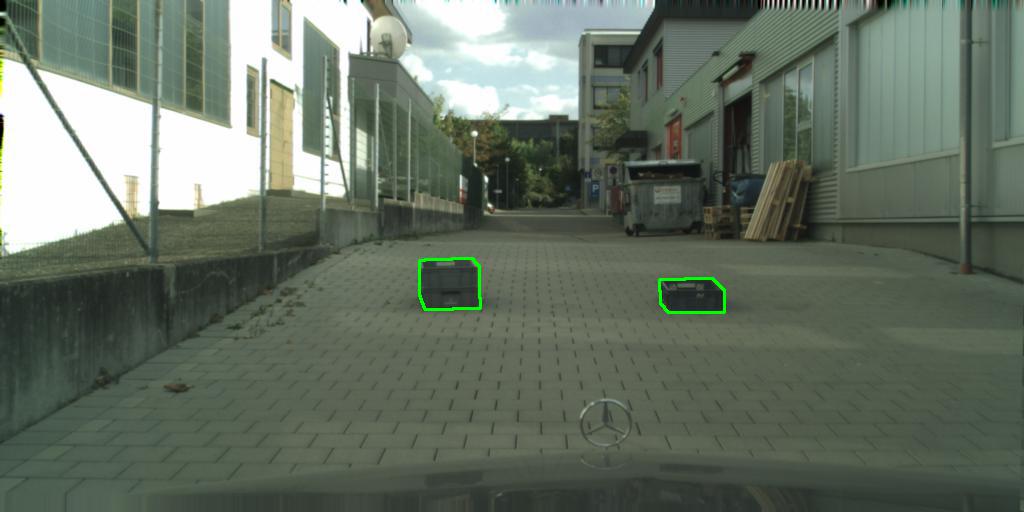}&
		\includegraphics[width=\localwidth]{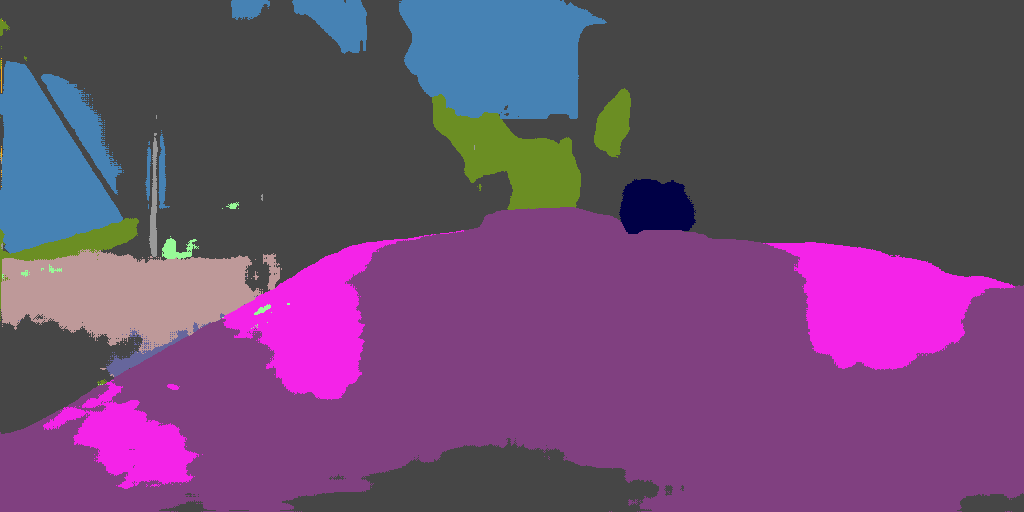}&
		\includegraphics[width=\localwidth]{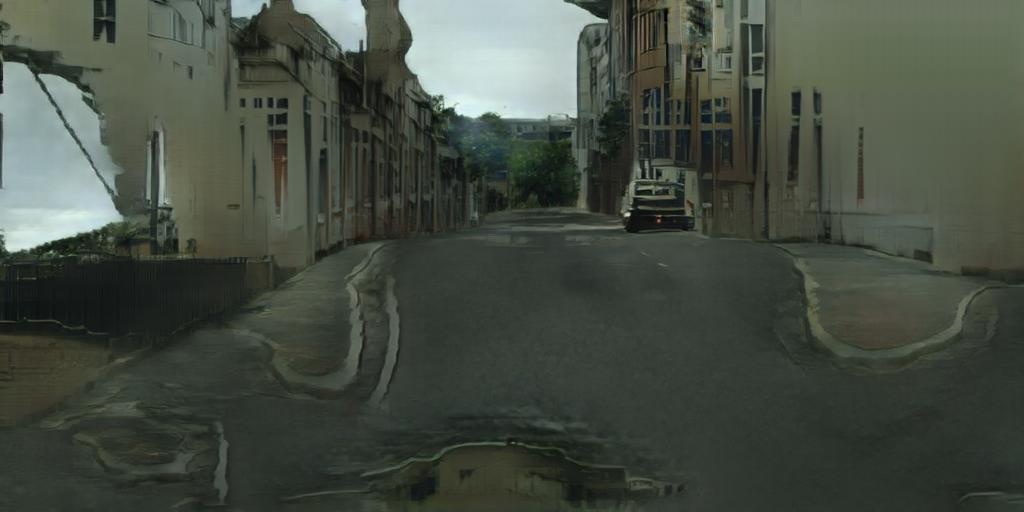}\\[-1mm]
				\small{Input image with anomalies highlighted}&
				\small{Predicted semantic map}&
				\small{Resynthesized image}\\
		\includegraphics[width=\localwidth]{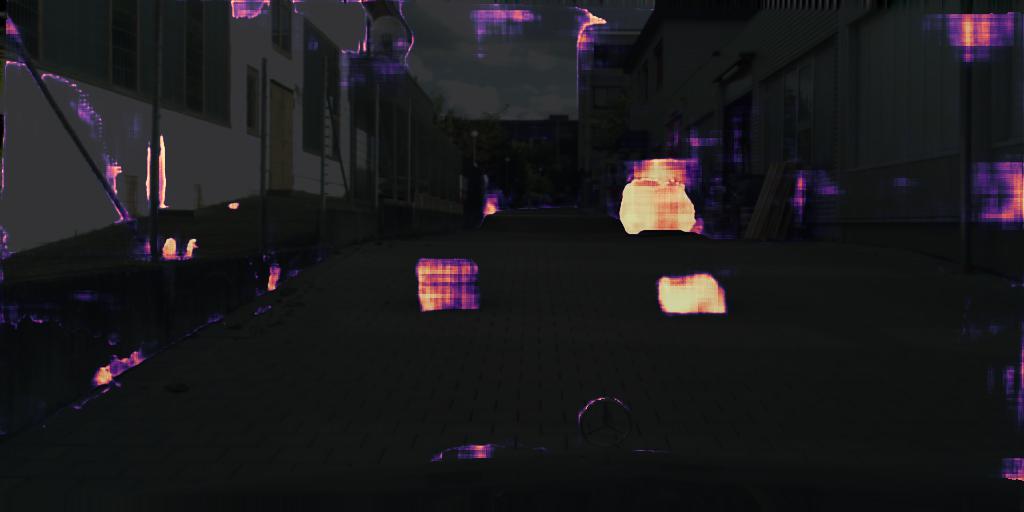}&
		\includegraphics[width=\localwidth]{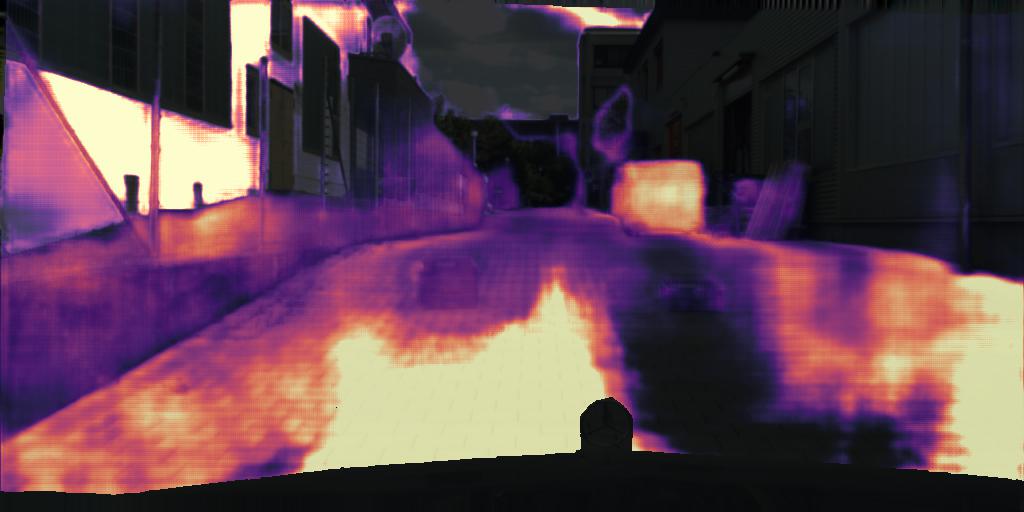}&
		\includegraphics[width=\localwidth]{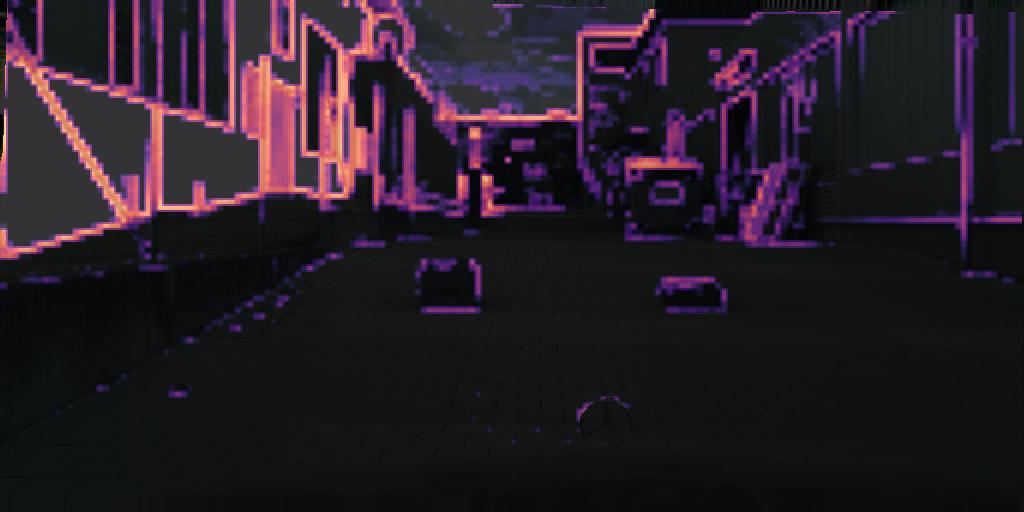}\\[-1mm]
				\small{Anomaly score - \textit{Ours}}&
				\small{Anomaly score - \textit{Uncertainty (Dropout)}}&
				\small{Anomaly score - \textit{RBM}}\\
	\end{tabular}
	\vspace{-3mm}
	\caption{{\bf Lost and Found results.} The top images depict algorithmic steps and the bottom ones our results along with those of the baselines.
		Our detector finds not only the obstacles on the road but also other unusual objects like the trash container on the right side of the road. %
		By contrast \textit{Uncertainty (Dropout)} reports high uncertainty in irrelevant regions and fails to localize the obstacles. 
		\textit{RBM} finds only the edges of the obstacles. %
		Our approach detects the unexpected objects correctly.
	}
	
	\label{fig:samples_laf}
\end{figure*}

\subsubsection{Our Road Anomaly Dataset}\label{sec:dataset_anomalies}

Motivated by the scarcity of available data for unexpected object detection,
we collected online images depicting anomalous objects, such as 
animals, rocks, lost tires, trash cans, and construction equipment, located on or near the road.
We then produced per-pixel annotations of these unexpected objects manually, using the {\it Grab Cut} algorithm~\cite{\GrabCut} to speed up the process.
The dataset contains 60 images  rescaled to a uniform size of $1280 \times 720$.
We will make this dataset and the labeling tool publicly available.

The results on this dataset are shown in the right column of Fig.~\ref{fig:rocs_all}, with example images in Fig.~\ref{fig:samples_road_anomaly}.
Our approach outperforms the baselines, demonstrating its ability to generalize to new environments. By contrast, the \textit{RBM} method's performance is strongly affected by the presence of road textures that differ significantly from the Cityscapes ones.

\providecommand{\localwidth}{}
\renewcommand{\localwidth}{0.30\linewidth}

\begin{figure*}[t]
	\centering
         \begin{tabular}{ccc}
		\includegraphics[width=\localwidth]{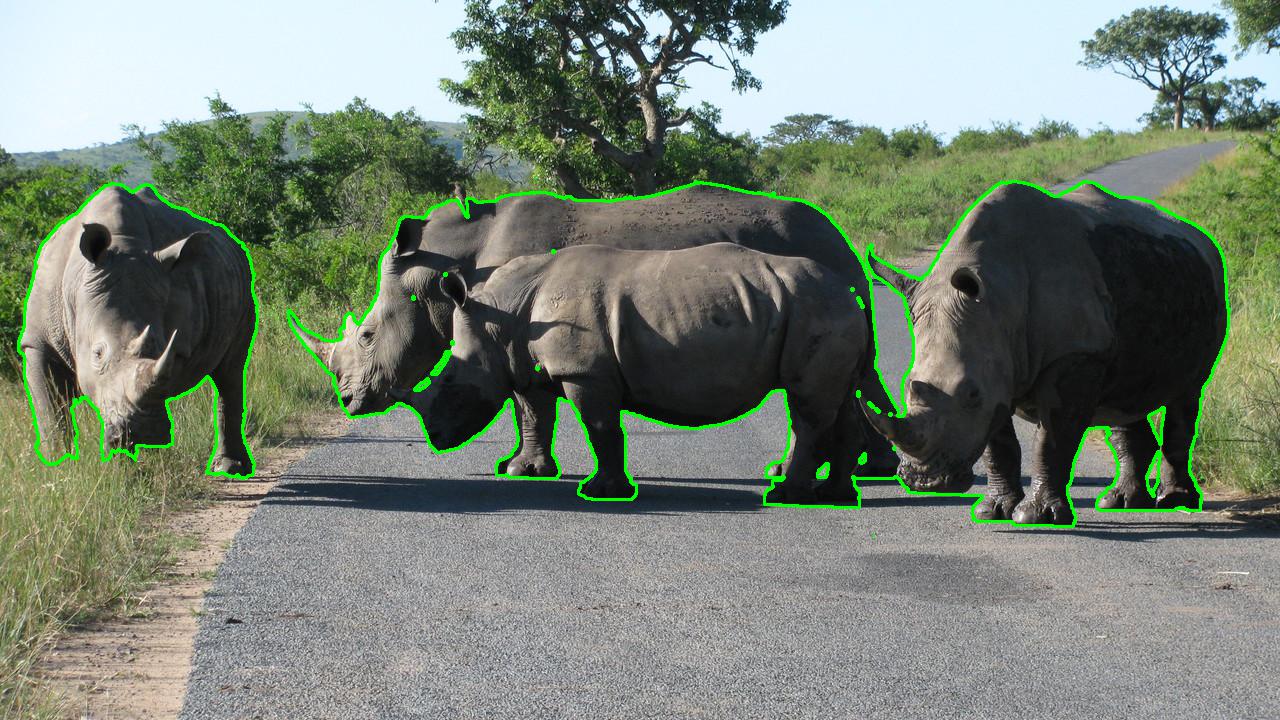}&
		\includegraphics[width=\localwidth]{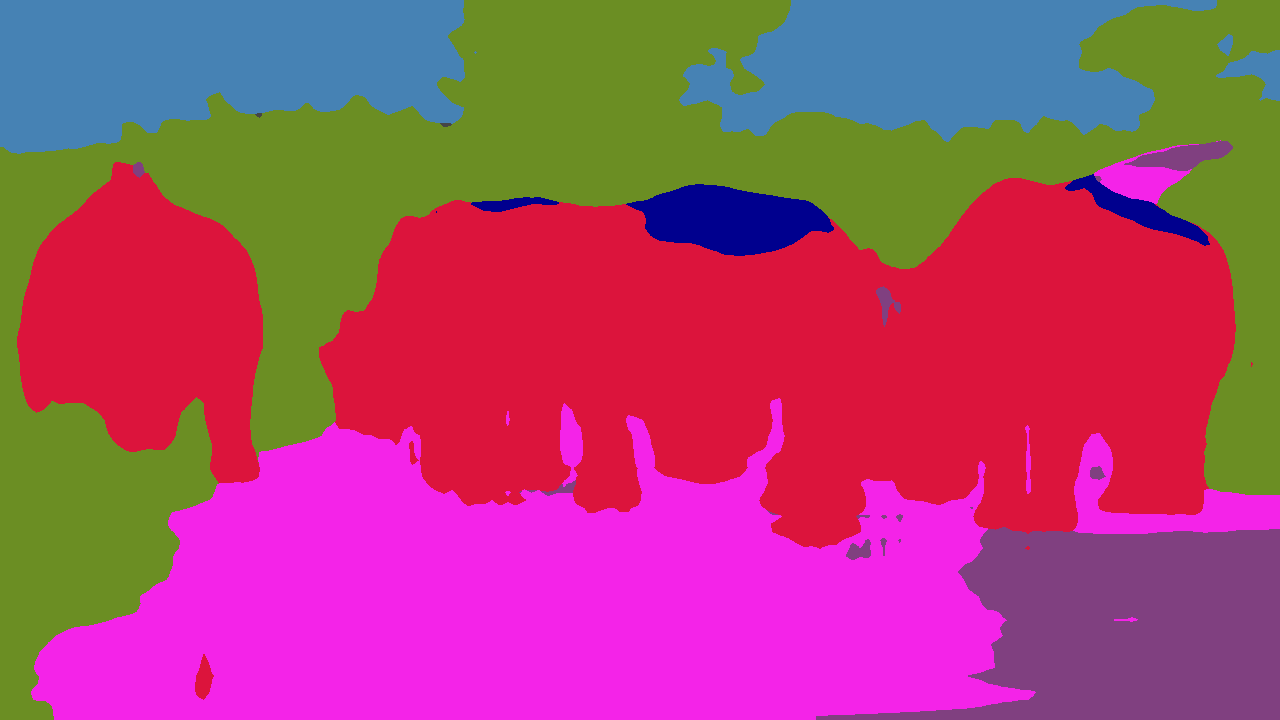}&
		\includegraphics[width=\localwidth]{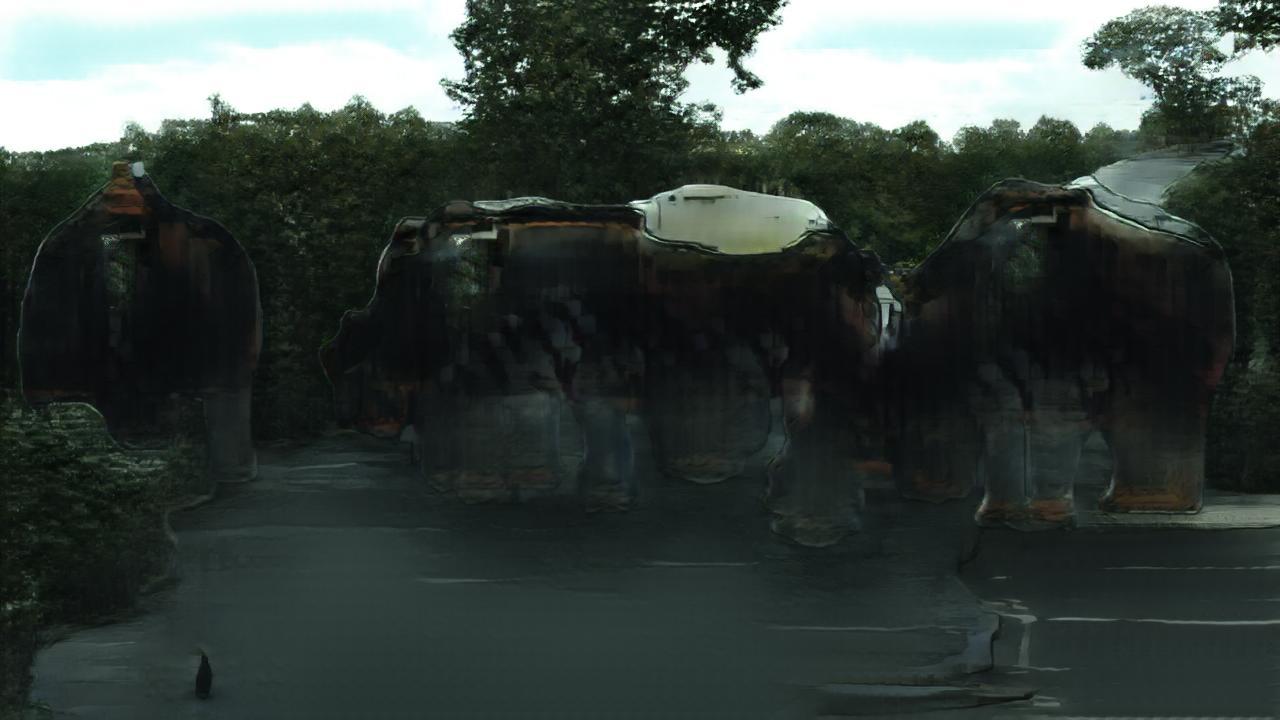}\\[-1mm]
                \small{Input image with anomalies highlighted}&
                \small{Predicted semantic map}&
				\small{Resynthesized image}\\
		\includegraphics[width=\localwidth]{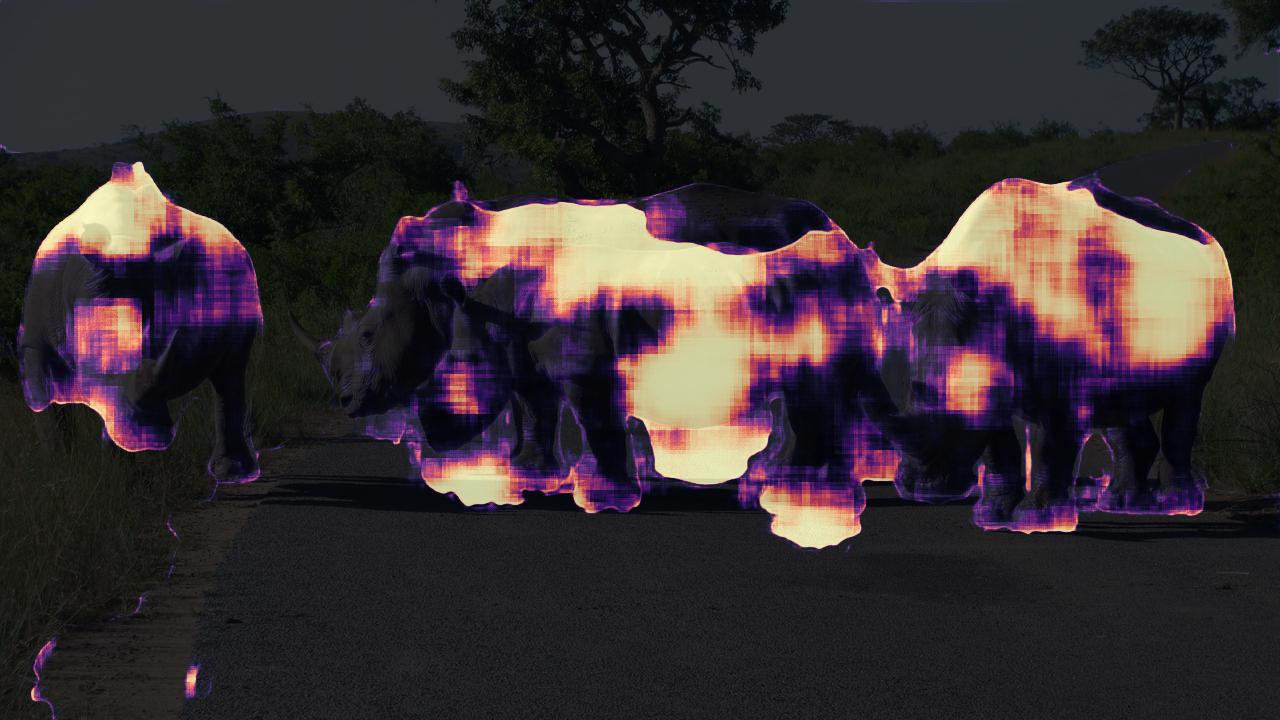}&
		\includegraphics[width=\localwidth]{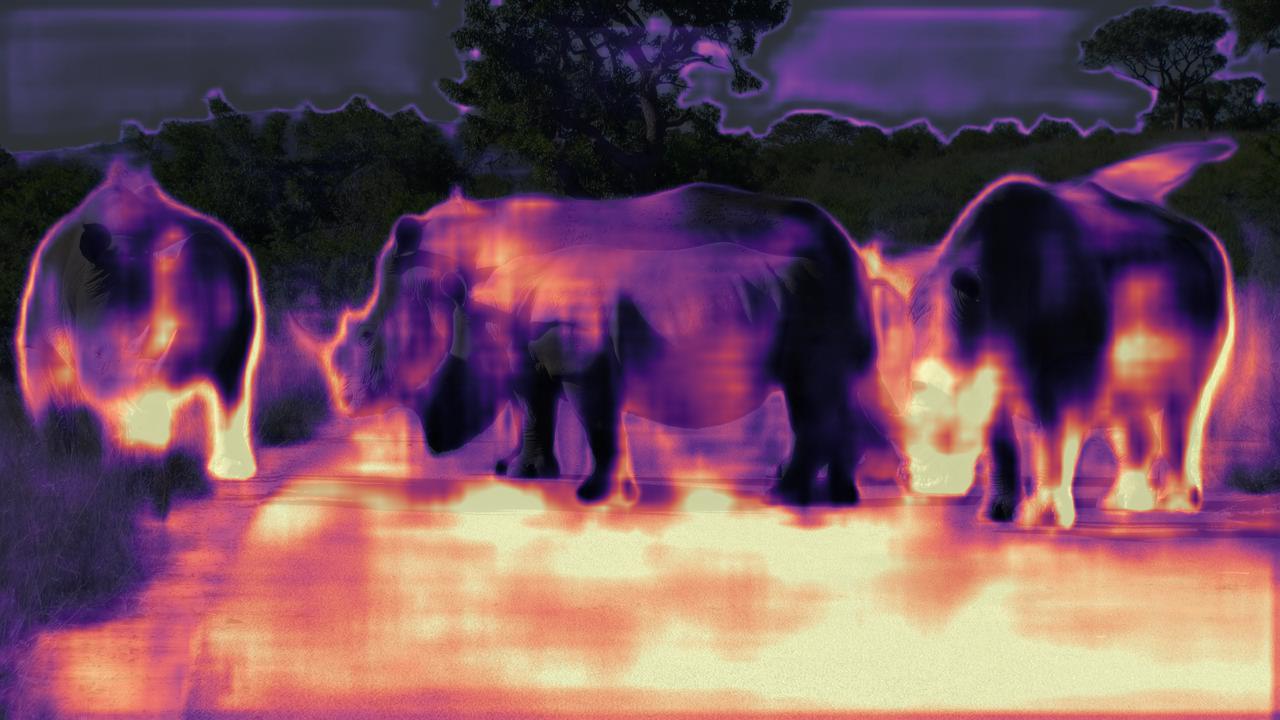}&
		\includegraphics[width=\localwidth]{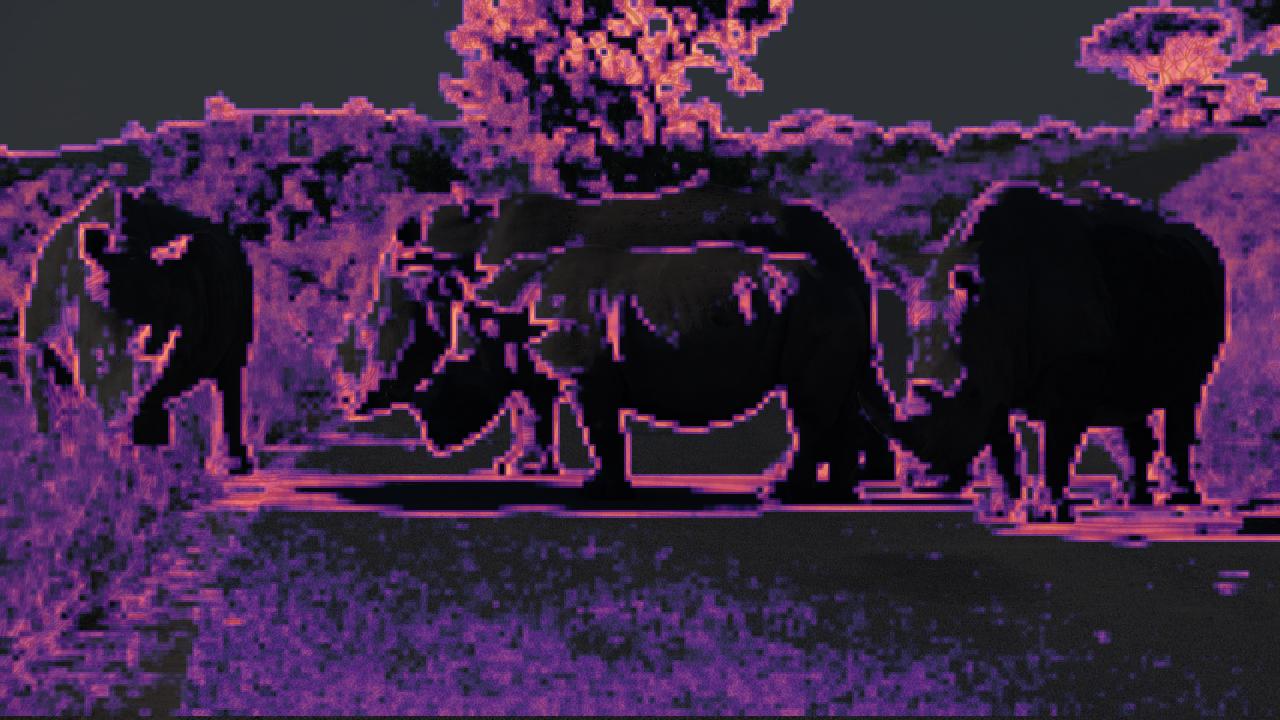}\\[-1mm]
				\small{Anomaly score - \textit{Ours}}&
                \small{Anomaly score - \textit{Uncertainty (Ensemble)}}&
				\small{Anomaly score - \textit{RBM}}\\
		\includegraphics[width=\localwidth]{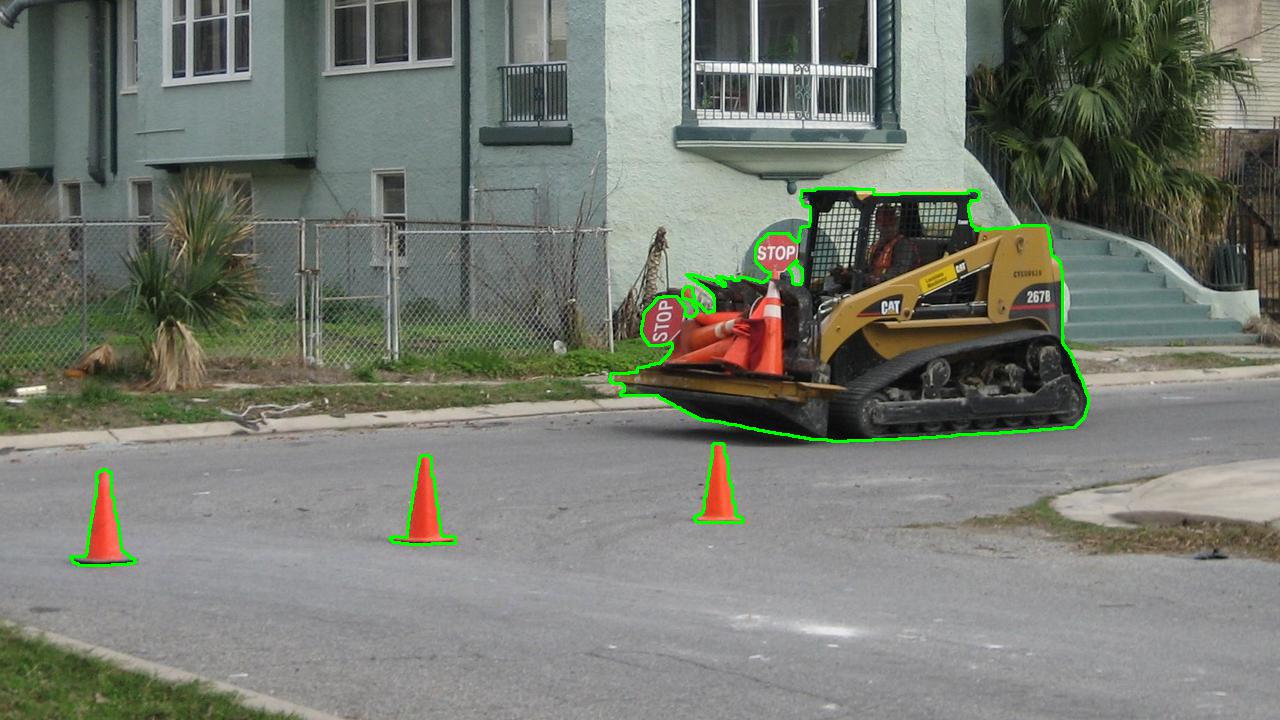}&
		\includegraphics[width=\localwidth]{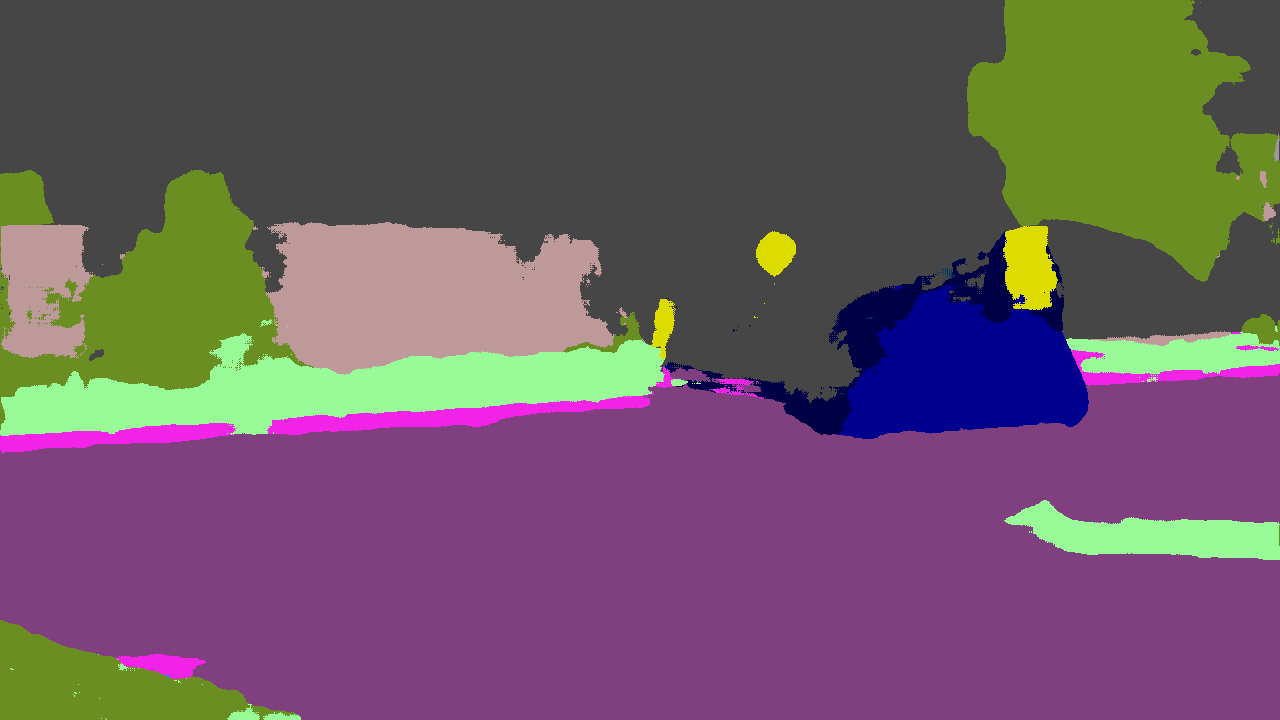}&
		\includegraphics[width=\localwidth]{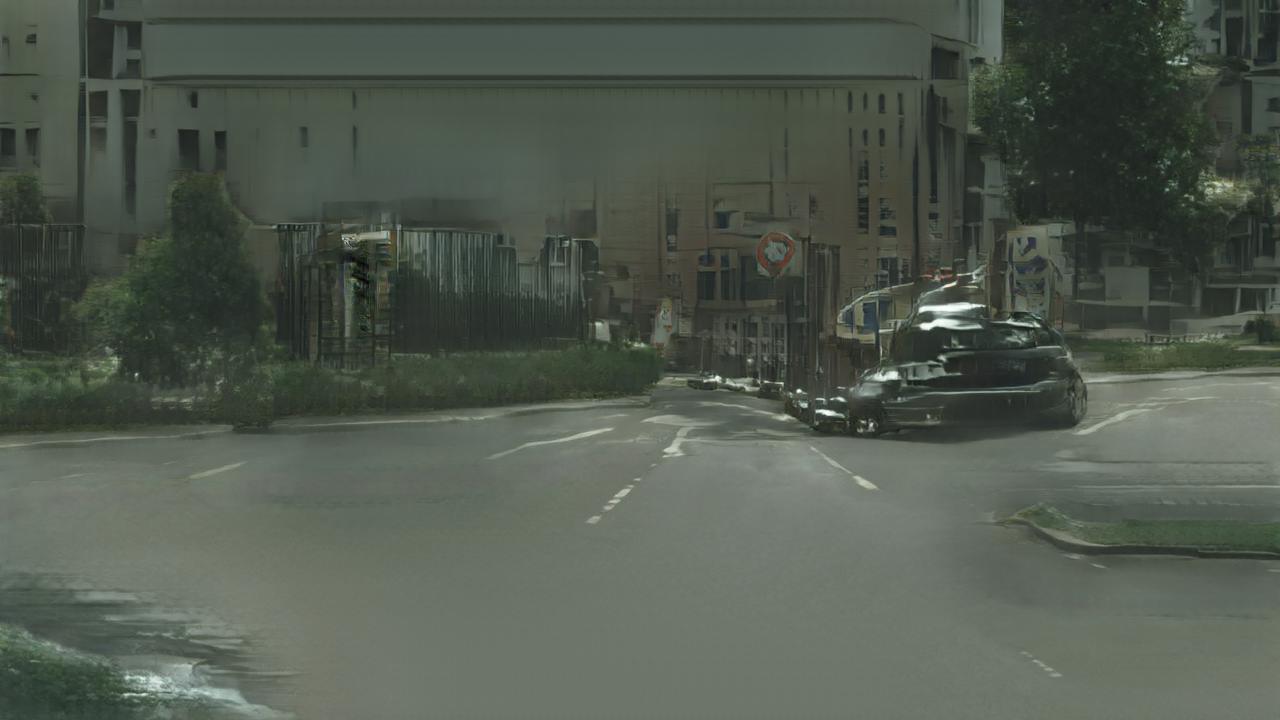}\\[-1mm]
                \small{Input image with anomalies highlighted}&
                \small{Predicted semantic map}&
				\small{Resynthesized image}\\
		\includegraphics[width=\localwidth]{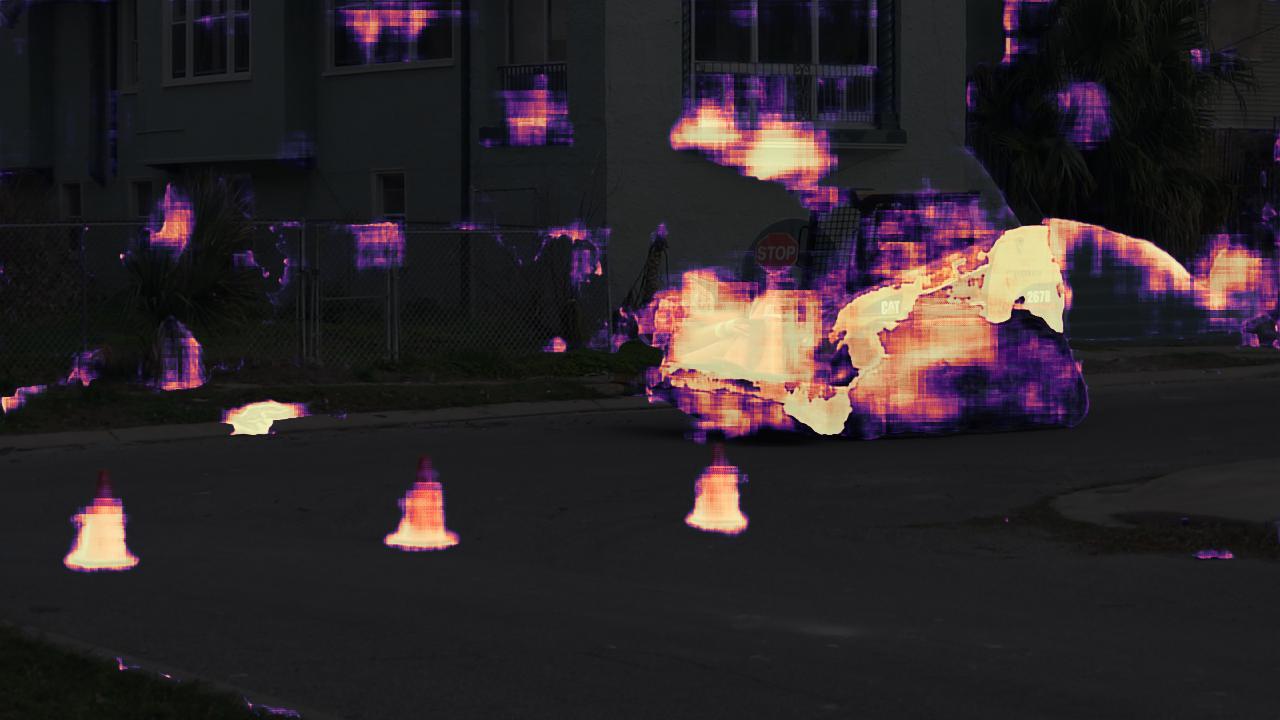}&
		\includegraphics[width=\localwidth]{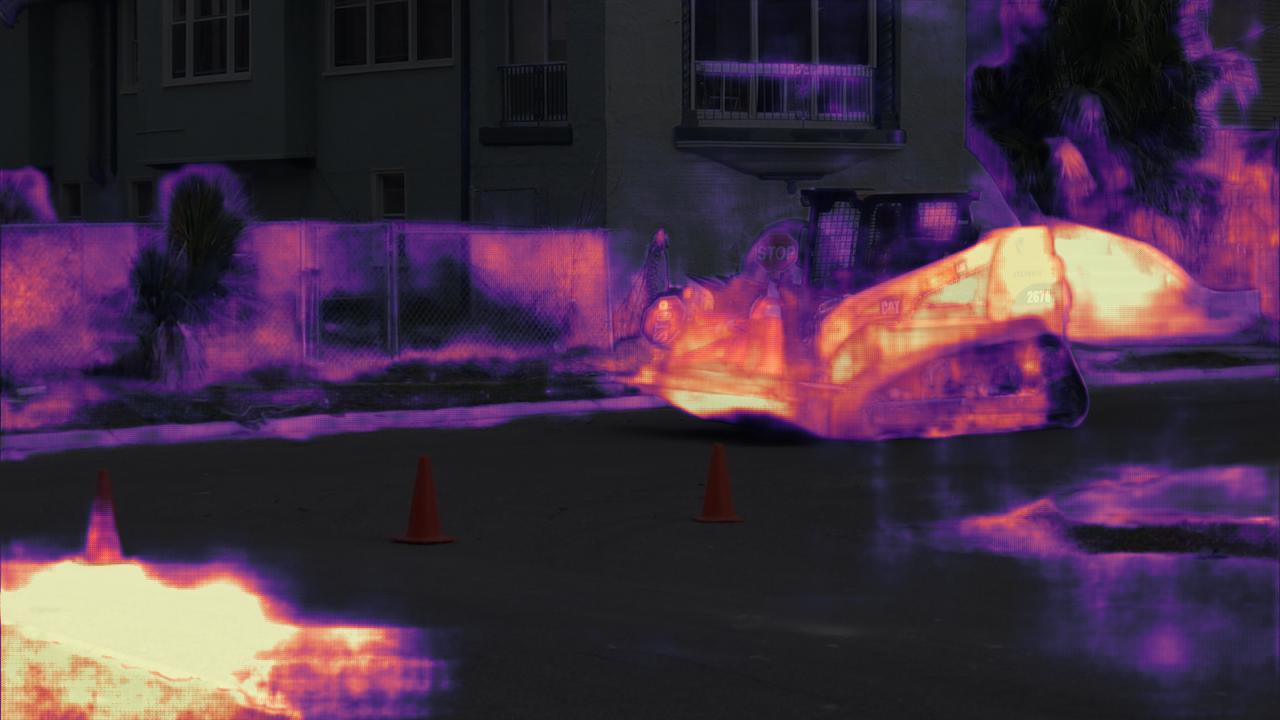}&
		\includegraphics[width=\localwidth]{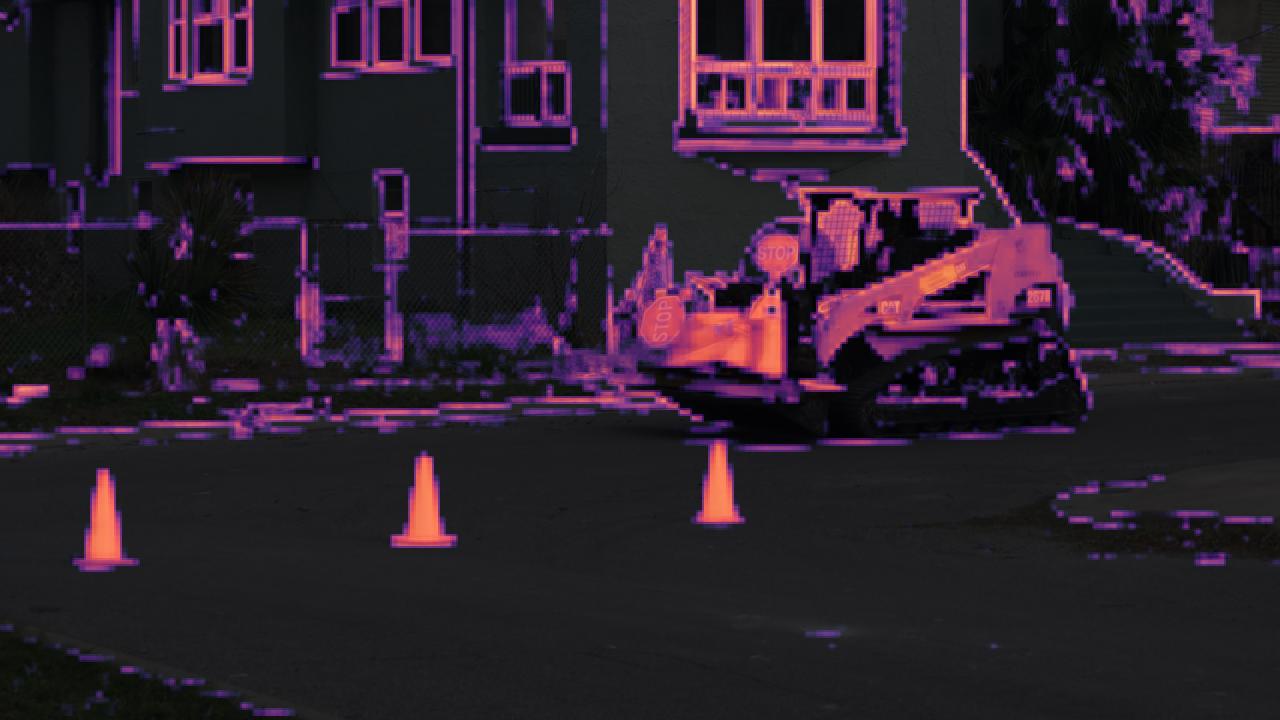}\\[-1mm]
				\small{Anomaly score - \textit{Ours}}&
                \small{Anomaly score - \textit{Uncertainty (Dropout)}}&
		        \small{Anomaly score - \textit{RBM}}
	\end{tabular}
	\vspace{-3mm}
	\caption{
		{\bf Road anomaly results.} As in Fig.~\ref{fig:samples_laf}, in each pairs of rows, the consecutive images at the top depict algorithmic steps and the ones at the bottom our results along with those of the baselines.
	}
	\label{fig:samples_road_anomaly}
\end{figure*}

\subsection{Adversarial Attack Detection}

We now evaluate our approach to detecting attacks using the two types of attack that have been used in the context of semantic segmentation.

\noindent\textbf{Adversarial Attacks:}  For semantic segmentation, the two state-of-the-art attack strategies are Dense Adversary Generation (DAG)~\cite{Xie17} and Houdini~\cite{cisse2017}.  While DAG is an iterative gradient-based method,  Houdini combines the standard task loss with an additional stochastic margin factor between the score of the actual and predicted semantic maps to yield less perturbed images. Following~\cite{Xiao18}, we generate adversarial examples with two different target semantic maps. In the first case (Shift), we shift the predicted label at each pixel by a constant offset and use the resulting label as target. In the second case (Pure), a single random  label is chosen as target for all pixels, thus generating a pure semantic map. We generate adversarial samples on the validation sets of the Cityscapes and BDD100K datasets, yielding 500 and 1000 images, respectively, with every normal sample having an attacked counterpart.

\noindent\textbf{Results:}
We compare our method with the state-of-the-art spatial consistency (SC) work of~\cite{Xiao18}, which crops random overlapping patches and computes the mean Intersection over Union (mIoU) of the overlapping regions. 

\begin{figure*}[t!]
	\centering
	\begin{subfigure}[t]{0.23\linewidth}
		\begin{subfigure}[t]{\linewidth}
		\includegraphics[width=\linewidth]{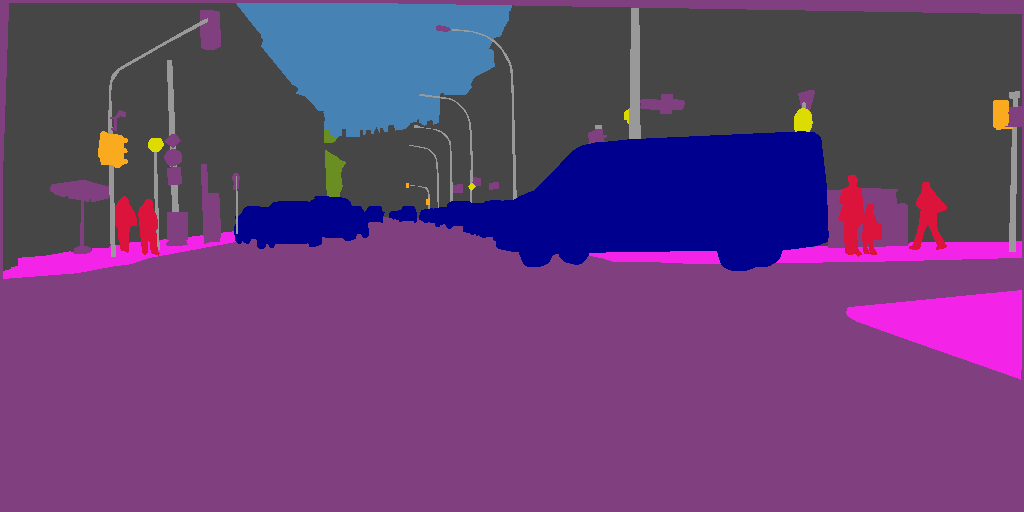}
		\caption{Ground truth map}
	\end{subfigure}
	\begin{subfigure}[t]{\linewidth}
	\includegraphics[width=\linewidth]{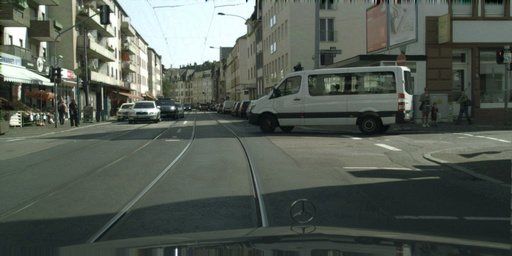}
	\caption{Input image (normal)}
\end{subfigure}
	\end{subfigure}
	\begin{subfigure}[t]{0.23\linewidth}
		\begin{subfigure}[t]{\linewidth}
			\includegraphics[width=\linewidth]{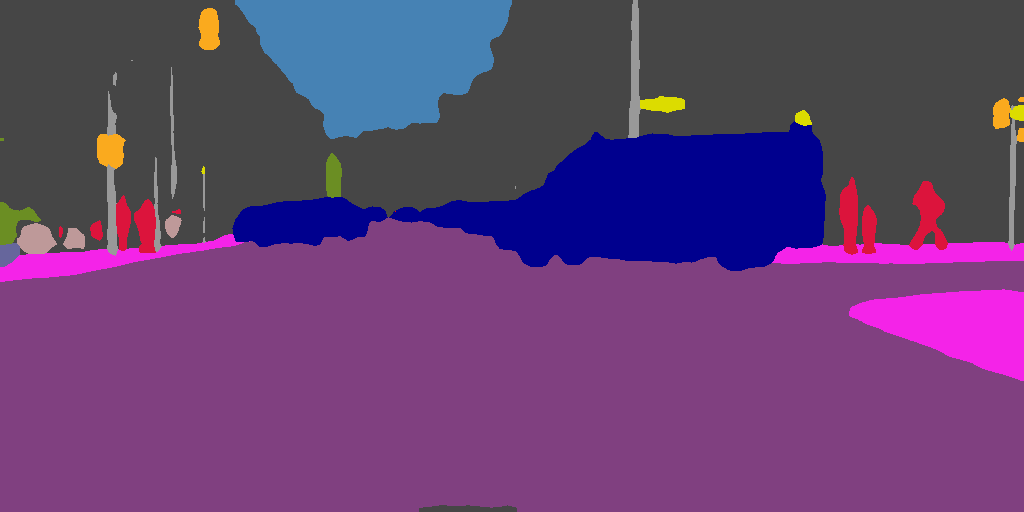}
			\caption{Predicted map (normal) }
		\end{subfigure}
		\begin{subfigure}[t]{\linewidth}
			\includegraphics[width=\linewidth]{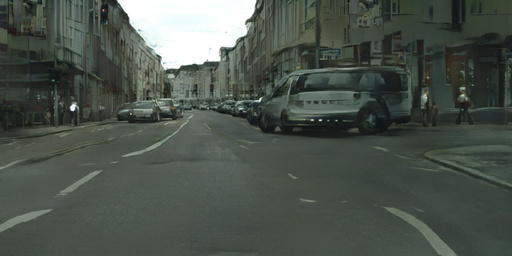}
			\caption{Resynthesized (normal)}
		\end{subfigure}
	\end{subfigure}
   \begin{subfigure}[t]{0.23\linewidth}
	\begin{subfigure}[t]{\linewidth}
		\includegraphics[width=\linewidth]{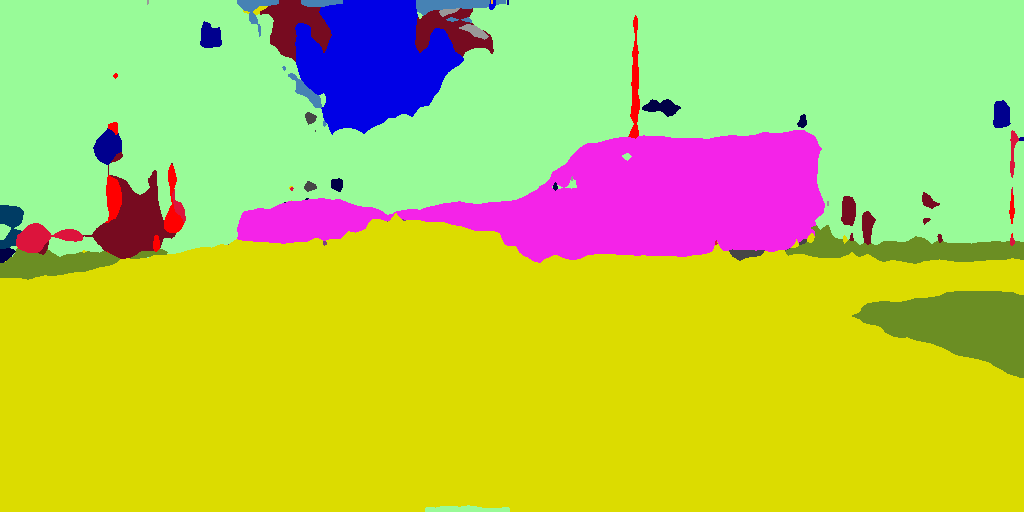}
		\caption{Predicted map (Shift)}
	\end{subfigure}
	\begin{subfigure}[t]{\linewidth}
		\includegraphics[width=\linewidth]{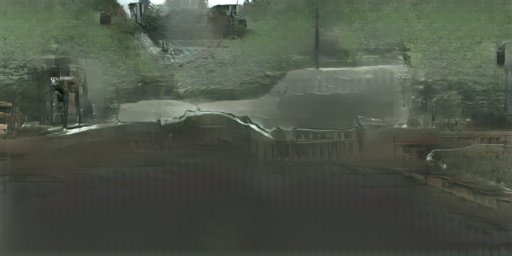}
		\caption{Resynthesized image (Shift)}
	\end{subfigure}
   \end{subfigure}
	\begin{subfigure}[t]{0.23\linewidth}
	\begin{subfigure}[t]{\linewidth}
		\includegraphics[width=\linewidth]{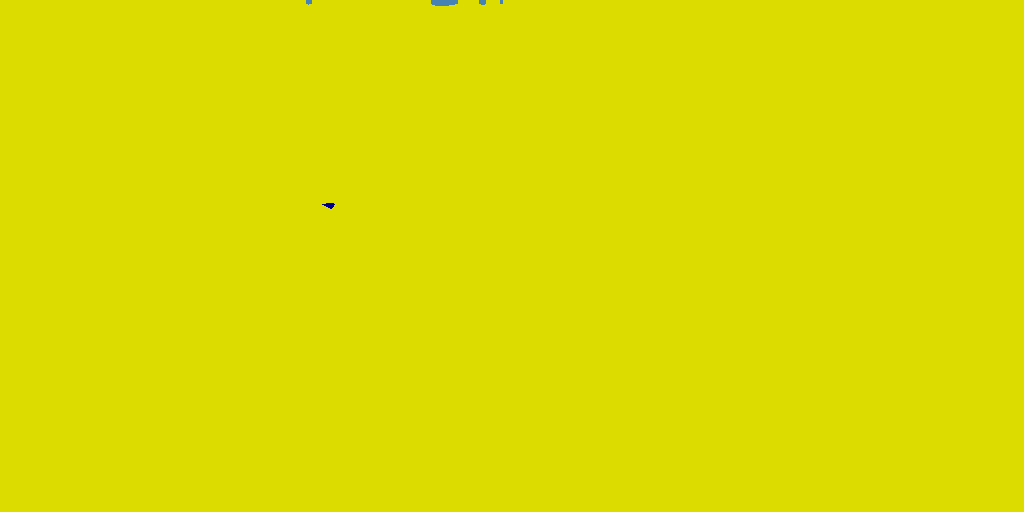}
		\caption{Predicted map (Pure)}
	\end{subfigure}		
	\begin{subfigure}[t]{\linewidth}
		\includegraphics[width=\linewidth]{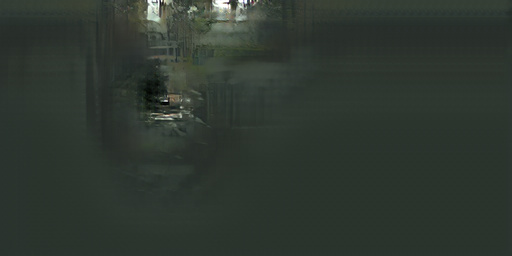}
		\caption{Resynthesized image (Pure)}
	\end{subfigure}
\end{subfigure}
	\caption{ \textbf{Visualizing adversarial attacks.} Without attacks, the resynthesized image {\bf (d)}, obtained from {\bf (c)}, looks similar to the input one {\bf (b)}. By contrast, resynthesized images ({\bf (f)} and {\bf (h)}) obtained from the semantic maps ({\bf (e)} and {\bf (g)}) computed from an attacked input differ massively from the original one.}
	\label{fig:advresults}
\end{figure*}

The results of this comparison are provided in Table~\ref{tbl:advresults}.
Our approach outperforms SC on Cityscapes and performs on par with it on BDD100K despite our use of a Cityscapes-trained generator to resynthesize the images.
Note that, in contrast with SC, which requires comparing 50 pairs of patches to detect the attack, our approach only requires a single forward pass through the segmentation and generator networks. In Fig.~\ref{fig:advresults}, we show the resynthesized images produced when using adversarial samples. Note that they massively differ from the input one. More examples are provided in the supplementary material.

\begin{table}[t]
	\resizebox{\linewidth}{!}{%
		
		\begin{tabular}{|c|c|c|c|c|c|c|}
			\toprule{\multirow{3}{*}{Dataset}} &
			\multicolumn{1}{c|}{\multirow{3}{*}{Model}} &
			\multicolumn{1}{c|}{\multirow{3}{*}{Method}} &
			\multicolumn{4}{c|}{{Detection} }\\
			\multicolumn{1}{|c|}{} &&&  \multicolumn{2}{c}{DAG} & \multicolumn{2}{c|}{Houdini}  \\ 
			\multicolumn{1}{|c|}{} &&&  Pure & Shift & Pure & Shift \\ \midrule
			\multirow{4}{*}{\shortstack{Cityscapes} } 
			& \multirow{2}{*}{\shortstack{BSeg} }  
			& SC   & 99\%  &98\%&\textbf{100\%}&\textbf{98\%} \\ 
			&& \textbf{Ours}  & \textbf{100\%} &\textbf{100\%}&\textbf{100\%}& \textbf{98\%}\\ 
			\cmidrule{2-7} 
			& \multirow{2}{*}{\shortstack{PSP} }   
			& SC &98\%&90\%&98\%&\textbf{100\%}\\ 
			&& \textbf{Ours}   & \textbf{100\%}& \textbf{99\%} &\textbf{99\%}& \textbf{100\%} \\ 
			\midrule
			\multirow{4}{*}{\shortstack{BDD} }  
			& \multirow{2}{*}{\shortstack{BSeg} } 
			& SC   &\textbf{100\%}  &\textbf{100\%}&98\%&\textbf{100\%}\\ 
			&& \textbf{Ours}  & 98\% &98\%&\textbf{100\%}&{90\%} \\
			\cmidrule{2-7} 
			& \multirow{2}{*}{\shortstack{PSP} }  
			& SC &92\%&\textbf{100\%}&96\%&\textbf{100\%}\\ 
			&& \textbf{Ours}   &\textbf{100\%} &96\%& \textbf{98\%}& {{ 95\%}} \\ 
			\hline
		\end{tabular}%
	}
	\caption{ {\small \textbf{Attack detection on Cityscapes and BDD100K.} Our method achieves higher AUROC on Cityscapes than SC and comparable ones on BDD100K, despite the fact that we rely on a generator trained on Cityscapes.}}
	\label{tbl:advresults}
\end{table}

\section{Conclusion}\label{sec:conclusion}
In this paper, we have introduced a drastically new approach to detecting the unexpected in images. Our method is built on the intuition that, because unexpected objects have not been seen during training, typical semantic segmentation networks will produce spurious labels in the corresponding regions. Therefore, resynthesizing an image from the semantic map will yield discrepancies with respect to the input image, and we introduced a network that learns to detect the meaningful ones. Our experiments have shown that our approach detects the unexpected objects much more reliably than uncertainty- and autoencoder-based techniques. We have also contributed a new dataset with annotated road anomalies, which we believe will facilitate research in this relatively unexplored field. Our approach still suffers from the presence of some false positives, which, in a real autonomous driving scenario would create a source of distraction. Reducing this false positive rate will therefore be the focus of our future research.

\FloatBarrier

{\small
\bibliographystyle{ieee}
\bibliography{string,vision,learning,biomed}

\begin{thebibliography}{10}\itemsep=-1pt

\bibitem{Akcay18}
S.~Akcay, A.~A. Abarghouei, and T.~P. Breckon.
\newblock {Ganomaly: Semi-Supervised Anomaly Detection via Adversarial
  Training}.
\newblock {\em arXiv Preprint}, abs/1805.06725, 2018.

\bibitem{Badrinarayanan15}
V.~Badrinarayanan, A.~Kendall, and R.~Cipolla.
\newblock {SegNet: A Deep Convolutional Encoder-Decoder Architecture for Image
  Segmentation}.
\newblock {\em arXiv Preprint}, 2015.

\bibitem{carlini2017}
N.~Carlini and D.~Wagner.
\newblock Towards evaluating the robustness of neural networks.
\newblock In {\em 2017 IEEE Symposium on Security and Privacy (SP)}, pages
  39--57. IEEE, 2017.

\bibitem{Chen17a}
L.~Chen, G.~Papandreou, F.~Schroff, and H.~Adam.
\newblock {Rethinking Atrous Convolution for Semantic Image Segmentation}.
\newblock {\em arXiv Preprint}, abs/1706.05587, 2017.

\bibitem{Chen18c}
L.~Chen, Y.~Zhu, G.~Papandreou, F.~Schroff, and H.~Adam.
\newblock {Encoder-Decoder with Atrous Separable Convolution for Semantic Image
  Segmentation}.
\newblock {\em arXiv Preprint}, abs/1802.02611, 2018.

\bibitem{cisse2017}
M.~M. Cisse, Y.~Adi, N.~Neverova, and J.~Keshet.
\newblock Houdini: Fooling deep structured visual and speech recognition models
  with adversarial examples.
\newblock In {\em Advances in Neural Information Processing Systems}, pages
  6977--6987, 2017.

\bibitem{Cordts16}
M.~Cordts, M.~Omran, S.~Ramos, T.~Rehfeld, M.~Enzweiler, R.~Benenson,
  U.~Franke, S.~Roth, and B.~Schiele.
\newblock {The Cityscapes Dataset for Semantic Urban Scene Understanding}.
\newblock In {\em Conference on Computer Vision and Pattern Recognition}, 2016.

\bibitem{Creusot15}
C.~Creuso and A.~Munawar.
\newblock {Real-Time Small Obstacle Detection on Highways Using Compressive RBM
  Road Reconstruction}.
\newblock In {\em Intelligent Vehicles Symposium}, 2015.

\bibitem{Dalal05}
N.~Dalal and B.~Triggs.
\newblock {Histograms of Oriented Gradients for Human Detection}.
\newblock In {\em Conference on Computer Vision and Pattern Recognition}, pages
  886--893, 2005.

\bibitem{Denker91}
J.~Denker and Y.~LeCun.
\newblock {Transforming Neural-Net Output Levels to Probability Distributions}.
\newblock In {\em Advances in Neural Information Processing Systems}, 1991.

\bibitem{Gal16}
Y.~Gal and Z.~Ghahramani.
\newblock {Dropout as a Bayesian Approximation: Representing Model Uncertainty
  in Deep Learning}.
\newblock In {\em International Conference on Machine Learning}, pages
  1050--1059, 2016.

\bibitem{Gast18}
J.~Gast and S.~Roth.
\newblock {Lightweight Probabilistic Deep Networks}.
\newblock In {\em Conference on Computer Vision and Pattern Recognition}, 2018.

\bibitem{goodfellow2014a}
I.~J. Goodfellow, J.~Shlens, and C.~Szegedy.
\newblock Explaining and harnessing adversarial examples.
\newblock {\em International Conference on Learning Representations}, 2015.

\bibitem{Huang18a}
X.~Huang, X.~Cheng, Q.~Geng, B.~Cao, D.~Zhou, P.~Wang, Y.~Lin, and R.~Yang.
\newblock {The Apolloscape Dataset for Autonomous Driving}.
\newblock {\em arXiv Preprint}, 1803.06184, 2018.

\bibitem{Isobe17a}
S.~Isobe and S.~Arai.
\newblock {A Semantic Segmentation Method Using Model Uncertainty}.
\newblock In {\em IIAE International Conference on Intelligent Systems and
  Image Processing}, 2017.

\bibitem{Isobe17c}
S.~Isobe and S.~Arai.
\newblock {Deep Convolutional Encoder-Decoder Network with Model Uncertainty
  for Semantic Segmentation}.
\newblock In {\em IEEE International Conference on INnovations in Intelligent
  SysTems and Applications}, 2017.

\bibitem{Isobe17b}
S.~Isobe and S.~Arai.
\newblock {Inference with Model Uncertainty on Indoor Scene for Semantic
  Segmentation}.
\newblock In {\em IEEE Global Conference on Signal and Information Processing},
  2017.

\bibitem{Kendall15b}
A.~Kendall, V.~Badrinarayanan, and R.~Cipolla.
\newblock {Bayesian Segnet: Model Uncertainty in Deep Convolutional
  Encoder-Decoder Architectures for Scene Understanding}.
\newblock {\em arXiv Preprint}, 2015.

\bibitem{Kendall17}
A.~Kendall and Y.~Gal.
\newblock {What Uncertainties Do We Need in Bayesian Deep Learning for Computer
  Vision?}
\newblock In {\em Advances in Neural Information Processing Systems}, 2017.

\bibitem{Kingma15}
D.~Kingma and J.~Ba.
\newblock {Adam: {A} Method for Stochastic Optimisation}.
\newblock In {\em International Conference on Learning Representations}, 2015.

\bibitem{Kiran18}
B.~Kiran, D.~Thomas, and R.~Parakkal.
\newblock {An Overview of Deep Learning Based Methods for Unsupervised and
  Semi-Supervised Anomaly Detection in Videos}.
\newblock {\em Journal of Imaging}, 2018.

\bibitem{Klambauer17}
G.~Klambauer, T.~Unterthiner, A.~Mayr, and S.~Hochreiter.
\newblock Self-normalizing neural networks.
\newblock In {\em Advances in Neural Information Processing Systems}, 2017.

\bibitem{kurakin2016}
A.~Kurakin, I.~Goodfellow, and S.~Bengio.
\newblock Adversarial machine learning at scale.
\newblock {\em International Conference on Learning Representations}, 2017.

\bibitem{Lakshminarayanan17}
B.~Lakshminarayanan, A.~Pritzel, and C.~Blundell.
\newblock {Simple and Scalable Predictive Uncertainty Estimation Using Deep
  Ensembles}.
\newblock In {\em Advances in Neural Information Processing Systems}, 2017.

\bibitem{lee2018adv}
K.~Lee, K.~Lee, H.~Lee, and J.~Shin.
\newblock A simple unified framework for detecting out-of-distribution samples
  and adversarial attacks.
\newblock In {\em Advances in Neural Information Processing Systems}, pages
  7167--7177, 2018.

\bibitem{Li18b}
W.~Li, O.~Jafari, and C.~Rother.
\newblock {Deep Object Co-Segmentation}.
\newblock {\em arXiv Preprint}, abs/1804.06423, 2018.

\bibitem{ma18b}
X.~Ma, B.~Li, Y.~Wang, S.~M. Erfani, S.~Wijewickrema, G.~Schoenebeck, D.~Song,
  M.~E. Houle, and J.~Bailey.
\newblock Characterizing adversarial subspaces using local intrinsic
  dimensionality.
\newblock {\em International Conference on Learning Representations}, 2018.

\bibitem{Mackay92}
D.~MacKay.
\newblock {A Practical Bayesian Framework for Backpropagation Networks}.
\newblock {\em Neural Computation}, 4(3):448--472, 1992.

\bibitem{MacKay95}
D.~Mackay.
\newblock {Bayesian Neural Networks and Density Networks}.
\newblock {\em {Nuclear Instruments and Methods in Physics Research Section A:
  Accelerators, Spectrometers, Detectors and Associated Equipment}},
  354(1):73--80, 1995.

\bibitem{metzen2017}
J.~H. Metzen, T.~Genewein, V.~Fischer, and B.~Bischoff.
\newblock On detecting adversarial perturbations.
\newblock {\em International Conference on Learning Representations}, 2017.

\bibitem{moosavi2016}
S.-M. Moosavi-Dezfooli, A.~Fawzi, and P.~Frossard.
\newblock Deepfool: a simple and accurate method to fool deep neural networks.
\newblock In {\em Conference on Computer Vision and Pattern Recognition}, pages
  2574--2582, 2016.

\bibitem{Munawar15}
A.~Munawar and C.~Creusot.
\newblock {Structural Inpainting of Road Patches for Anomaly Detection}.
\newblock In {\em IAPR International Conference on Machine Vision
  Applications}, 2015.

\bibitem{Munawar17}
A.~Munawar, P.~Vinayavekhin, and G.~D. Magistris.
\newblock {Limiting the Reconstruction Capability of Generative Neural Network
  Using Negative Learning}.
\newblock In {\em IEEE International Workshop on Machine Learning for Signal
  Processing}, 2017.

\bibitem{Paszke16}
A.~Paszke, A.~Chaurasia, S.~Kim, and E.~Culurciello.
\newblock {Enet: A deep neural network architecture for real-time semantic
  segmentation}.
\newblock {\em arXiv Preprint}, abs/1606.02147, 2016.

\bibitem{Pinggera16}
P.~Pinggera, S.~Ramos, S.~Gehrig, U.~Franke, C.~Rother, and R.~Mester.
\newblock {Lost and Found: Detecting Small Road Hazards for Self-Driving
  Vehicles}.
\newblock In {\em International Conference on Intelligent Robots and Systems},
  2016.

\bibitem{Ravanbakhsh17}
M.~Ravanbakhsh, M.~Nabi, E.~Sangineto, L.~Marcenaro, C.~Regazzoni, and N.~Sebe.
\newblock {Abnormal Event Detection in Videos Using Generative Adversarial
  Nets}.
\newblock In {\em International Conference on Image Processing}, 2017.

\bibitem{Rother04}
C.~Rother, V.~Kolmogorov, and A.~Blake.
\newblock {"{GrabCut}" - Interactive Foreground Extraction Using Iterated Graph
  Cuts}.
\newblock In {\em ACM SIGGRAPH}, pages 309--314, 2004.

\bibitem{Schlegl17}
T.~Schlegl, P.~Seeb{\"o}ck, S.~Waldstein, U.~Schmidt-Erfurth, and G.~Langs.
\newblock {Unsupervised Anomaly Detection with Generative Adversarial Networks
  to Guide Marker Discovery}.
\newblock In {\em International Conference on Information Processing in Medical
  Imaging}, 2017.

\bibitem{Simonyan15}
K.~Simonyan and A.~Zisserman.
\newblock {Very Deep Convolutional Networks for Large-Scale Image Recognition}.
\newblock In {\em International Conference on Learning Representations}, 2015.

\bibitem{Srivastava14}
N.~Srivastava, G.~Hinton, A.~Krizhevsky, I.~Sutskever, and R.~Salakhutdinov.
\newblock {Dropout: A Simple Way to Prevent Neural Networks from Overfitting}.
\newblock {\em Journal of Machine Learning Research}, 15:1929--1958, 2014.

\bibitem{tramer2017}
F.~Tram{\`e}r, A.~Kurakin, N.~Papernot, I.~Goodfellow, D.~Boneh, and
  P.~McDaniel.
\newblock Ensemble adversarial training: Attacks and defenses.
\newblock {\em arXiv}, 2017.

\bibitem{Wang18c}
T.~Wang, M.-Y. Liu, J.-Y. Zhu, A.~Tao, J.~Kautz, and B.~Catanzaro.
\newblock {High-Resolution Image Synthesis and Semantic Manipulation with
  Conditional {GANs}}.
\newblock {\em Conference on Computer Vision and Pattern Recognition}, 2018.

\bibitem{Xiao18}
C.~Xiao, R.~Deng, B.~Li, F.~Yu, M.~Liu, and D.~Song.
\newblock Characterizing adversarial examples based on spatial consistency
  information for semantic segmentation.
\newblock In {\em European Conference on Computer Vision}, pages 217--234,
  2018.

\bibitem{Xie17}
C.~Xie, J.~Wang, Z.~Zhang, Y.~Zhou, L.~Xie, and A.~Yuille.
\newblock Adversarial examples for semantic segmentation and object detection.
\newblock In {\em International Conference on Computer Vision}, 2017.

\bibitem{Yu18b}
C.~Yu, J.~Wang, C.~Peng, C.~Gao, G.~Yu, and N.~Sang.
\newblock {Learning a Discriminative Feature Network for Semantic
  Segmentation}.
\newblock In {\em Conference on Computer Vision and Pattern Recognition}, 2018.

\bibitem{Yu18c}
F.~Yu, W.~Xian, Y.~Chen, F.~Liu, M.~Liao, V.~Madhavan, and T.~Darrell.
\newblock {{BDD100K:} {A} Diverse Driving Video Database with Scalable
  Annotation Tooling}.
\newblock {\em arXiv Preprint}, 2018.

\bibitem{Zhao17b}
H.~Zhao, J.~Shi, X.~Qi, X.~Wang, and J.~Jia.
\newblock {Pyramid Scene Parsing Network}.
\newblock In {\em Conference on Computer Vision and Pattern Recognition}, 2017.

\end{thebibliography}
}

\clearpage
\appendix
\appendixpage

\section{Detecting Unexpected Objects}

\begin{figure}[t]
	\centering
	\includegraphics[width=\linewidth]{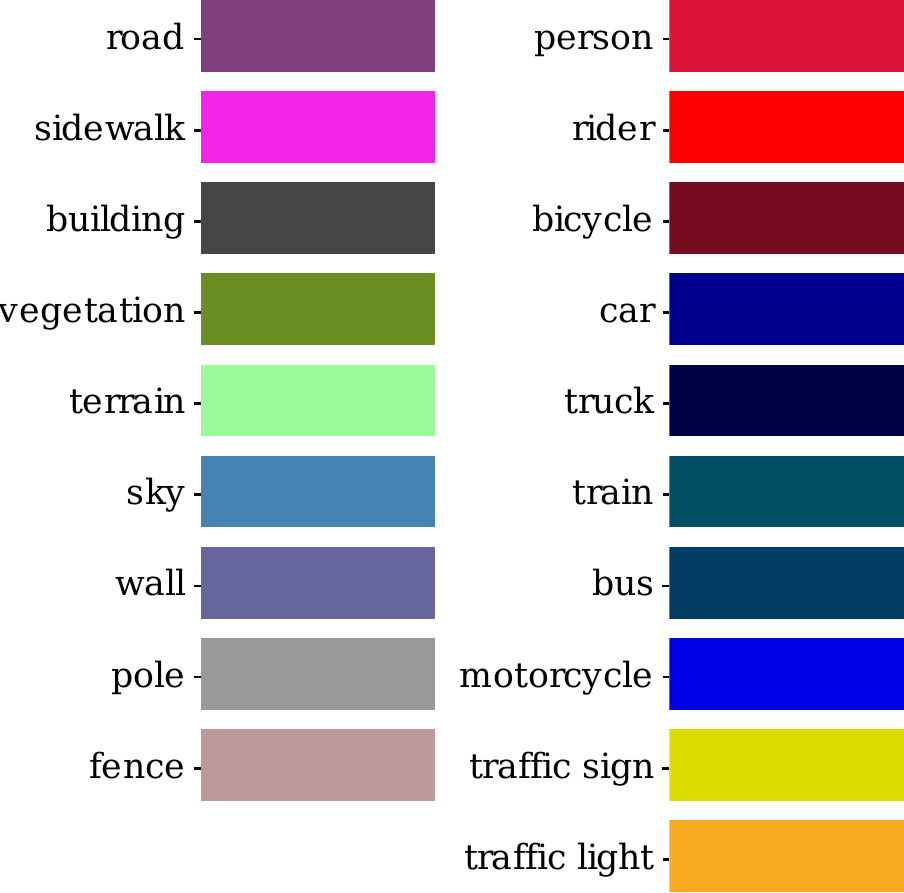}
	\vspace{-3mm}
	\caption{{\bf Semantic map legend.} The colors used in semantic maps throughout this article correspond to the object classes listed above.
	}
	\label{fig:supplement_semantic_colors}
\end{figure}

The legend for the semantic class colors used throughout the article is given in Fig.~\ref{fig:supplement_semantic_colors}.
We present additional examples of the anomaly detection task in Fig.~\ref{fig:supplement_extra_samples}.

The synthetic training process alters only foreground objects. A potential failure mode could therefore be for the network to detect {\it all} foreground objects as anomalies, thus finding not only the true obstacles but also everything else. 
In Fig.~\ref{fig:supplement_laf_foreground}, we show that this does not happen and that objects correctly labeled in the semantic segmentation are not detected as discrepancies.

In Fig.~\ref{fig:supplement_road_anomaly_variation}, we illustrate the fact that, sometimes, objects of known classes differ strongly in appearance from the instances of this class present in the training data, resulting in them being marked as unexpected. 

We present a failure case of our method in Fig.~\ref{fig:supplement_anomaly_failure_case}: 
Anomalies similar to an existing semantic class are sometimes not detected as discrepancies
if the semantic segmentation marks them as this similar class.
For example, an animal is assigned to the {\it person} class and missed by the discrepancy network.
In that case, however, the system as a whole is still aware of the obstacle because of its presence in the semantic map.

\begin{figure*}[t]
	\centering
	\includegraphics[width=\linewidth]{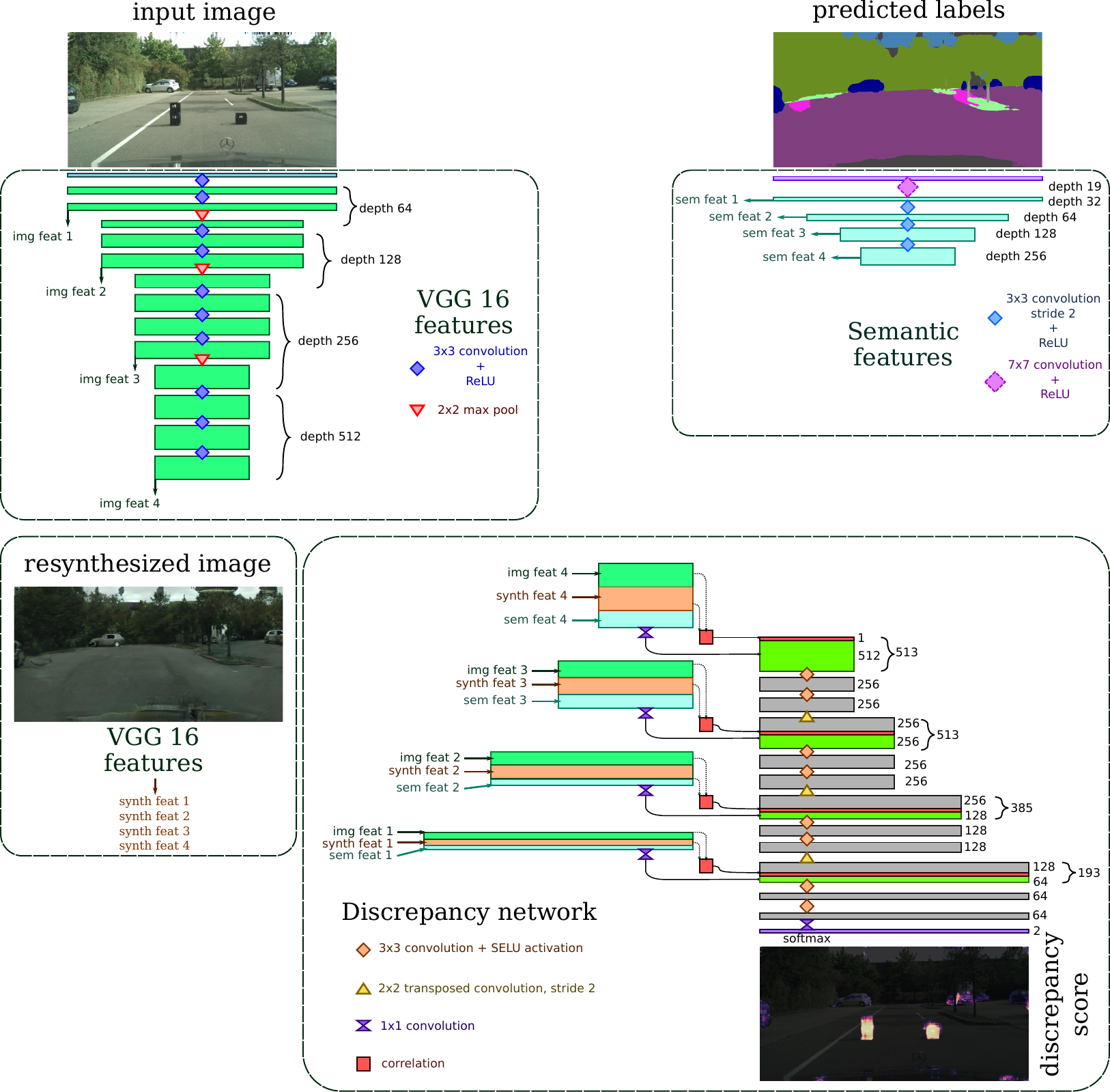}	
	\vspace{-3mm}
	\caption{{\bf Architecture of our discrepancy network.}
	}
	\label{fig:supplement_discrepancy_arch}
\end{figure*}

Our discrepancy network relies on the implementations of {\it PSP Net}~\cite{\PSPnet} and {\it SegNet}~\cite{\SegNet} kindly provided by Zijun Deng.
The detailed architecture of the discrepancy network is shown in Fig.~\ref{fig:supplement_discrepancy_arch}.
We utilize a pre-trained VGG16~\cite{\VGG} to extract features from images
and calculate their pointwise correlation, inspired by the co-segmentation network of~\cite{Li18b}.
The up-convolution part of the network contains SELU activation functions~\cite{\SELU}.
The discrepancy network was trained for 50 epochs using the Cityscapes~\cite{\Cityscapes} training set with synthetically changed labels as described in Section~3.2 of the main paper.
We used the Adam~\cite{\AdamOpt} optimizer with a learning rate of 0.0001 and the per-pixel cross-entropy loss.
We utilized the class weighting scheme introduced in~\cite{Paszke16} to offset the unbalanced numbers of pixels belonging to each class.

\providecommand{\localwidth}{}
\renewcommand{\localwidth}{0.30\linewidth}

\newcommand{\SamplesLafD}[1]{supplement/figures/extra_samples/02_Hanns_Klemm_Str_44__02_Hanns_Klemm_Str_44_000017_000170_#1} %
\newcommand{\SamplesRanoConeE}[1]{supplement/figures/extra_samples/Federation_chantier_aout_2006_-_5_#1} %

\begin{figure*}[t]
	\centering
    \begin{tabular}{ccc}
		\includegraphics[width=\localwidth]{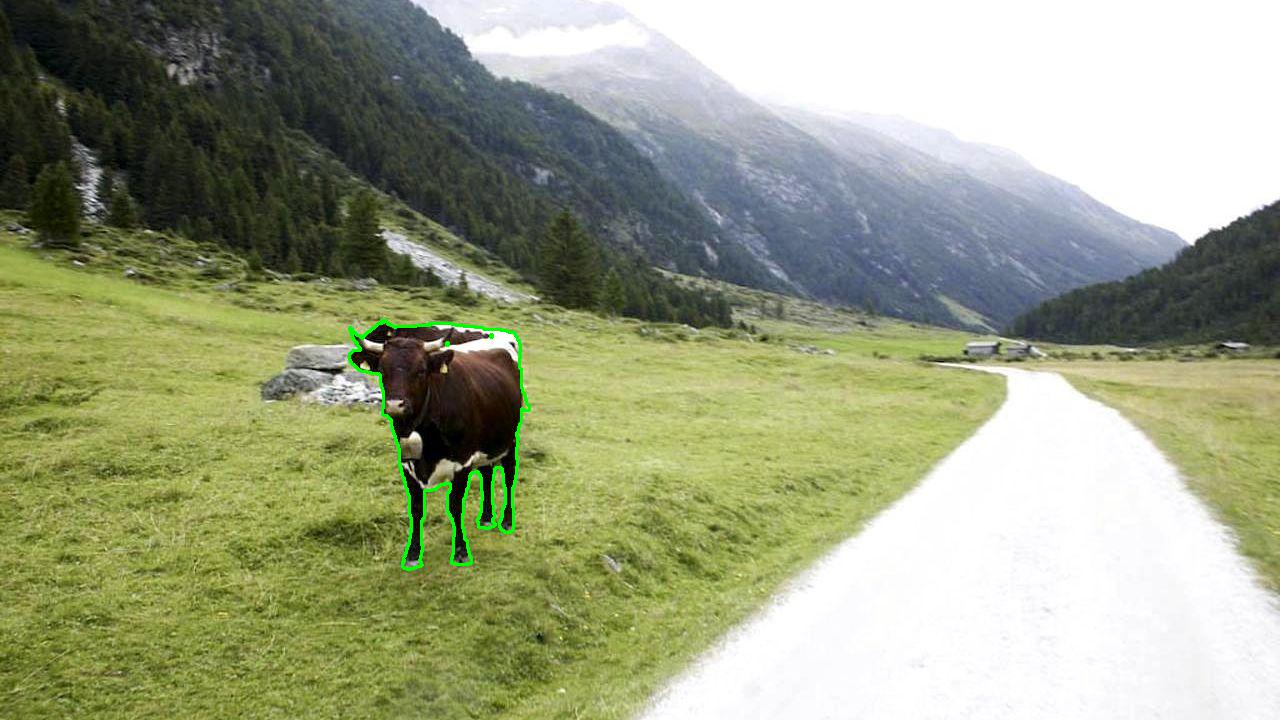}&
		\includegraphics[width=\localwidth]{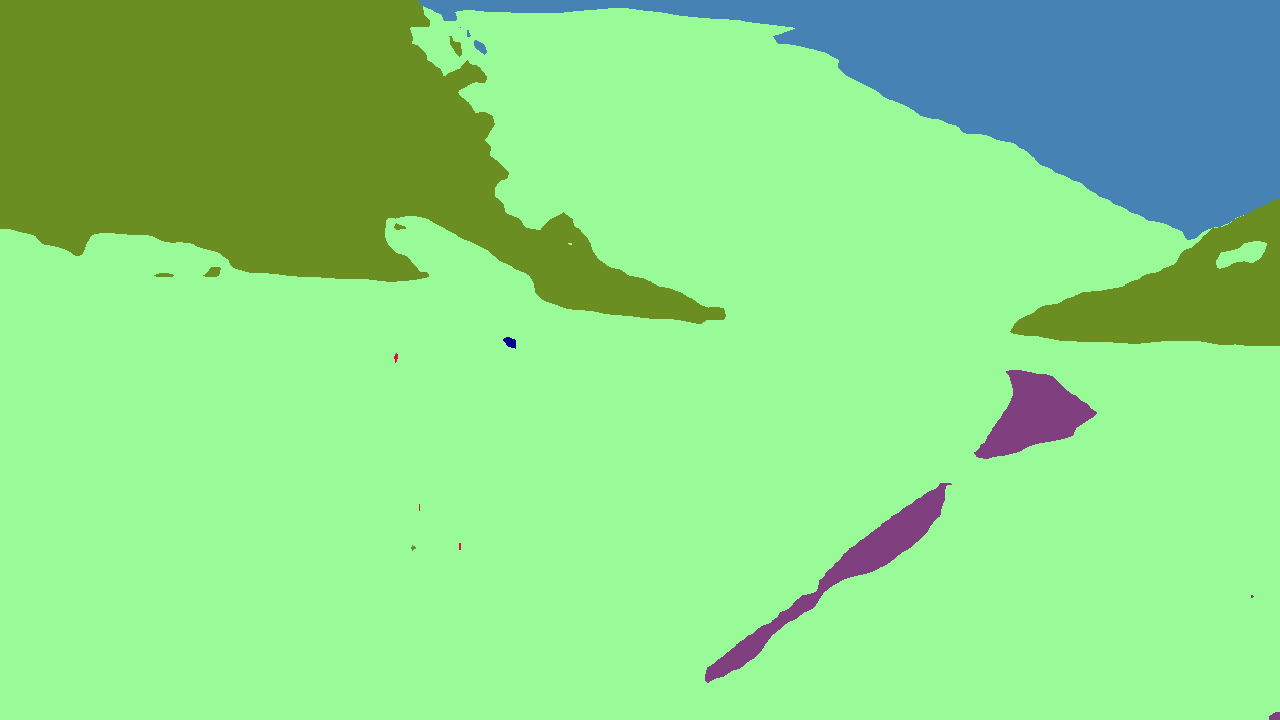}&
		\includegraphics[width=\localwidth]{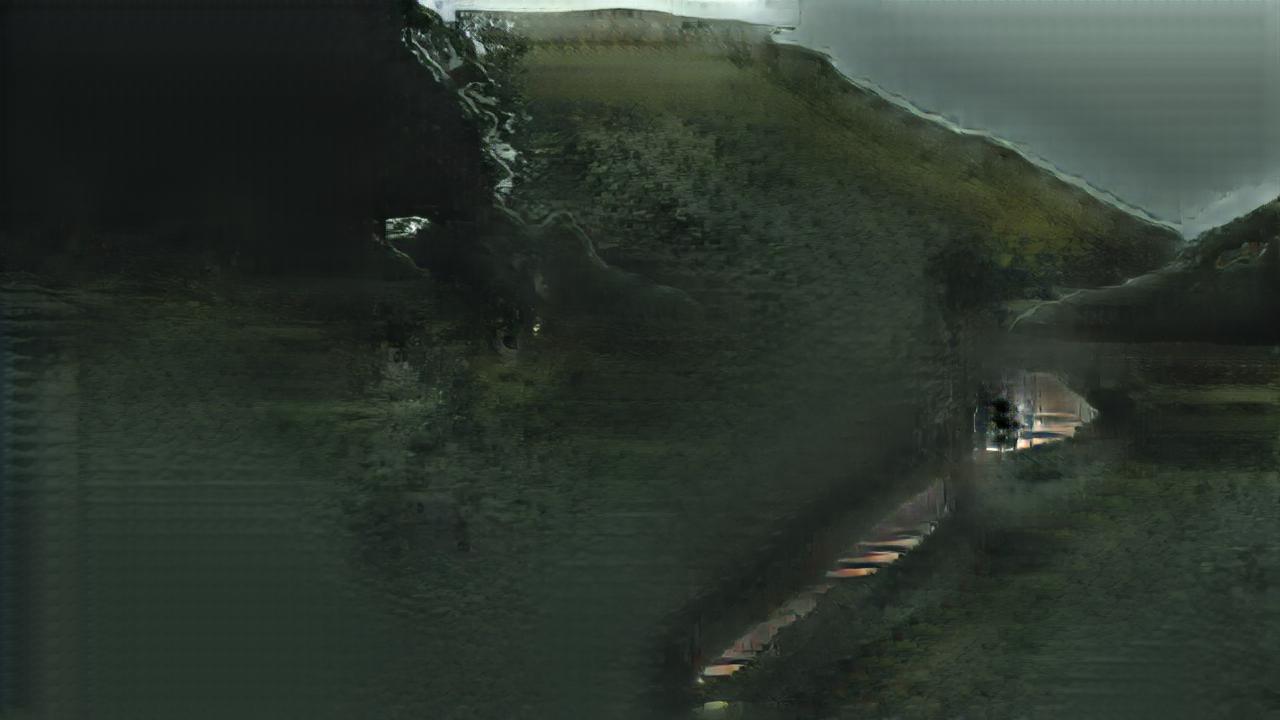}\\[-1mm]
			\small{Input image with anomalies highlighted}&
			\small{Predicted semantic map}&
			\small{Resynthesized image}\\
		\includegraphics[width=\localwidth]{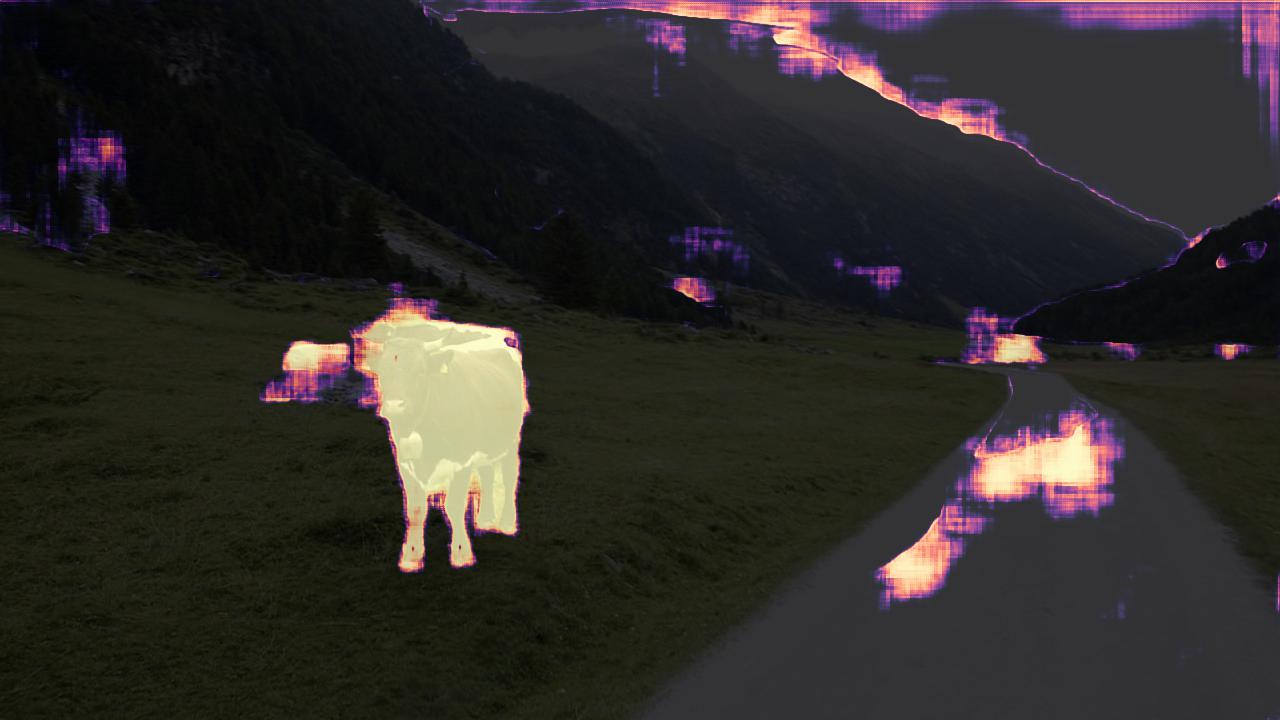}&
		\includegraphics[width=\localwidth]{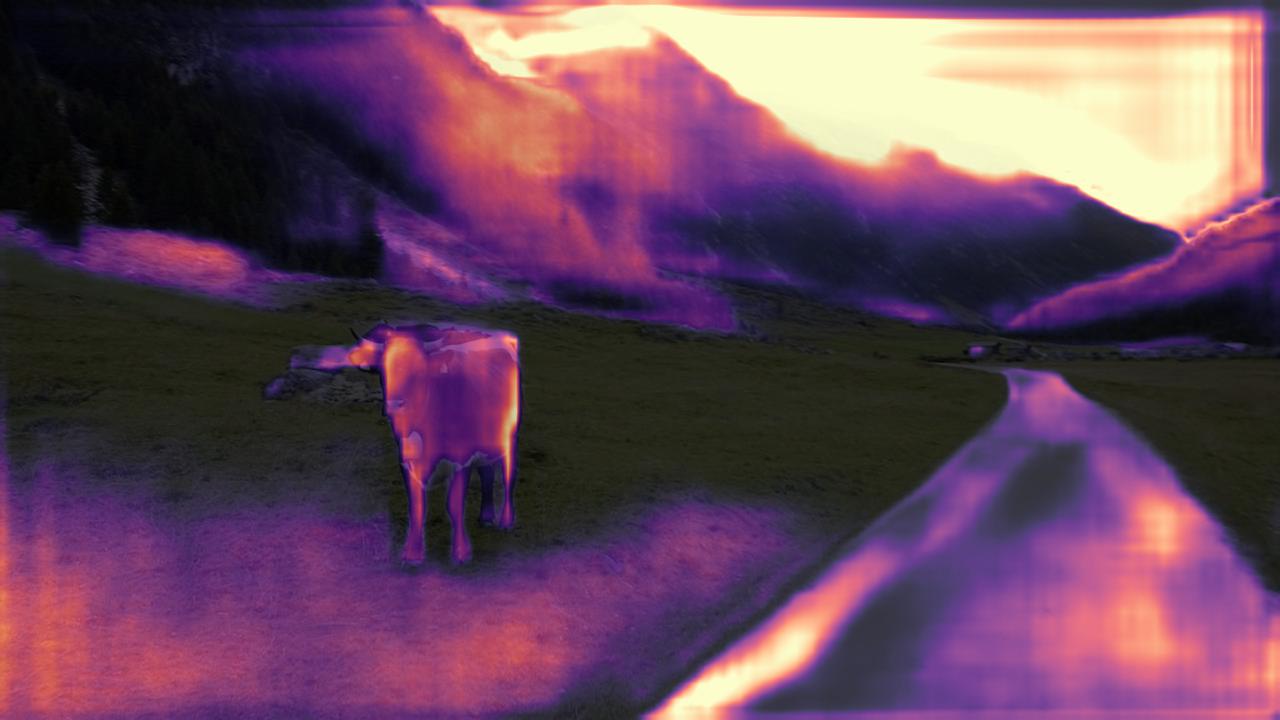}&
		\includegraphics[width=\localwidth]{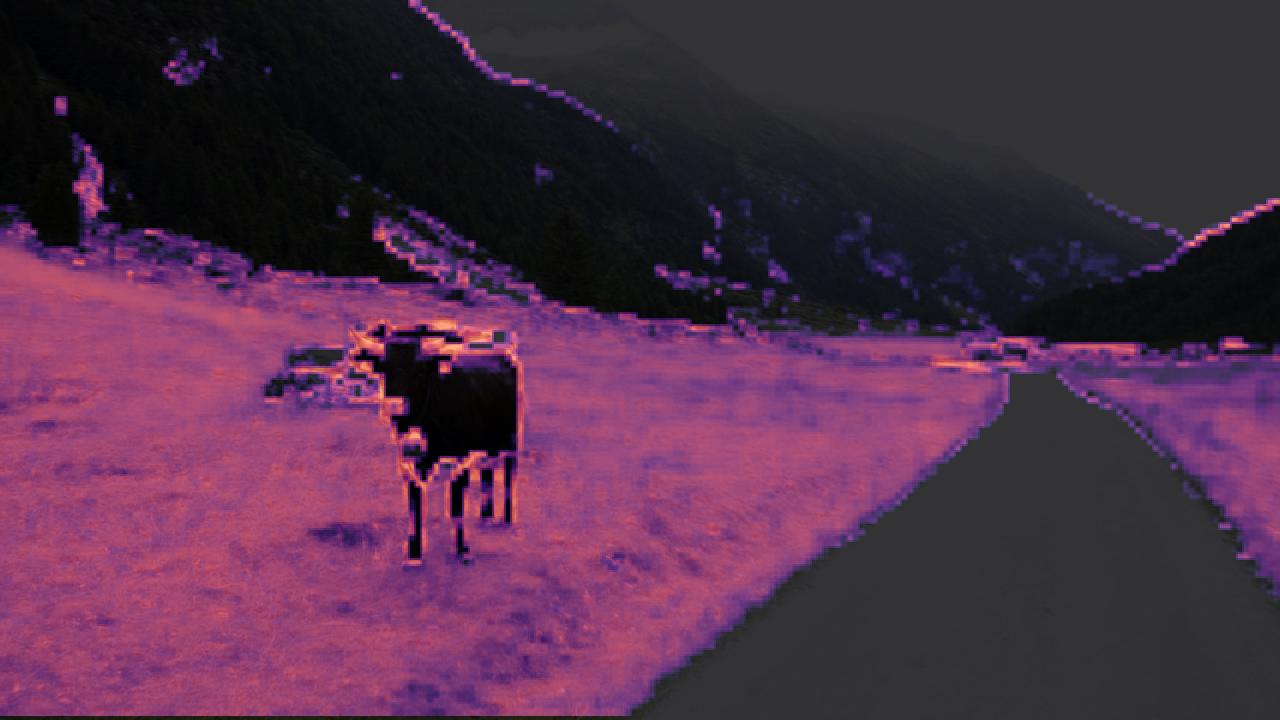}\\[-1mm]
			\small{Anomaly score - \textit{Ours}}&
			\small{Anomaly score - \textit{Uncertainty (Ensemble)}}&
			\small{Anomaly score - \textit{RBM}}\\
		\includegraphics[width=\localwidth]{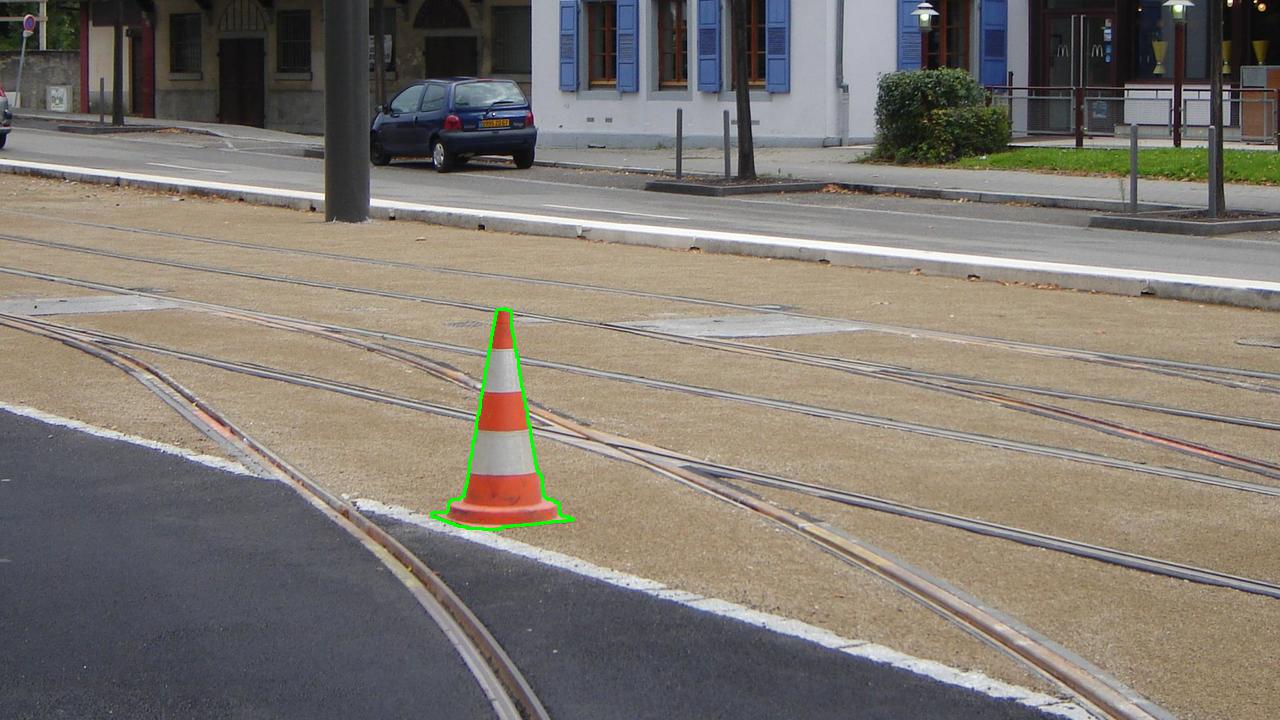}&
		\includegraphics[width=\localwidth]{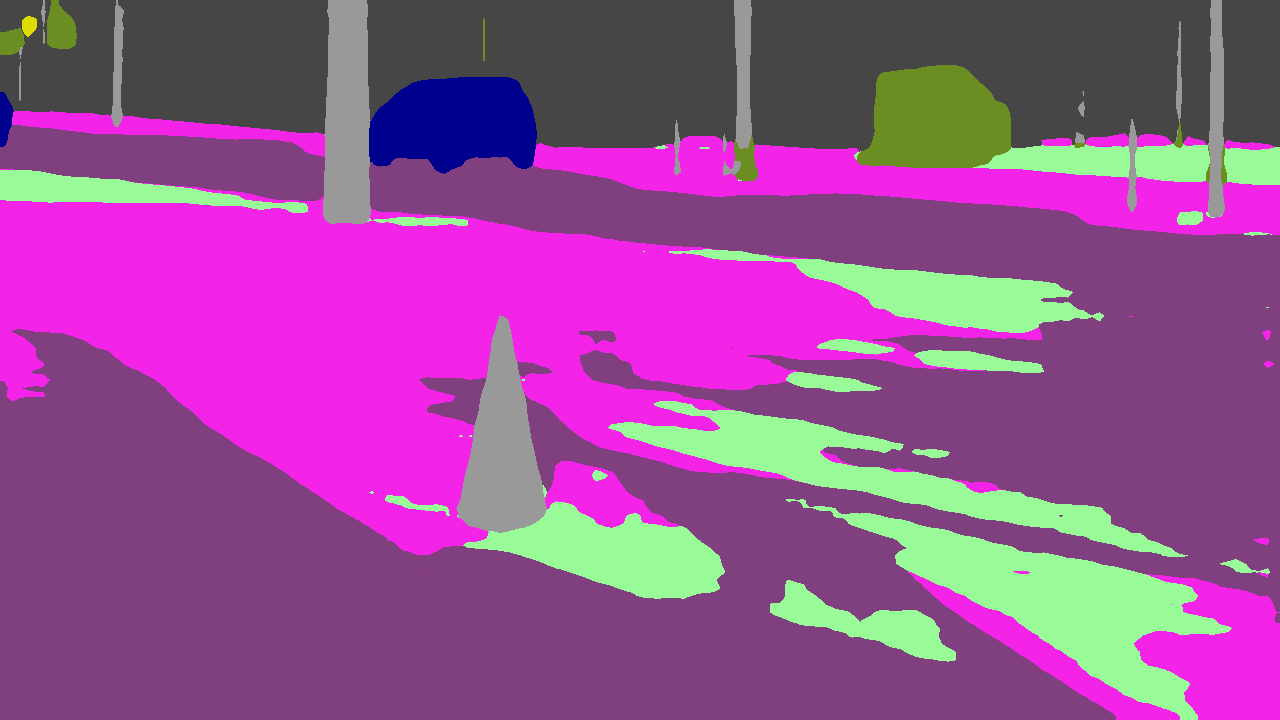}&
		\includegraphics[width=\localwidth]{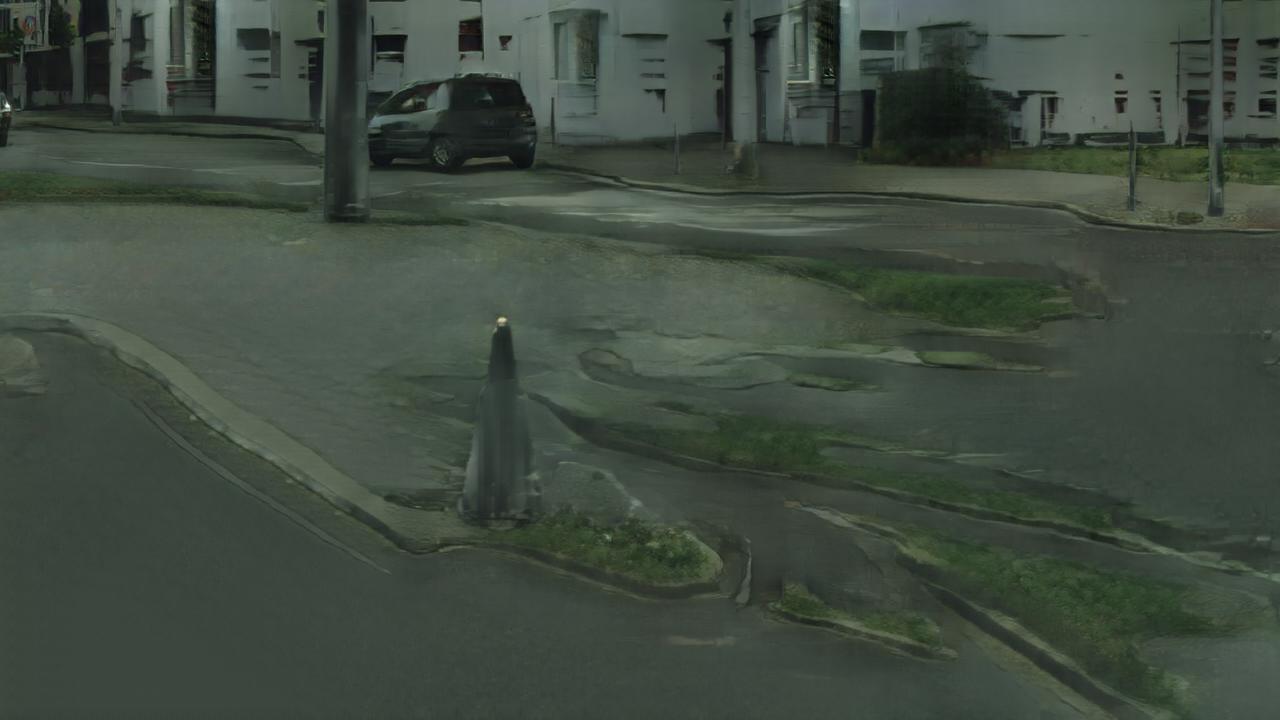}\\[-1mm]
			\small{Input image with anomalies highlighted}&
			\small{Predicted semantic map}&
			\small{Resynthesized image}\\
		\includegraphics[width=\localwidth]{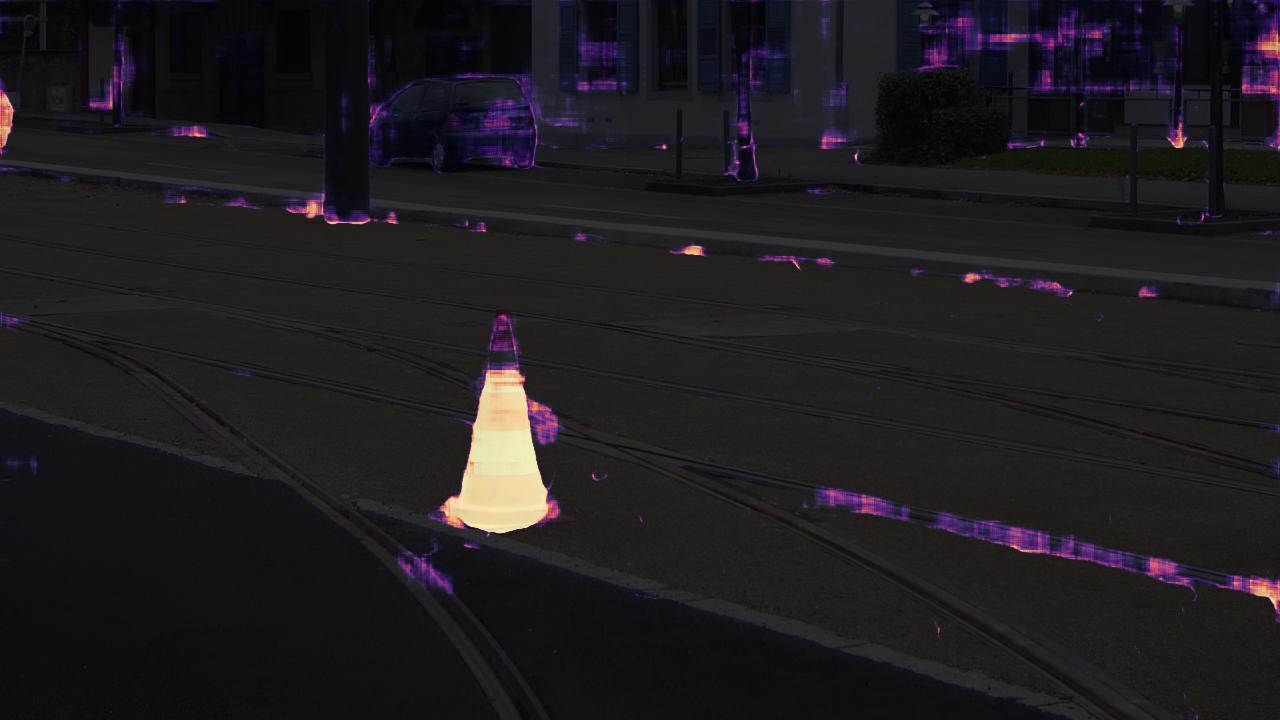}&
		\includegraphics[width=\localwidth]{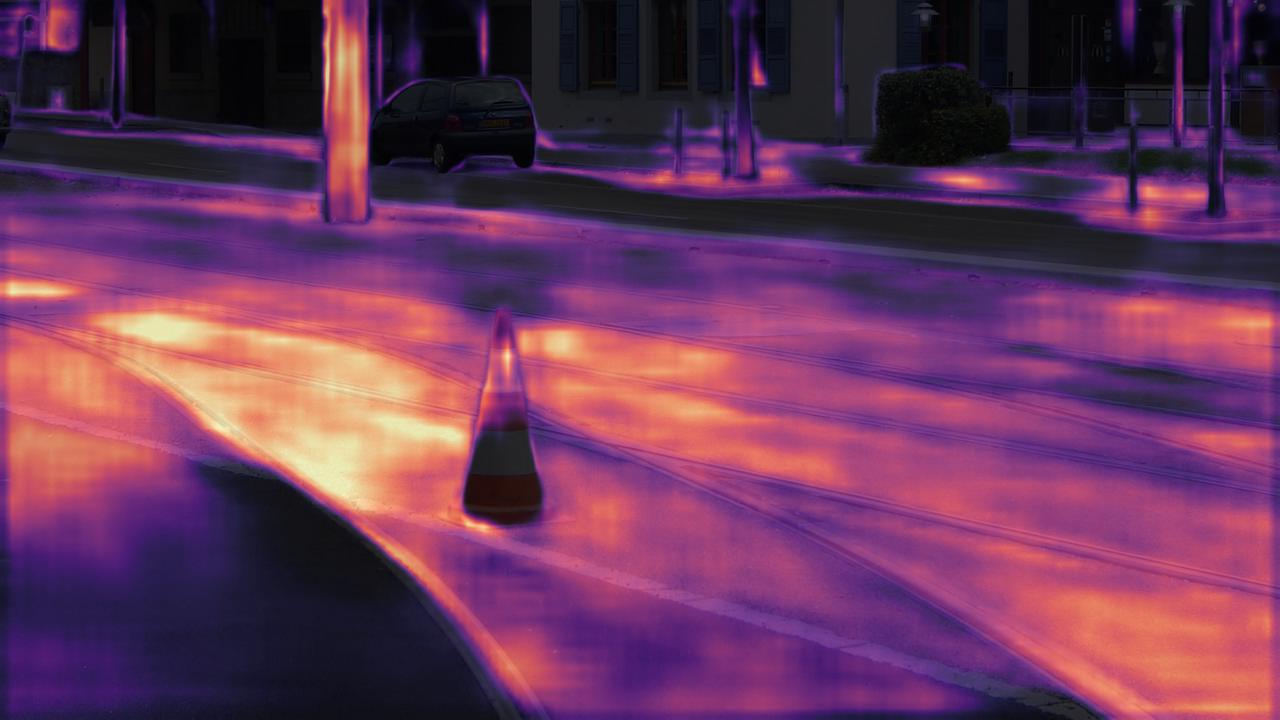}&
		\includegraphics[width=\localwidth]{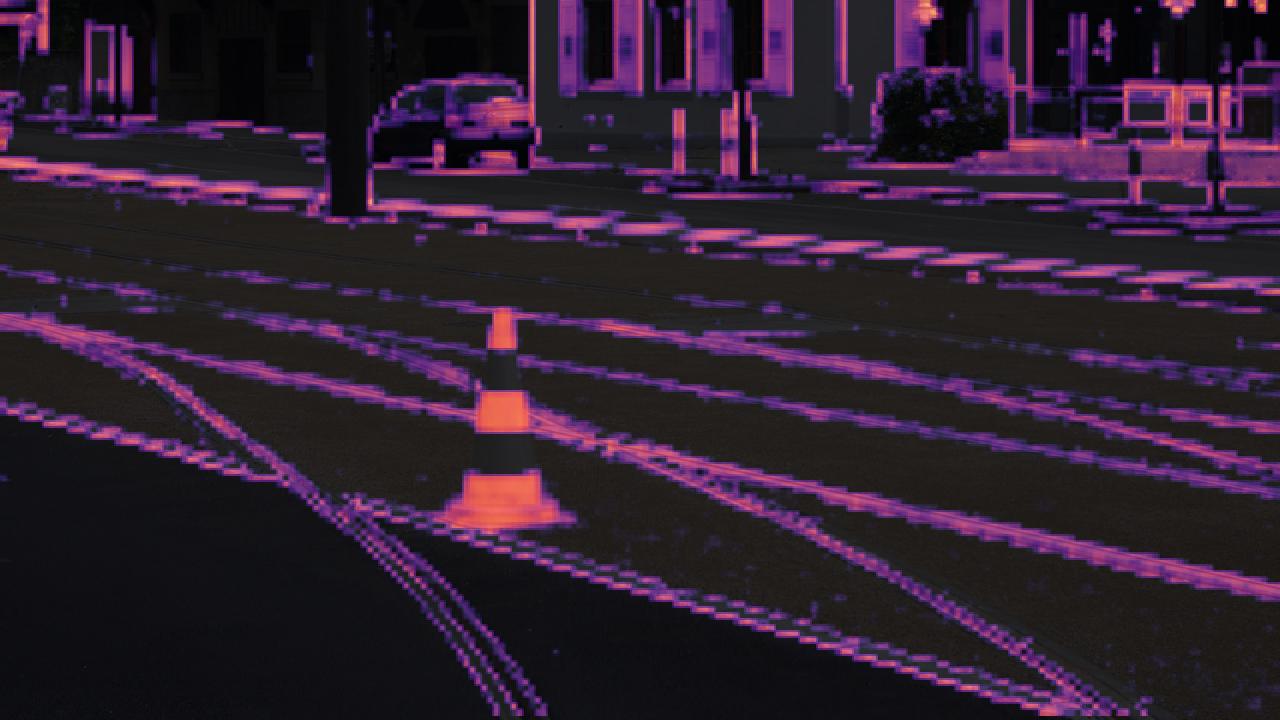}\\[-1mm]
			\small{Anomaly score - \textit{Ours}}&
			\small{Anomaly score - \textit{Uncertainty (Ensemble)}}&
			\small{Anomaly score - \textit{RBM}}\\
		\includegraphics[width=\localwidth]{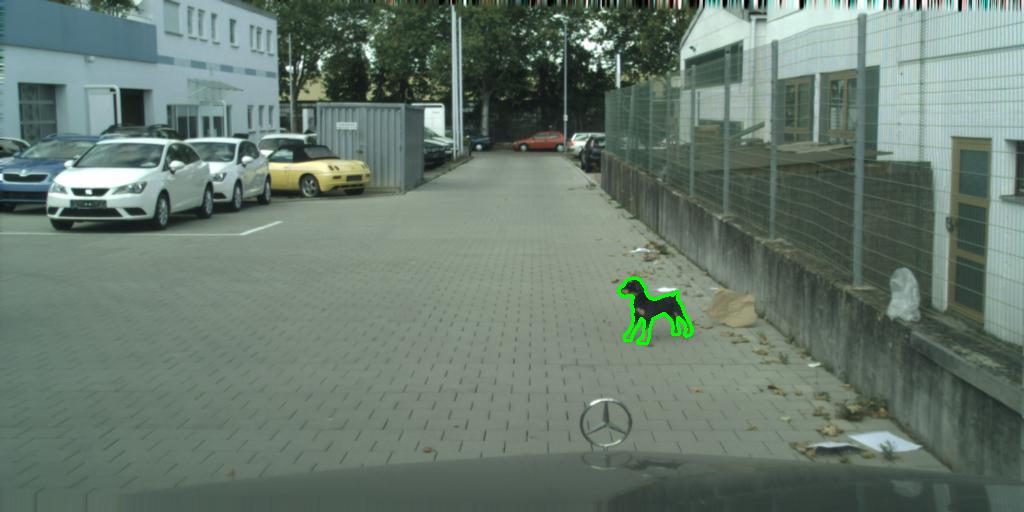}&
		\includegraphics[width=\localwidth]{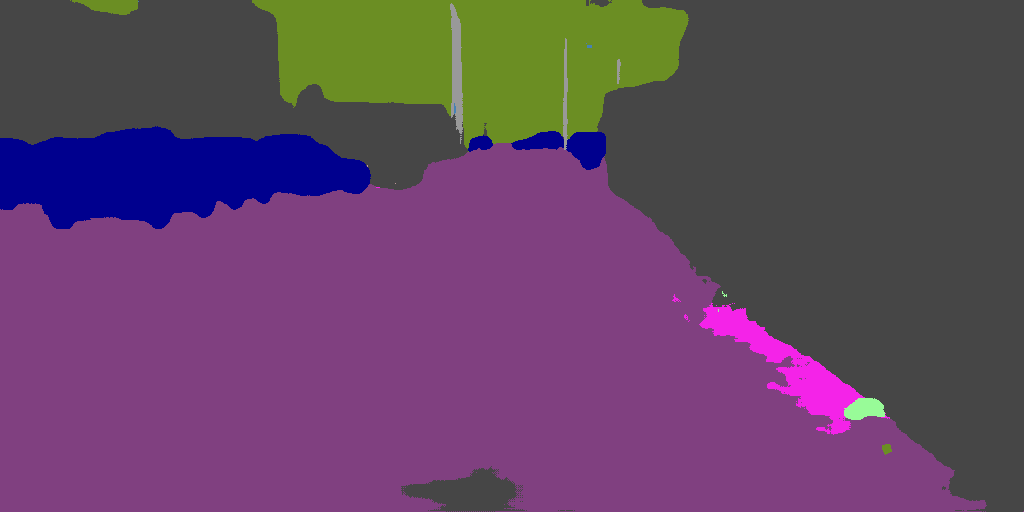}&
		\includegraphics[width=\localwidth]{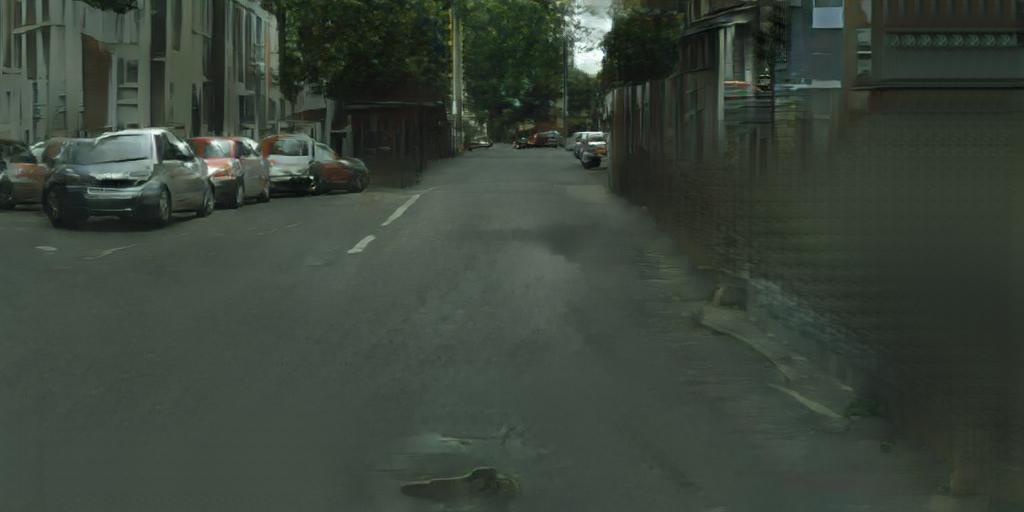}\\[-1mm]
			\small{Input image with anomalies highlighted}&
			\small{Predicted semantic map}&
			\small{Resynthesized image}\\
		\includegraphics[width=\localwidth]{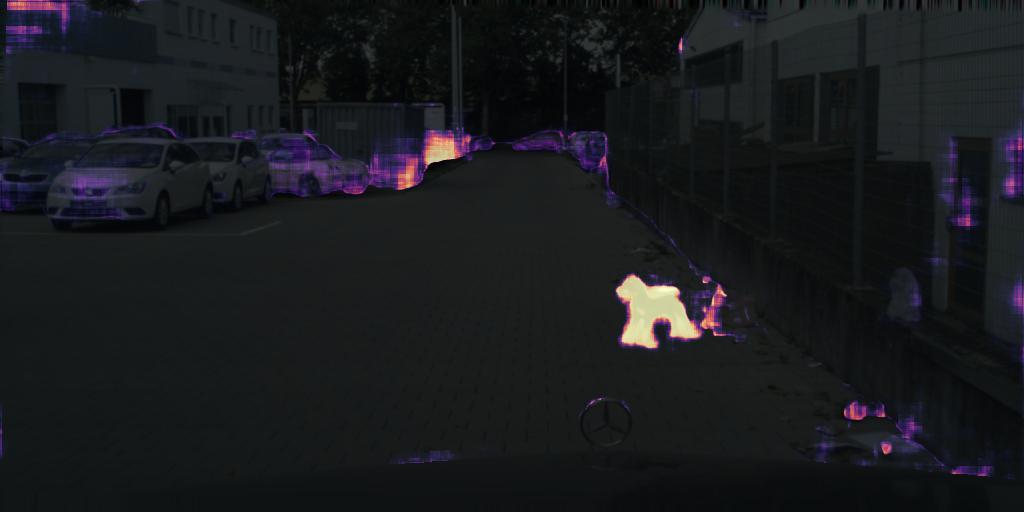}&
		\includegraphics[width=\localwidth]{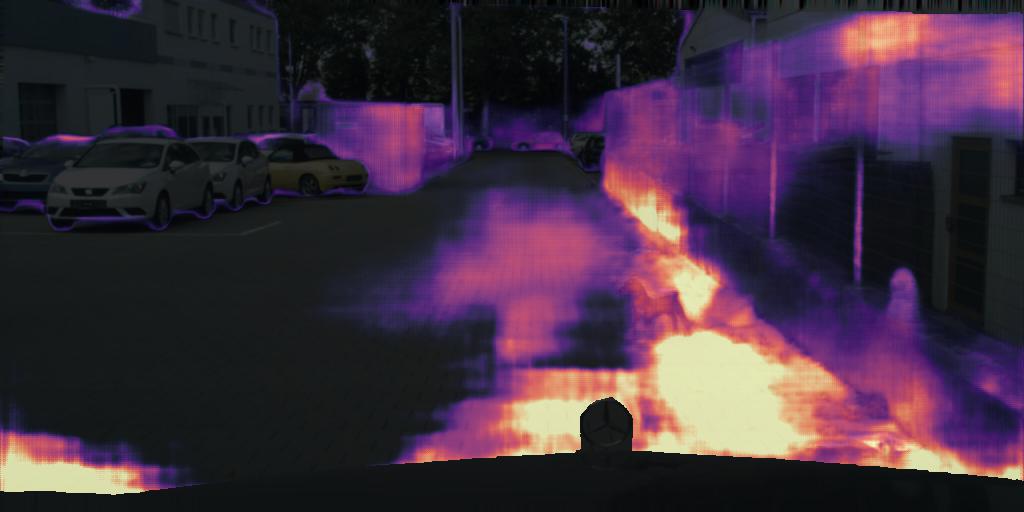}&
		\includegraphics[width=\localwidth]{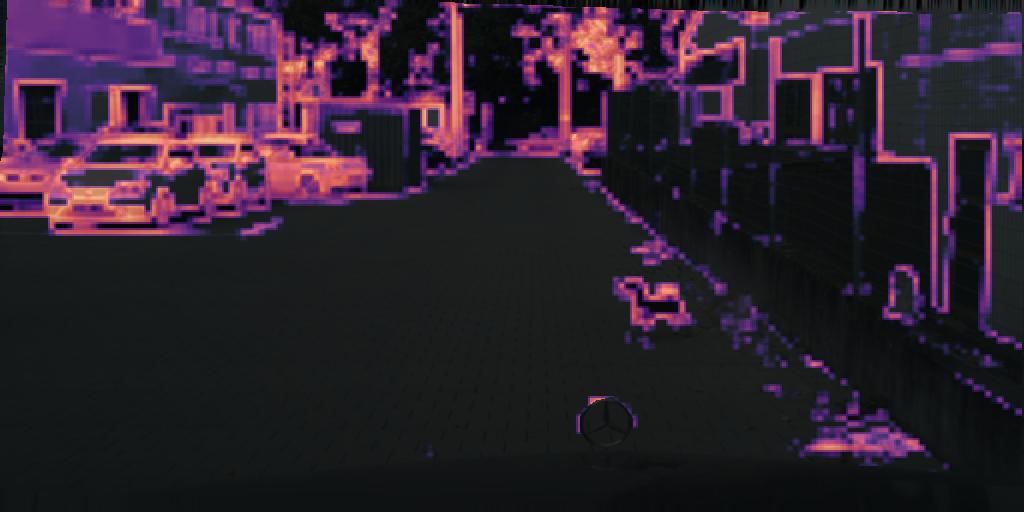}\\[-1mm]
			\small{Anomaly score - \textit{Ours}}&
			\small{Anomaly score - \textit{Uncertainty (Dropout)}}&
			\small{Anomaly score - \textit{RBM}}\\
	\end{tabular}
	\vspace{-3mm}
	\caption{Additional examples of the anomaly detection task
	}
	\label{fig:supplement_extra_samples}
\end{figure*}

\providecommand{\localwidth}{}
\renewcommand{\localwidth}{0.48\linewidth}

\newcommand{\SamplesLafForegroundBayseg}[1]{supplement/figures/laf_foreground_objects/bayseg/15_Rechbergstr_Deckenpfronn__15_Rechbergstr_Deckenpfronn_000007_000360_#1}
\newcommand{\SamplesLafForegroundPSP}[1]{supplement/figures/laf_foreground_objects/psp/15_Rechbergstr_Deckenpfronn__15_Rechbergstr_Deckenpfronn_000007_000360_#1}

\begin{figure*}[t]
	\centering
    \begin{tabular}{cc}
		\includegraphics[width=\localwidth]{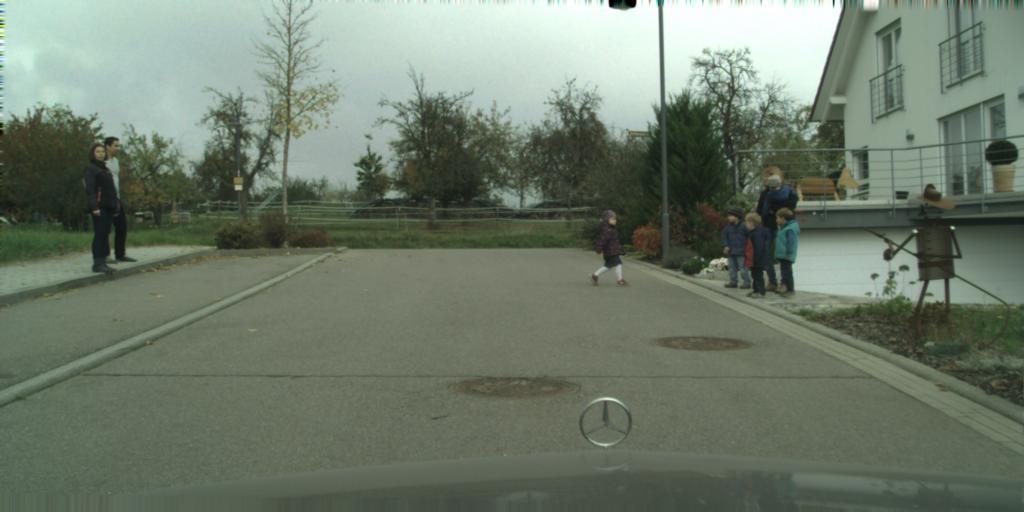}&
		\includegraphics[width=\localwidth]{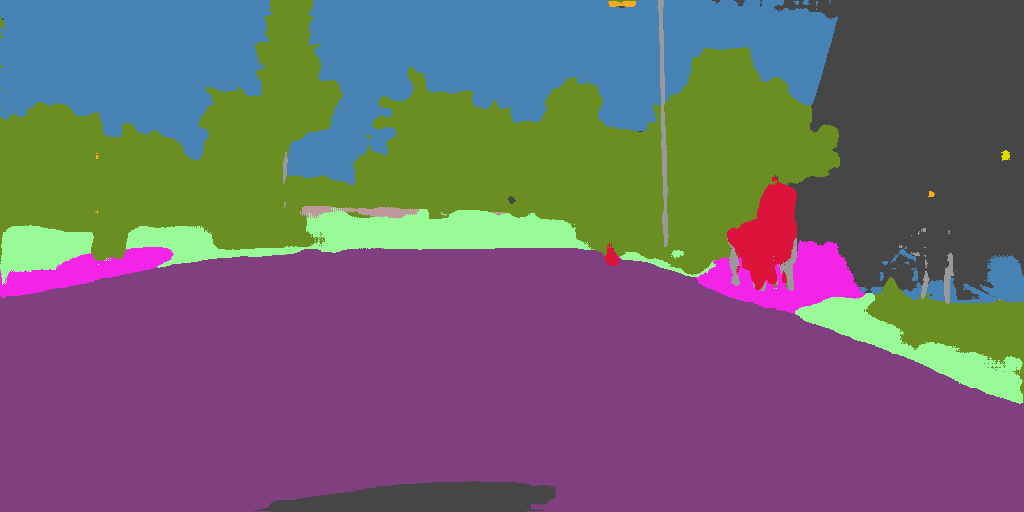}\\[-1mm]
			\small{Input image}&
			\small{Predicted semantic map - Baysesian Seg Net}\\
		\includegraphics[width=\localwidth]{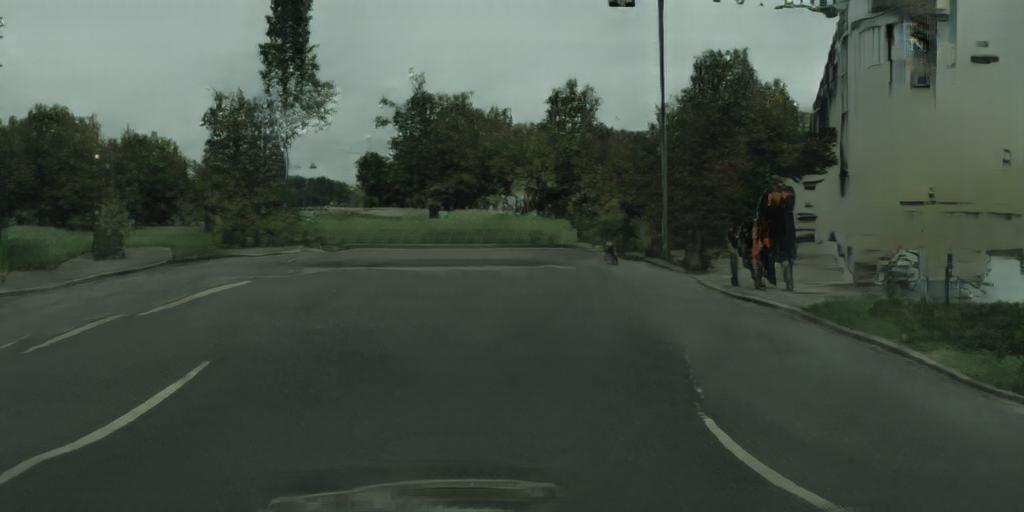}&
		\includegraphics[width=\localwidth]{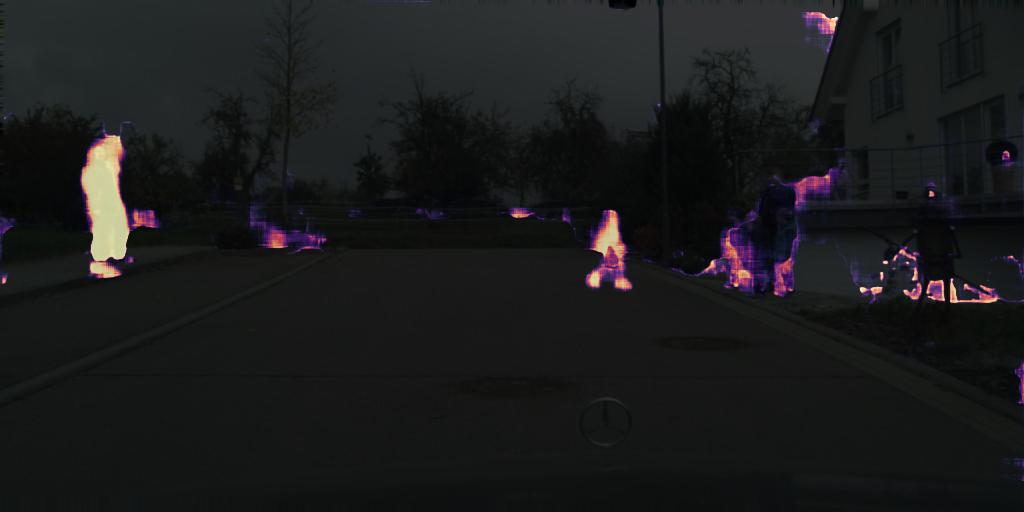}\\[-1mm]
			\small{Resynthesized image (labels from Baysesian Seg Net)}&
			\small{Anomaly score - \textit{Ours}}\\
		\includegraphics[width=\localwidth]{supplement/figures/laf_foreground_objects/15_Rechbergstr_Deckenpfronn__15_Rechbergstr_Deckenpfronn_000007_000360_image.jpg}&
		\includegraphics[width=\localwidth]{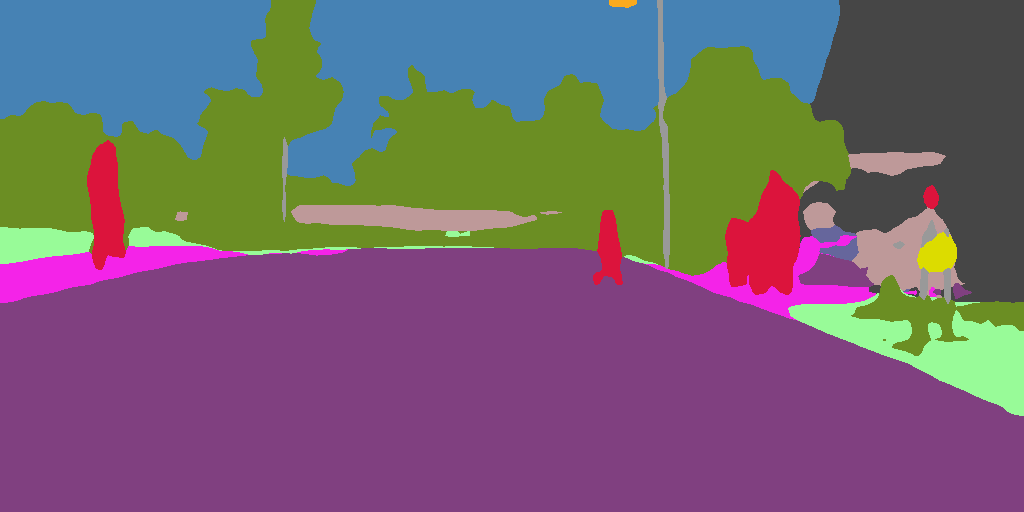}\\[-1mm]
			\small{Input image}&
			\small{Predicted semantic map - PSP Net}\\
		\includegraphics[width=\localwidth]{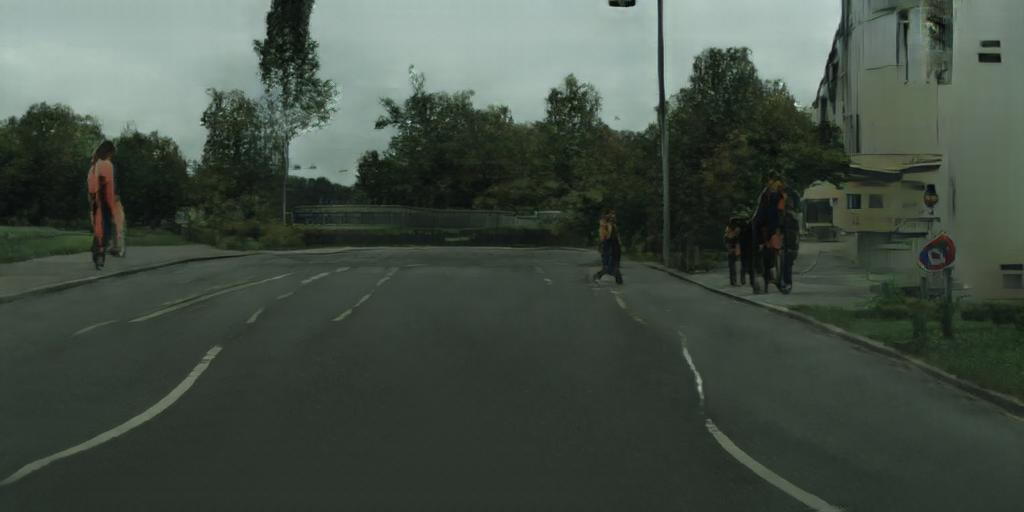}&
		\includegraphics[width=\localwidth]{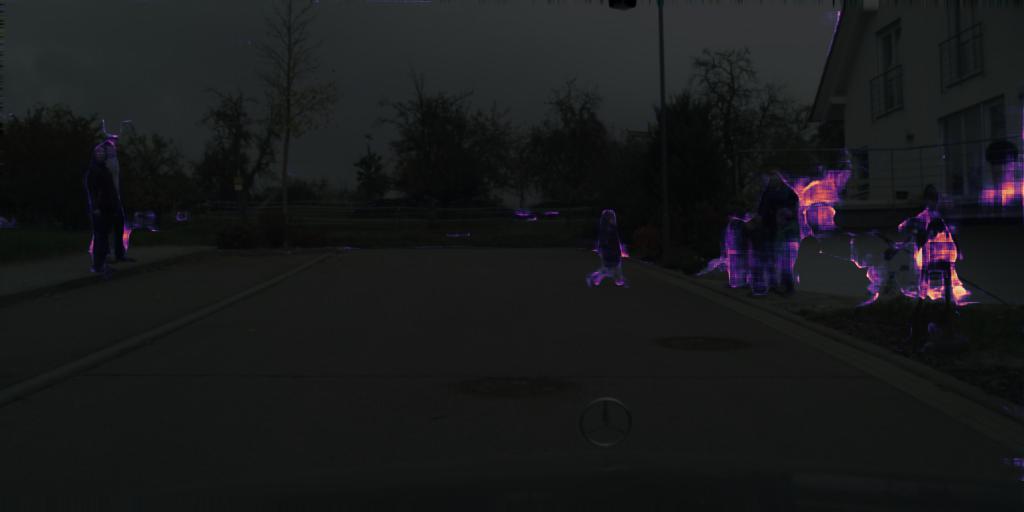}\\[-1mm]
			\small{Resynthesized image (labels from PSP Net)}&
			\small{Anomaly score - \textit{Ours}}\\
	\end{tabular}
	\vspace{-3mm}
	\caption{%
		The synthetic training process alters only foreground objects, but that does not mean our discrepancy network
		learns to blindly mark all such objects.
		In the top row, we show an example where the {\it Bayesian SegNet} failed to correctly label some of the people present, and
		this discrepancy is detected by our network.
		However, our detector reports no discrepancy when the {\it PSP Net} correctly labels the people in the image (third row).
	}
	\label{fig:supplement_laf_foreground}
\end{figure*}

\providecommand{\localwidth}{}
\renewcommand{\localwidth}{0.30\linewidth}

\newcommand{\SamplesZebras}[1]{supplement/figures/road_anomaly_object_variation/Zebra_Crossing_Abbey_Road_Style__63894353__#1} %

\begin{figure*}[t]
	\centering
	\begin{tabular}{ccc}
		\includegraphics[width=\localwidth]{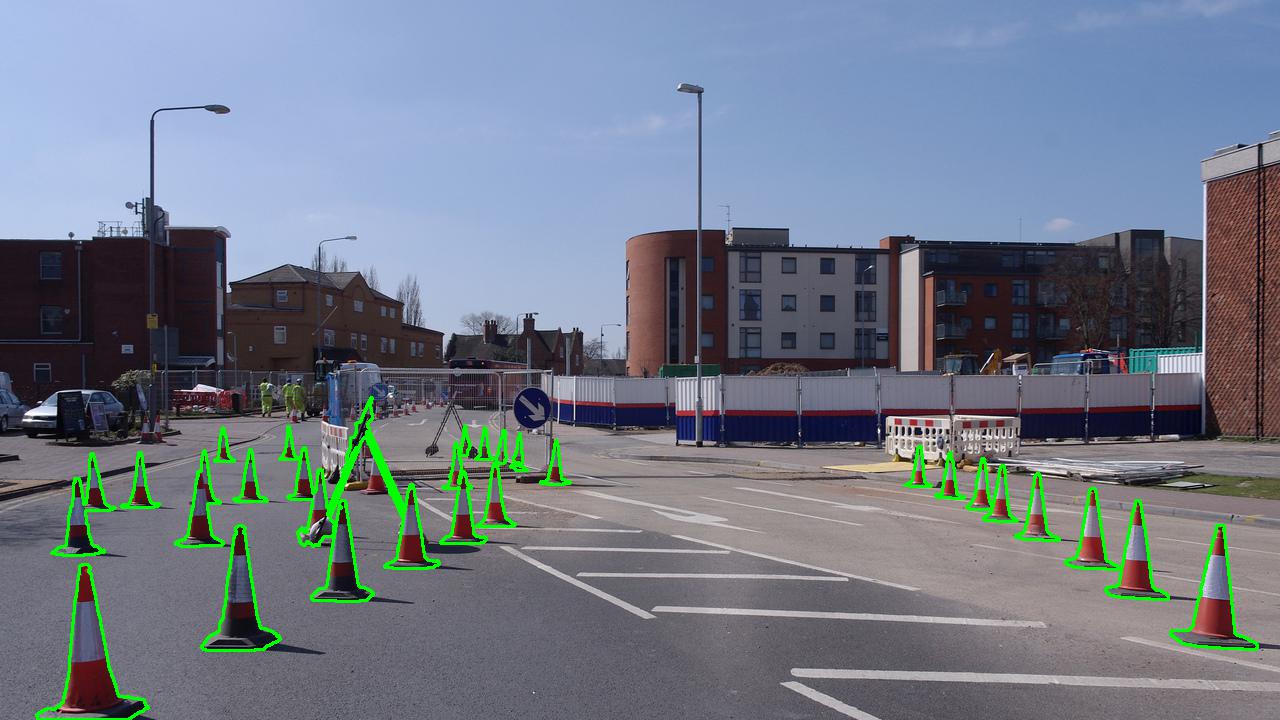}&
		\includegraphics[width=\localwidth]{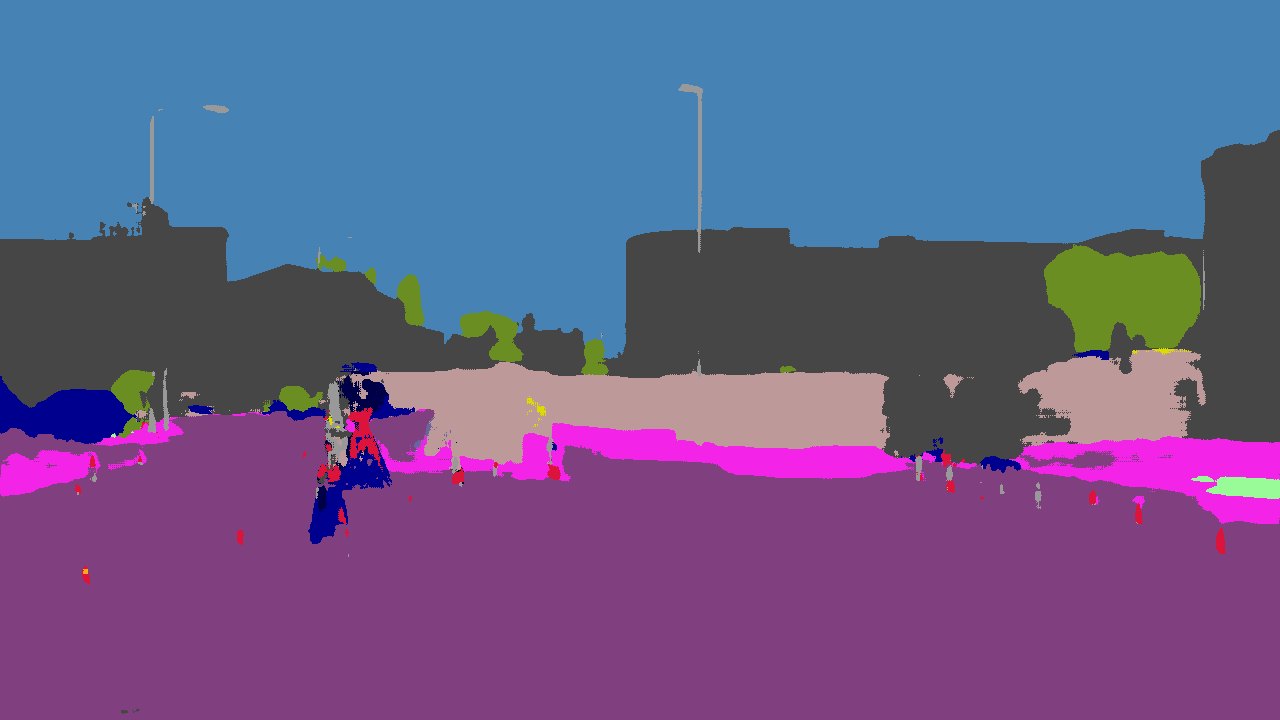}&
		\includegraphics[width=\localwidth]{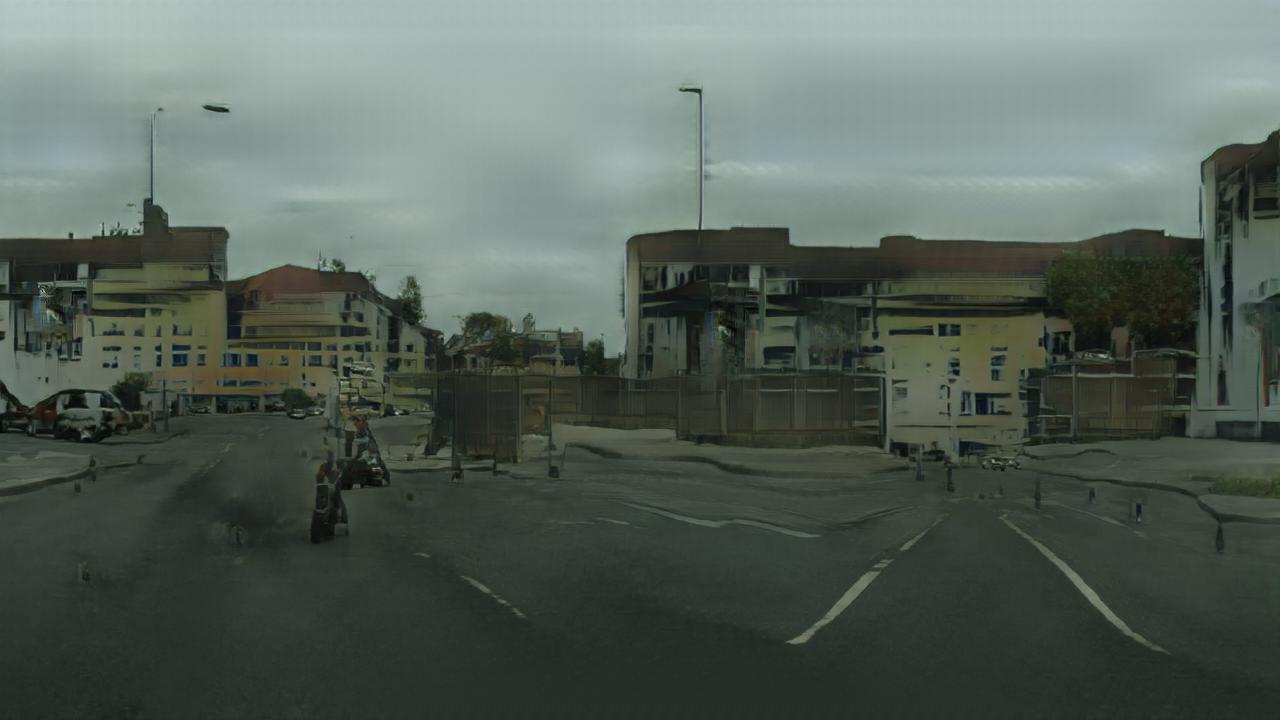}\\[-1mm]
			\small{Input image with anomalies highlighted}&
			\small{Predicted semantic map}&
			\small{Resynthesized image}\\
		\includegraphics[width=\localwidth]{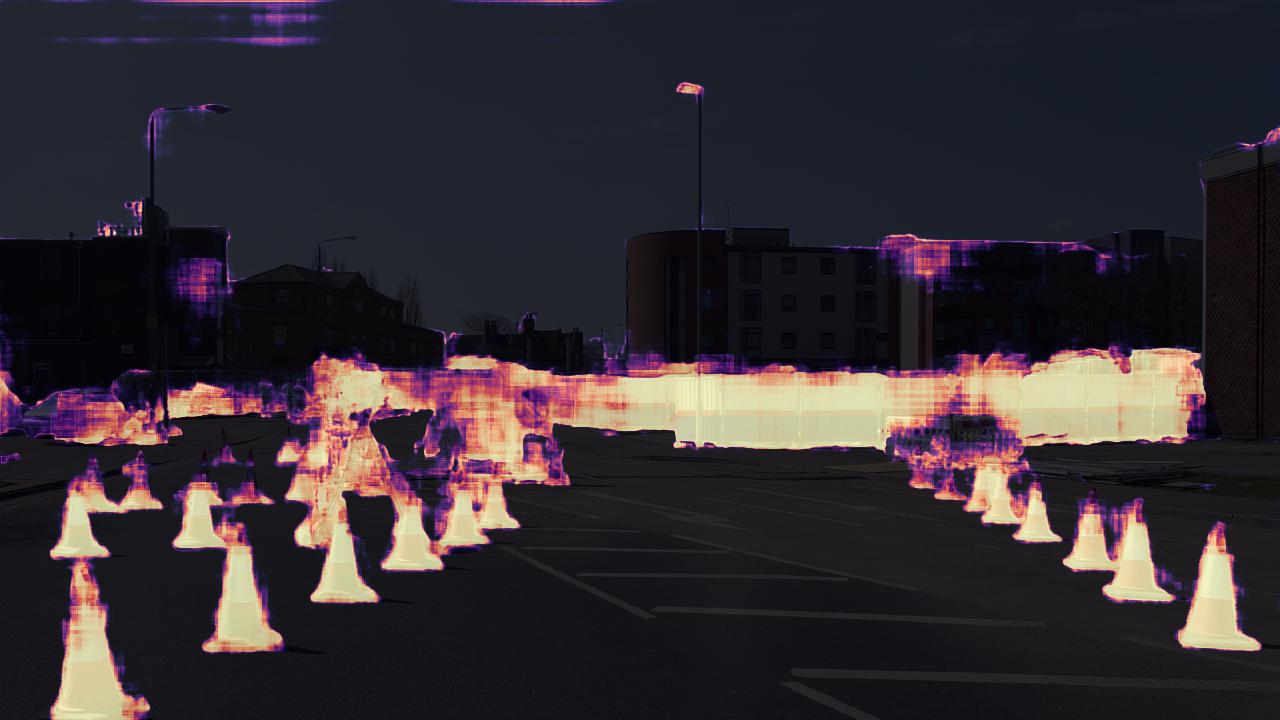}&
		\includegraphics[width=\localwidth]{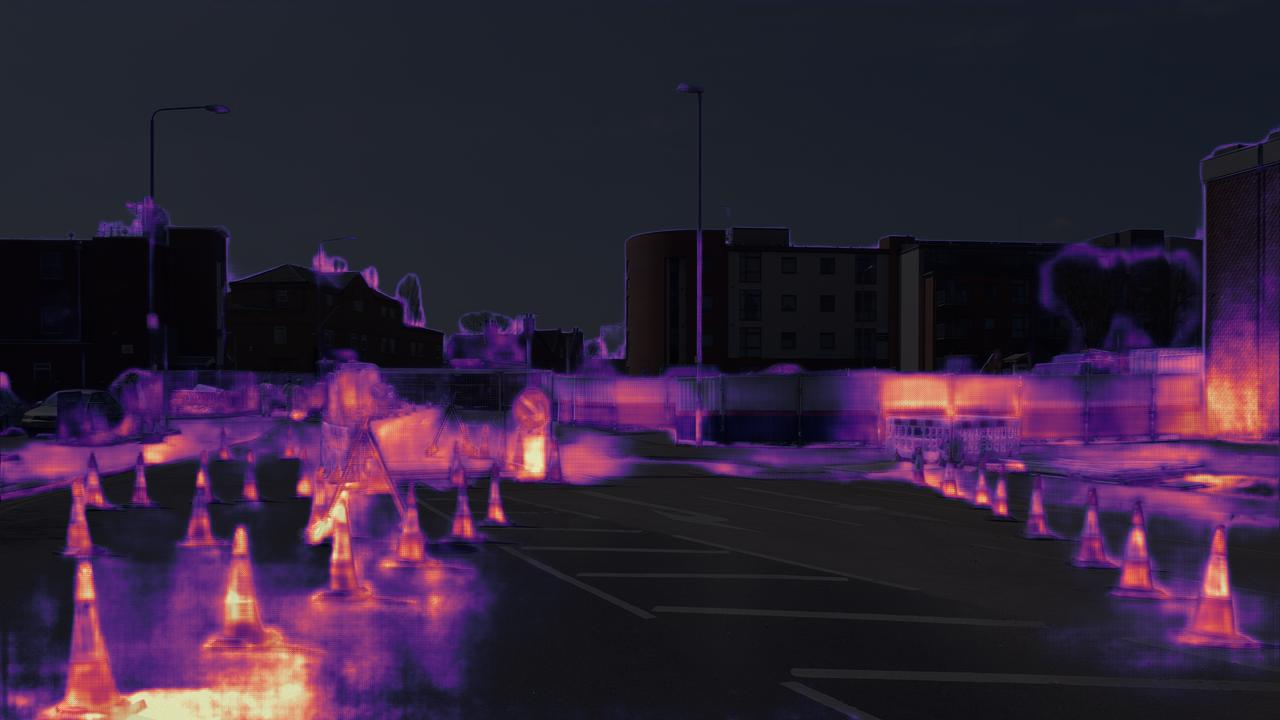}&
		\includegraphics[width=\localwidth]{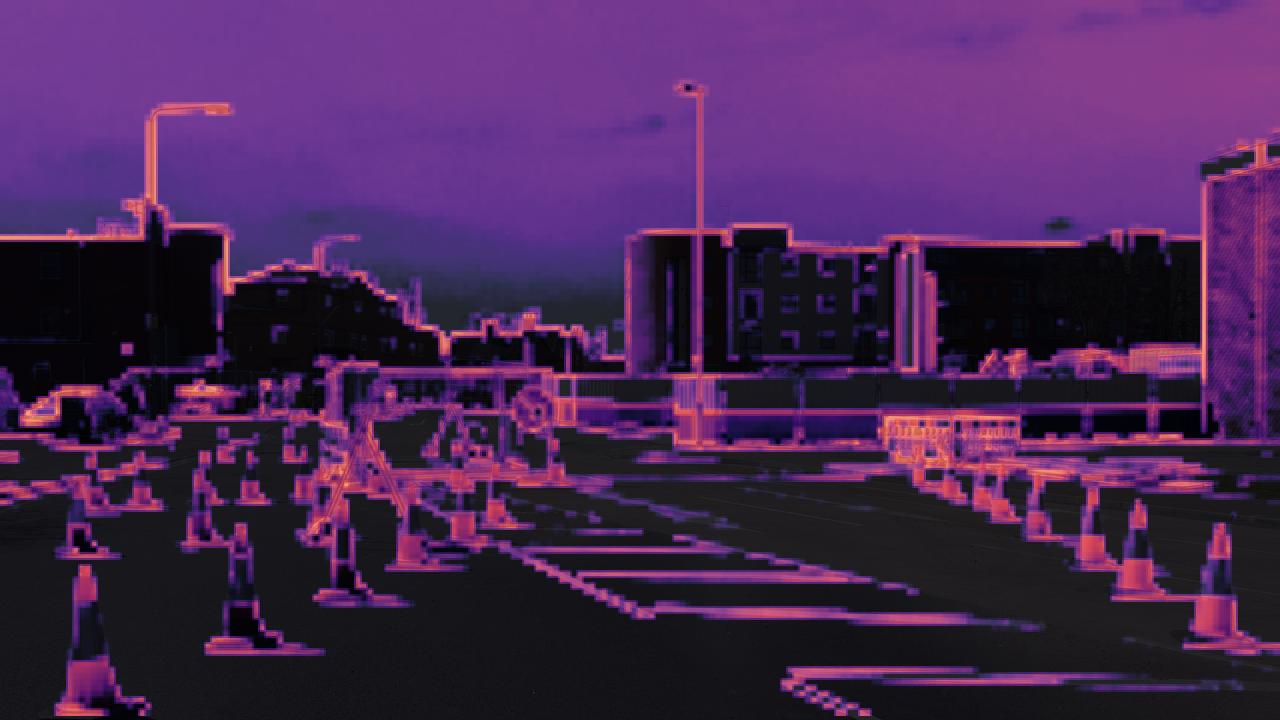}\\[-1mm]
			\small{Anomaly score - \textit{Ours}}&
			\small{Anomaly score - \textit{Uncertainty (Dropout)}}&
			\small{Anomaly score - \textit{RBM}}\\
		\includegraphics[width=\localwidth]{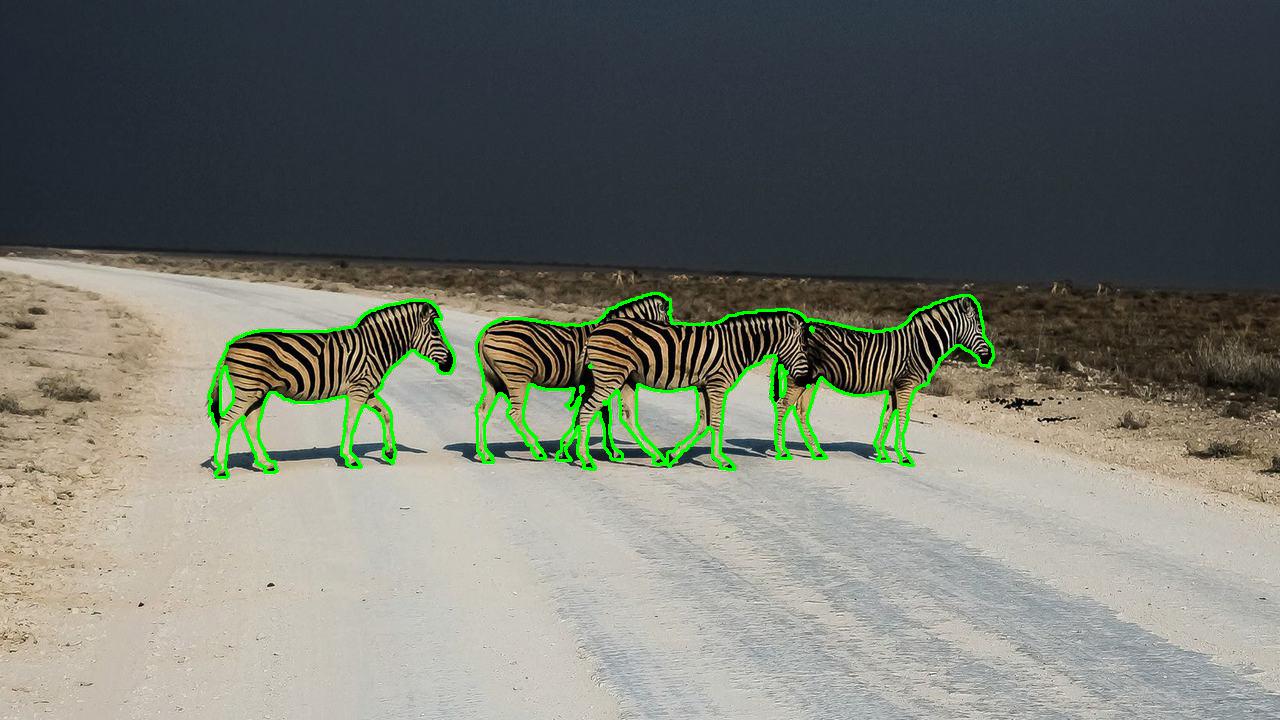}&
		\includegraphics[width=\localwidth]{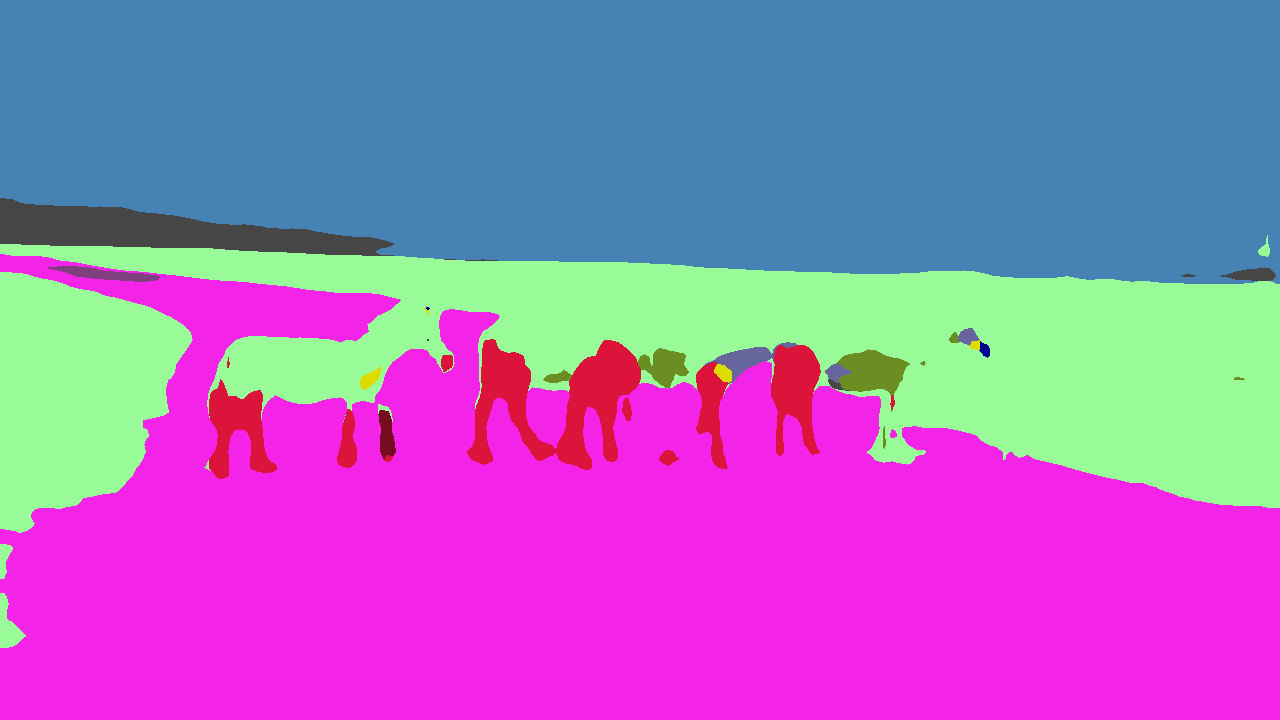}&
		\includegraphics[width=\localwidth]{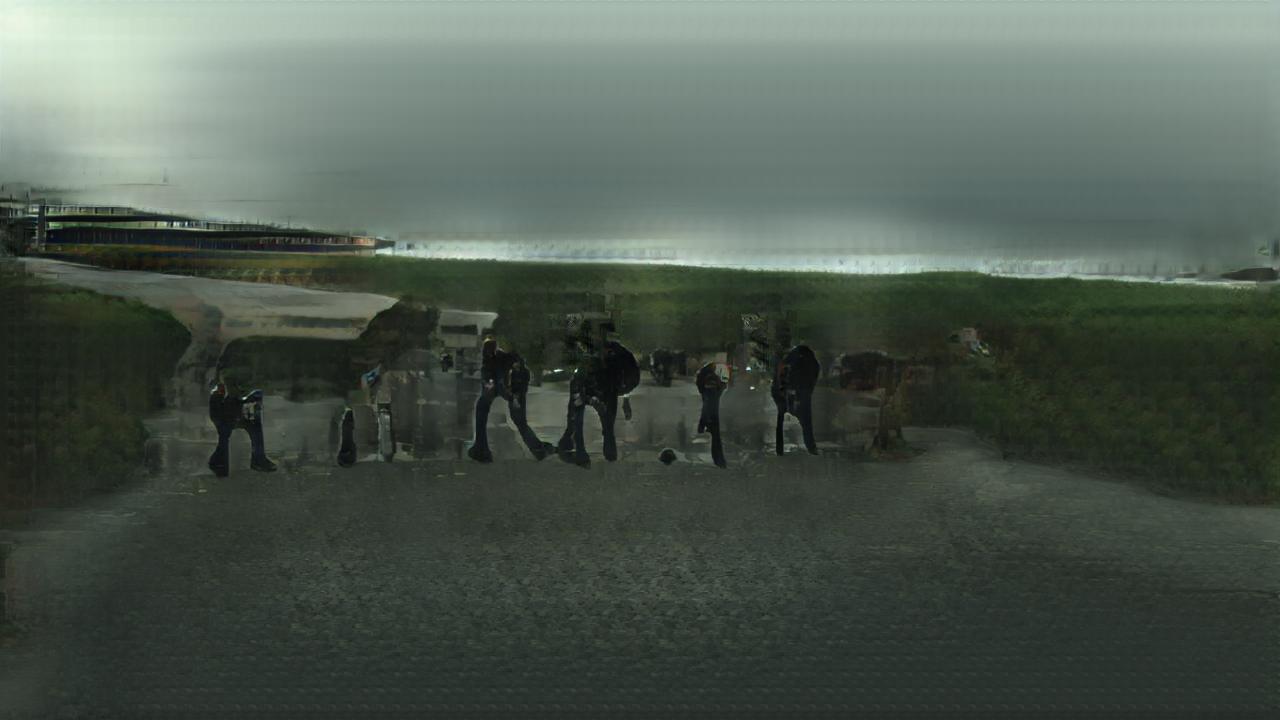}\\[-1mm]
			\small{Input image with anomalies highlighted}&
			\small{Predicted semantic map}&
			\small{Resynthesized image}\\
		\includegraphics[width=\localwidth]{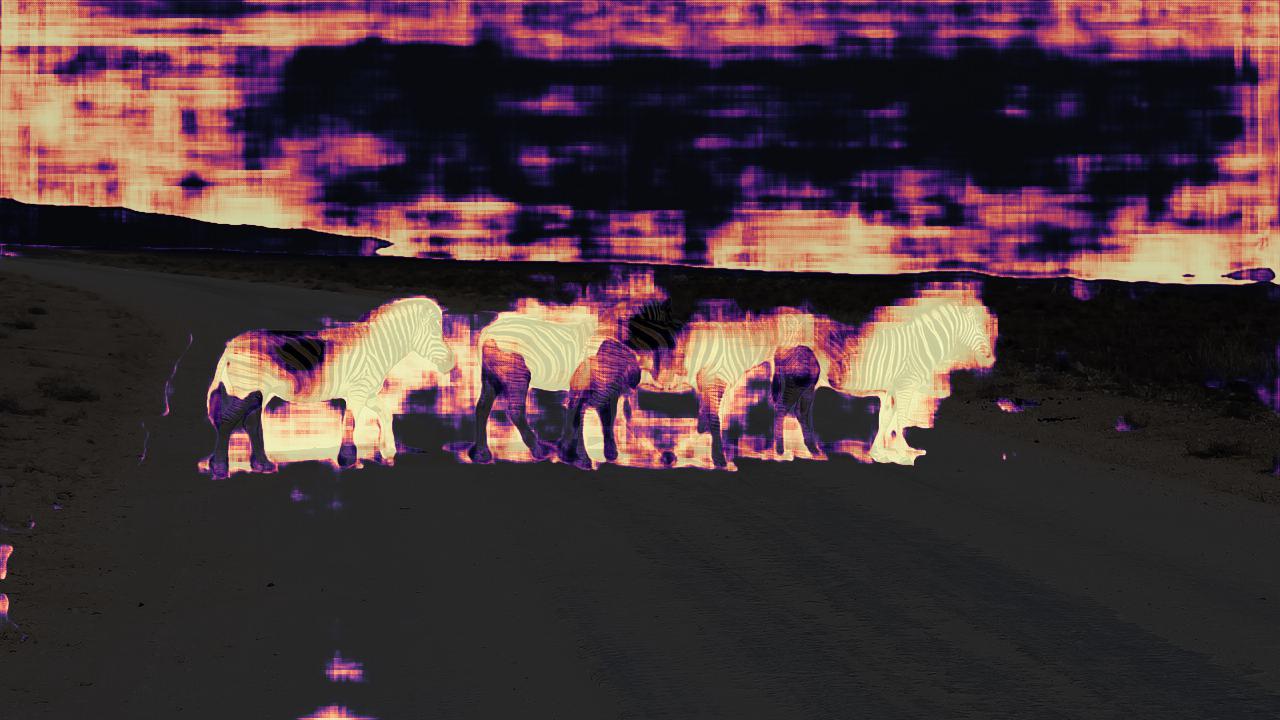}&
		\includegraphics[width=\localwidth]{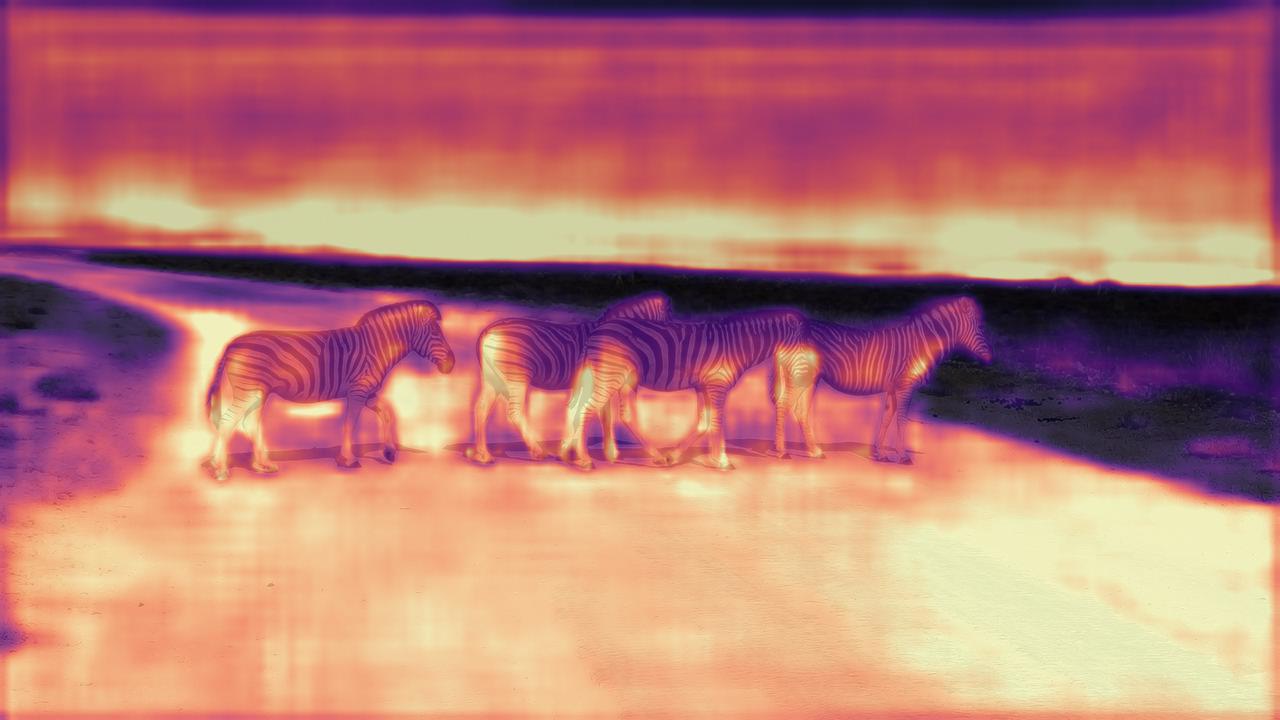}&
		\includegraphics[width=\localwidth]{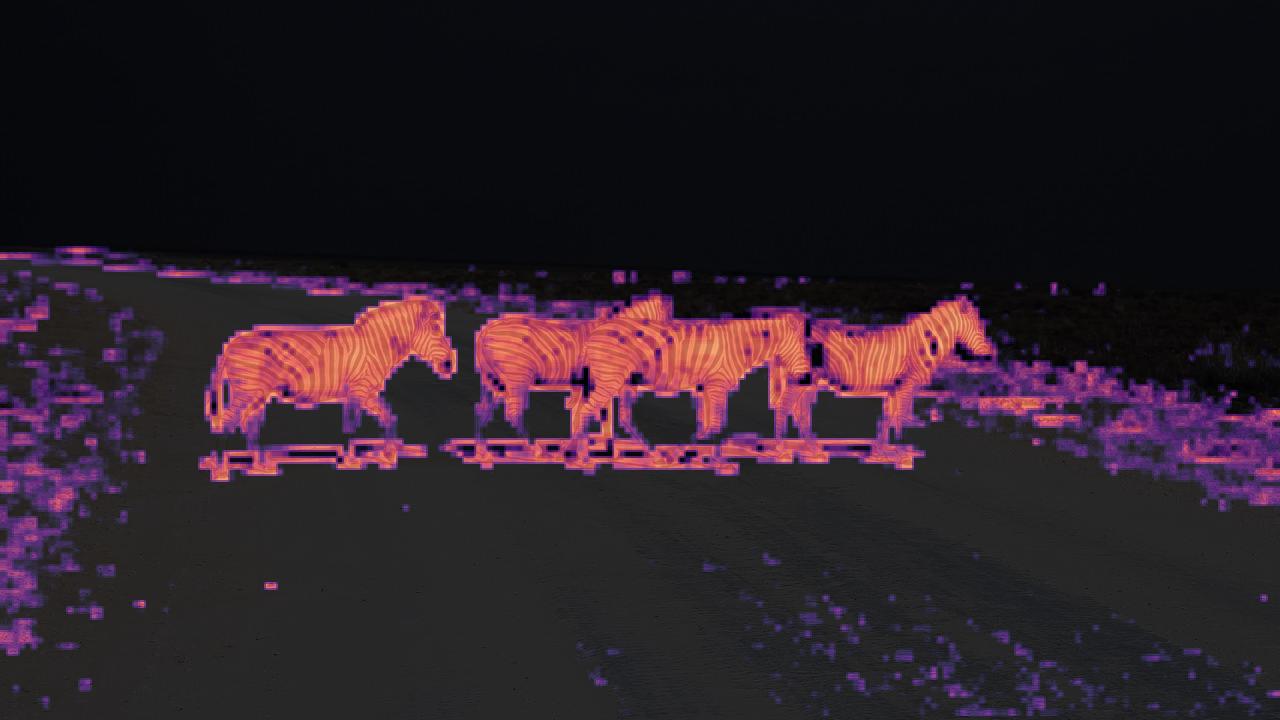}\\[-1mm]
			\small{Anomaly score - \textit{Ours}}&
			\small{Anomaly score - \textit{Uncertainty (Ensemble)}}&
			\small{Anomaly score - \textit{RBM}}
	\end{tabular}
	\vspace{-3mm}
	\caption{
		{\bf Unusual versions of known objects}. 
		Objects of known classes are marked as anomalies because their appearance differs from the examples of this class present in the training data, for example
		the fence in the first row (\textit{fence} class)
		and the dark sky in the third row.
		Note that the \textit{RBM} patch-based method~\cite{Creusot15} is especially sensitive to edges
		and so it detects the zebras very well.
	}
	\label{fig:supplement_road_anomaly_variation}
\end{figure*}

\providecommand{\localwidth}{}
\renewcommand{\localwidth}{0.30\linewidth}

\newcommand{\SamplesFailureHorned}[1]{supplement/figures/anomaly_failure_case/Aihole-Pattadakal_road_#1} %
\newcommand{\SamplesFailureHay}[1]{supplement/figures/anomaly_failure_case/A_man_carrying_dry_grass_on_bicycle_for_domestic_animal_like_cows_#1} %

\begin{figure*}[t]
	\centering
	\begin{tabular}{ccc}
		\includegraphics[width=\localwidth]{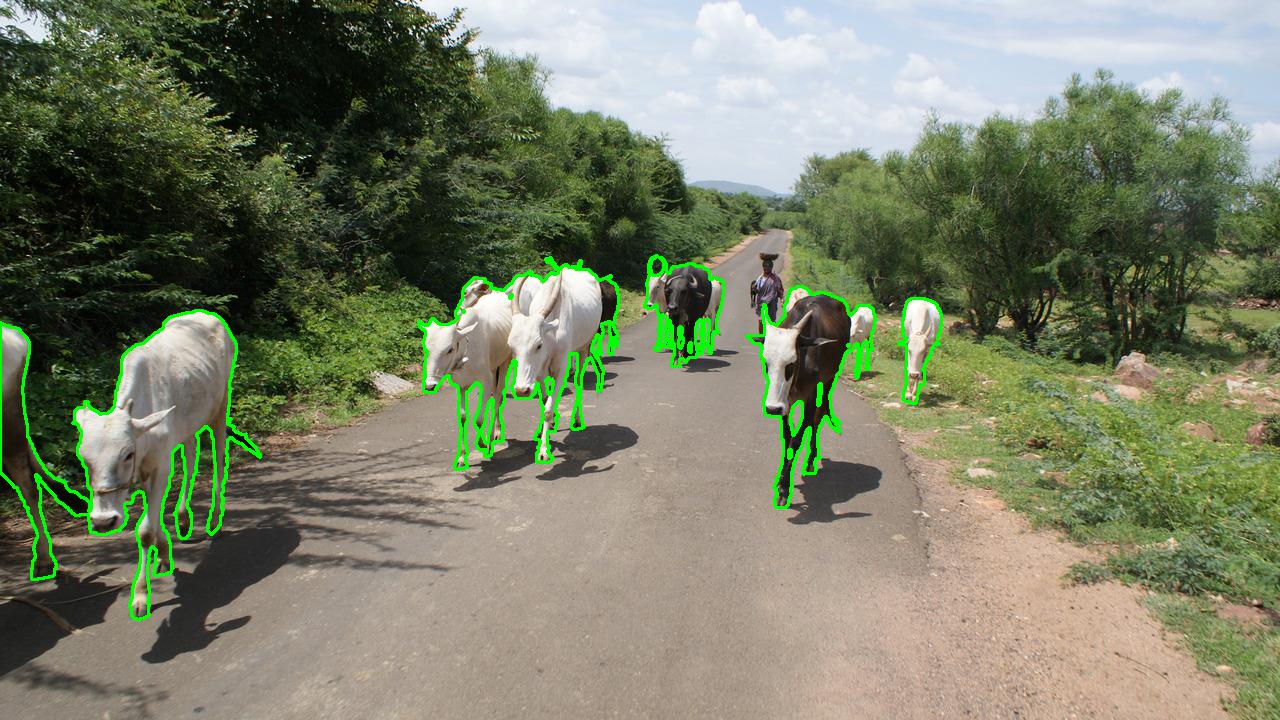}&
		\includegraphics[width=\localwidth]{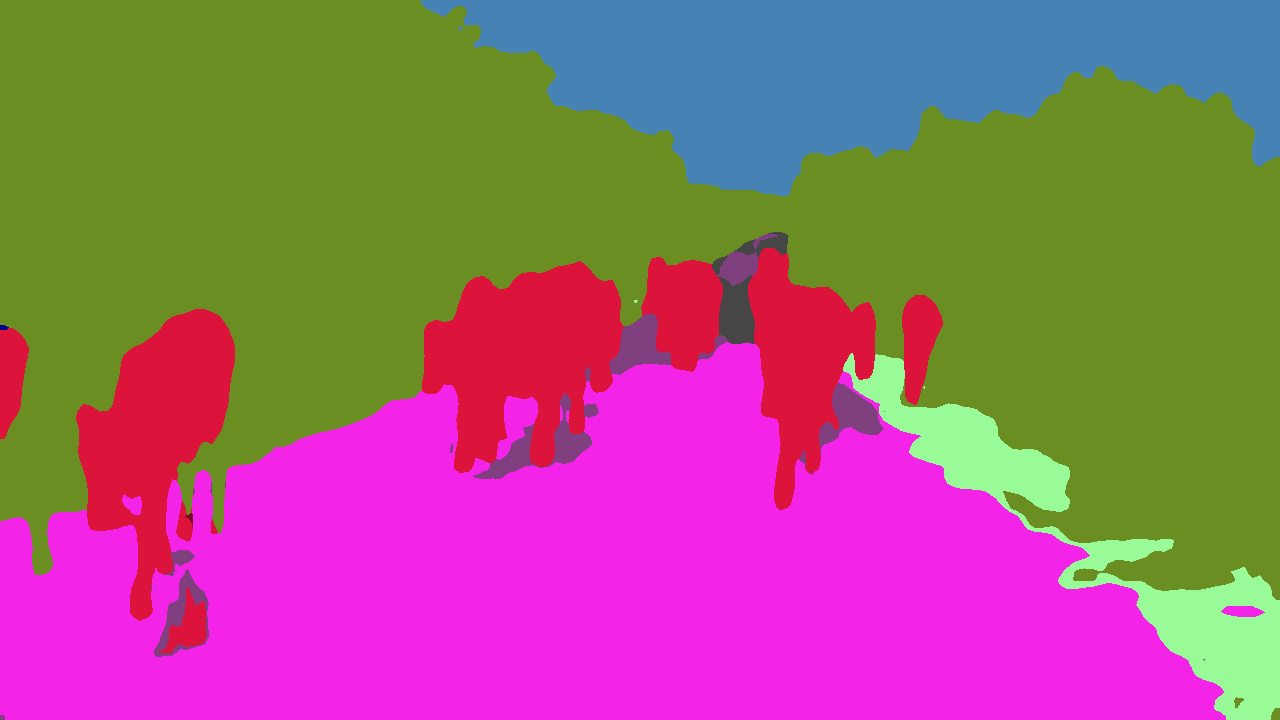}&
		\includegraphics[width=\localwidth]{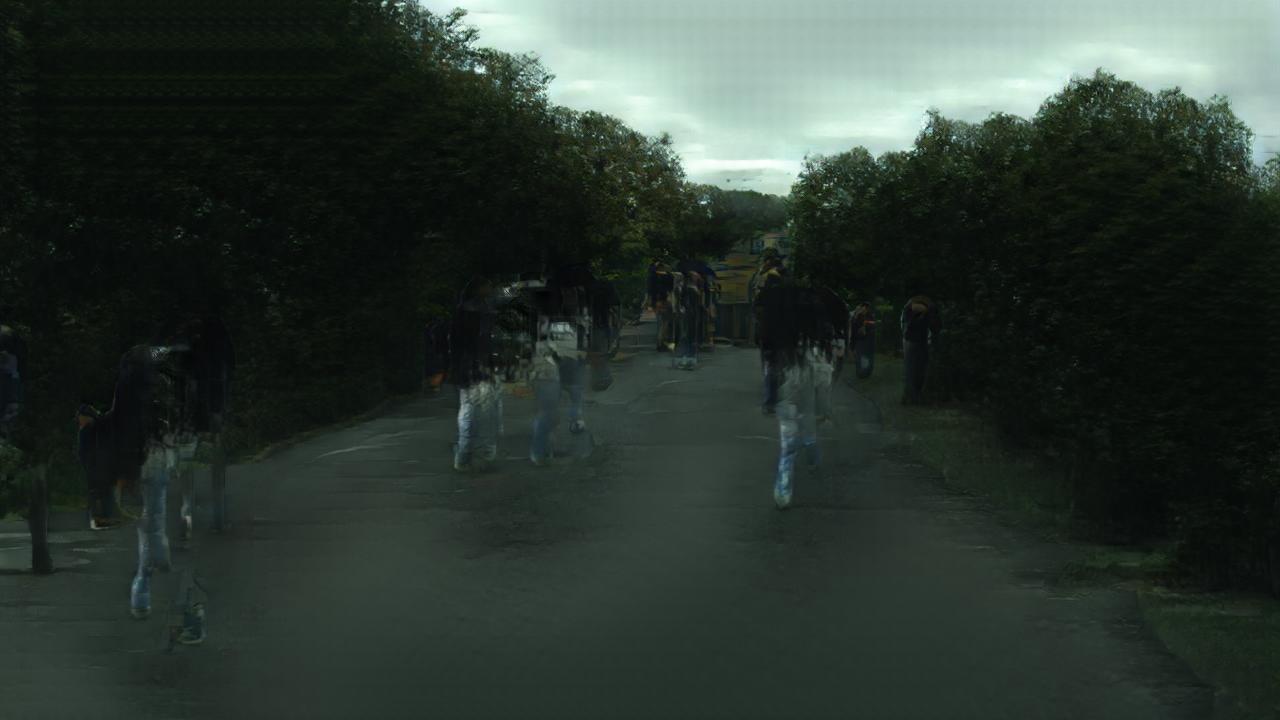}\\[-1mm]
			\small{Input image with anomalies highlighted}&
			\small{Predicted semantic map}&
			\small{Resynthesized image}\\
		\includegraphics[width=\localwidth]{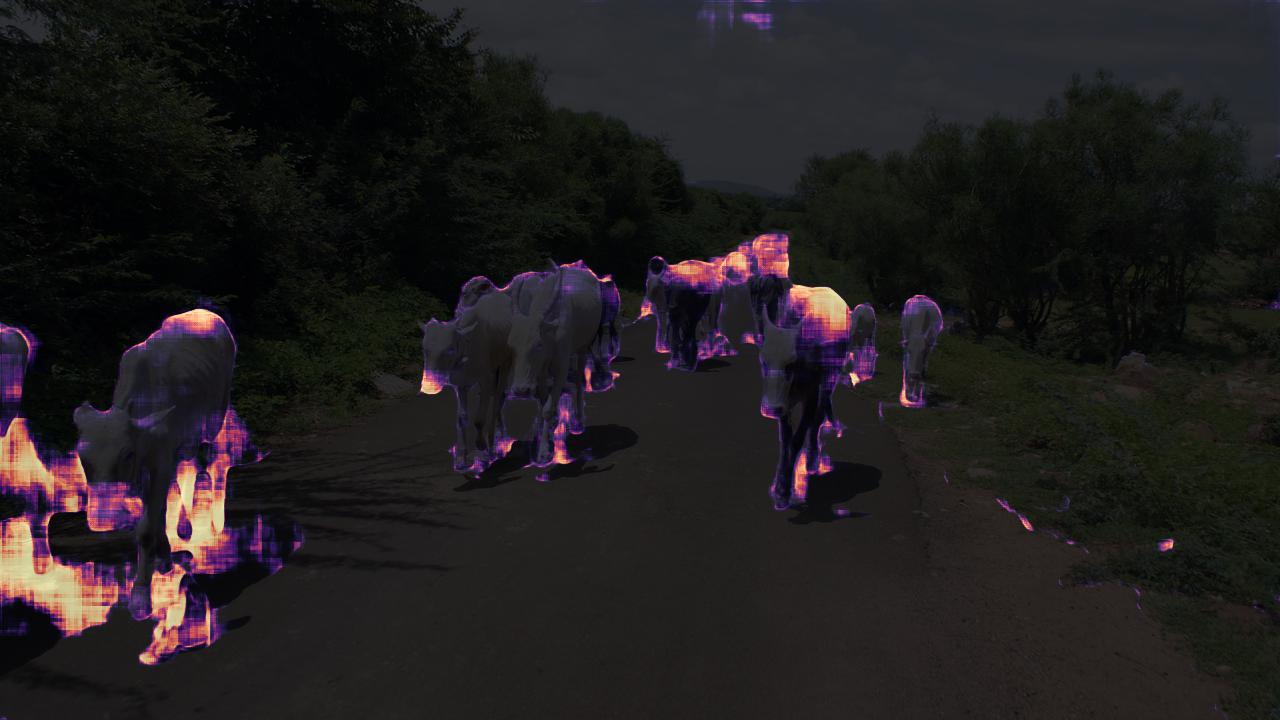}&
		\includegraphics[width=\localwidth]{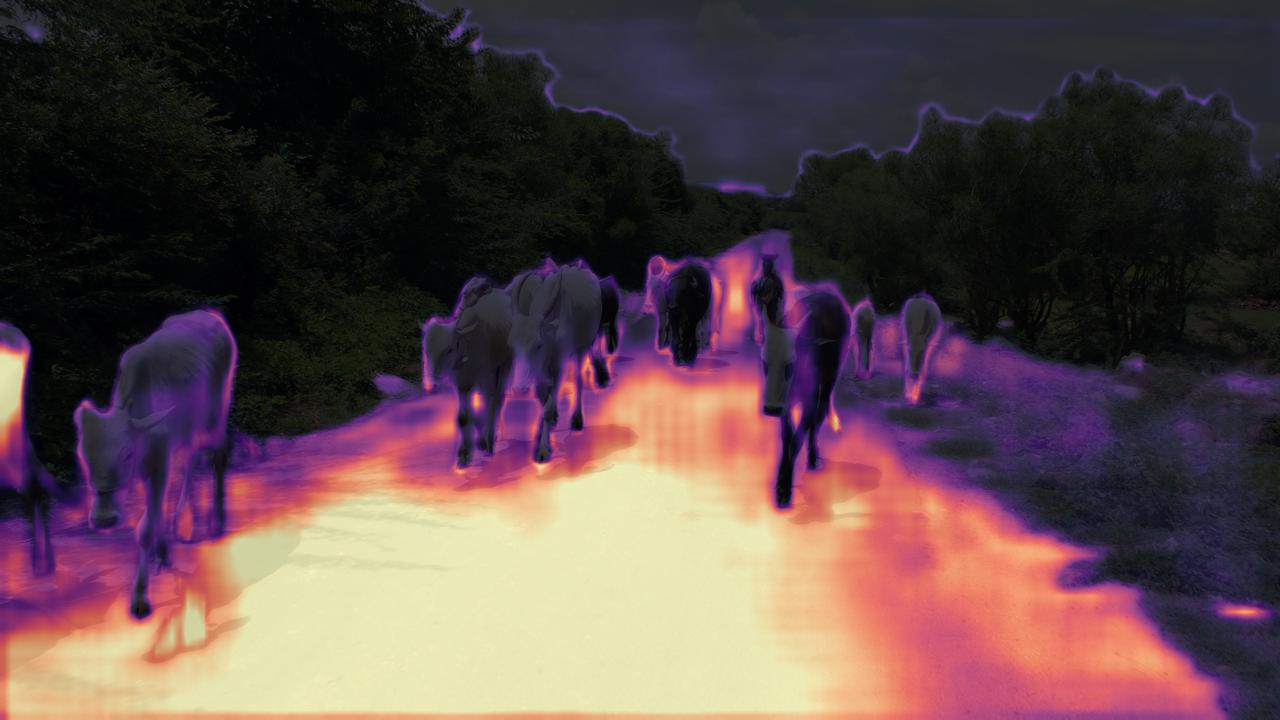}&
		\includegraphics[width=\localwidth]{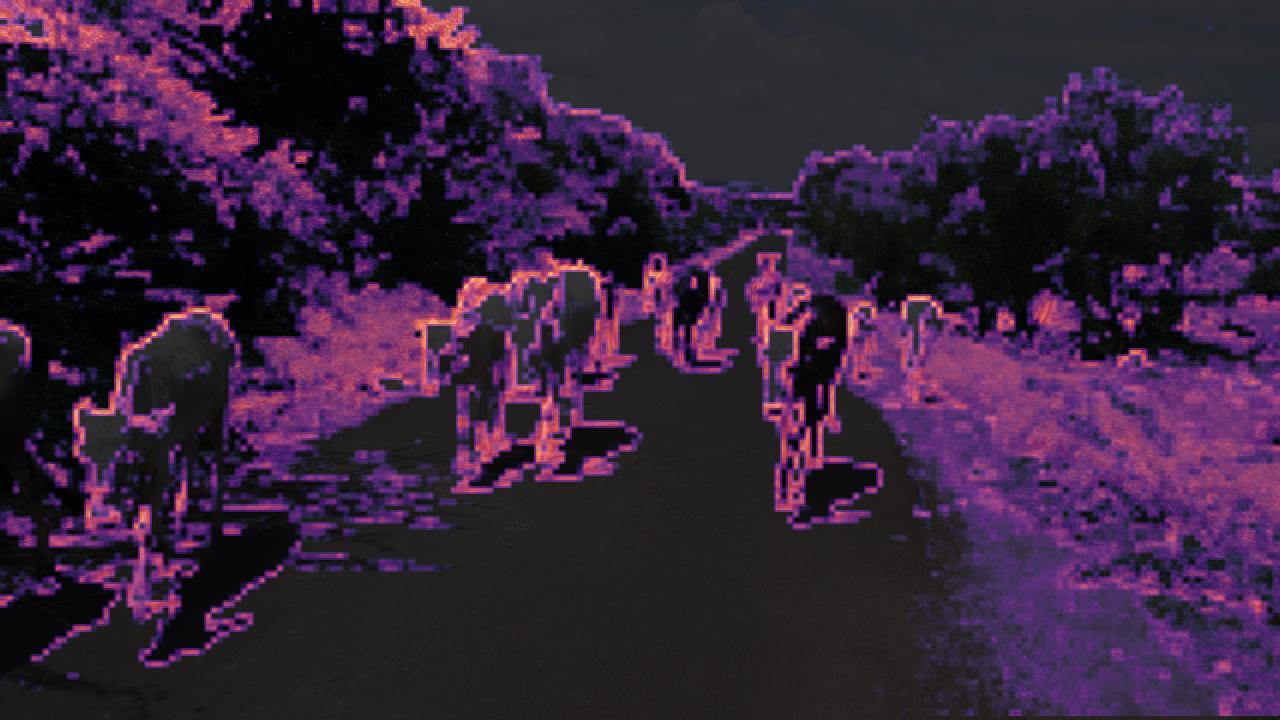}\\[-1mm]
			\small{Anomaly score - \textit{Ours}}&
			\small{Anomaly score - \textit{Uncertainty (Ensemble)}}&
			\small{Anomaly score - \textit{RBM}}\\
		\includegraphics[width=\localwidth]{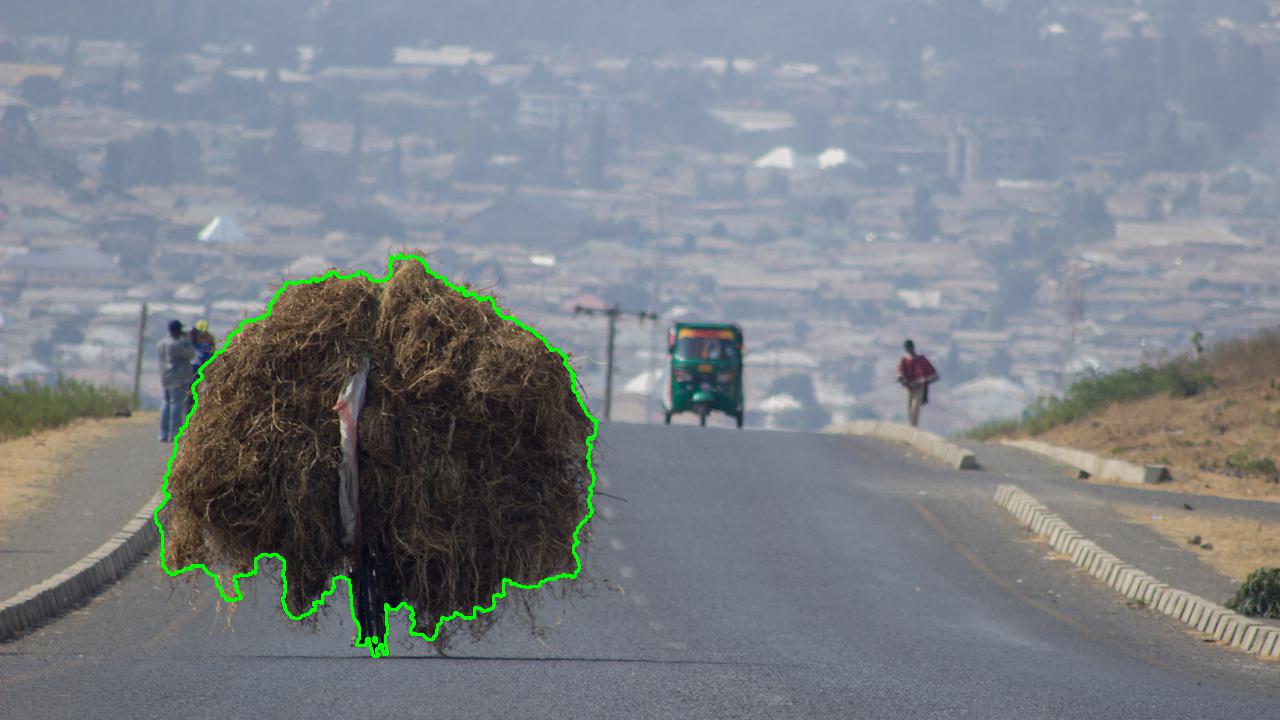}&
		\includegraphics[width=\localwidth]{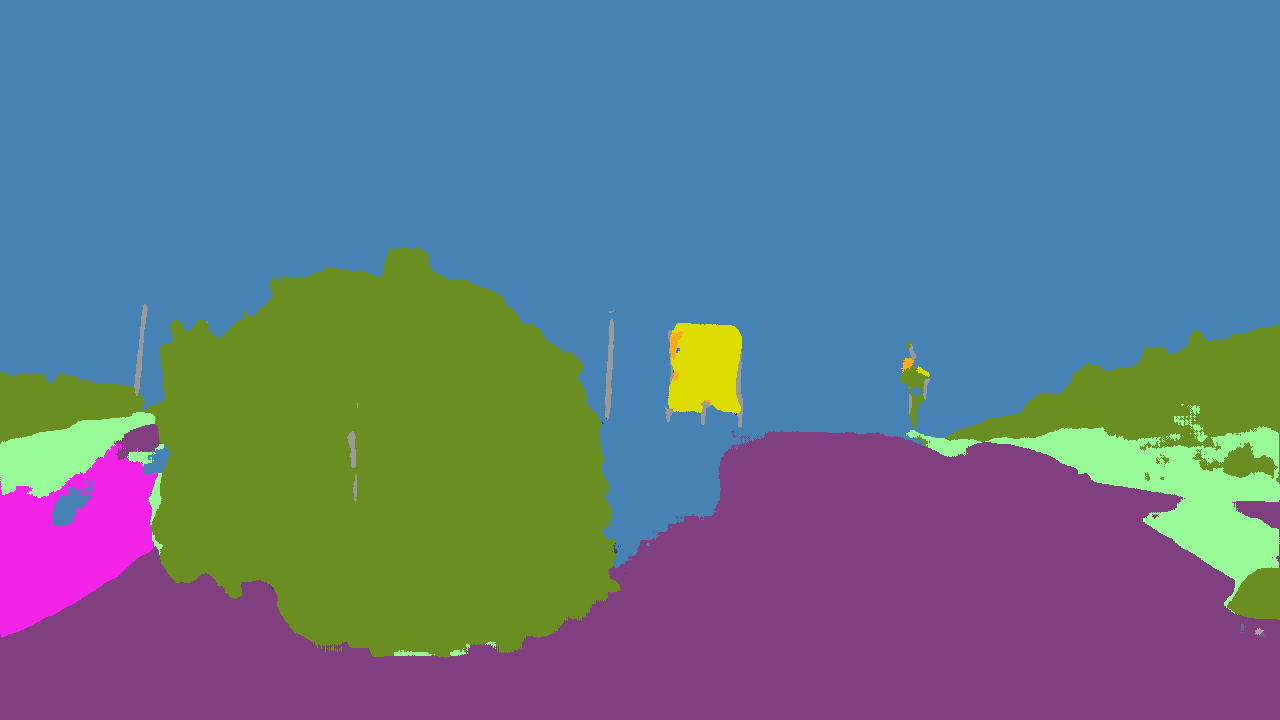}&
		\includegraphics[width=\localwidth]{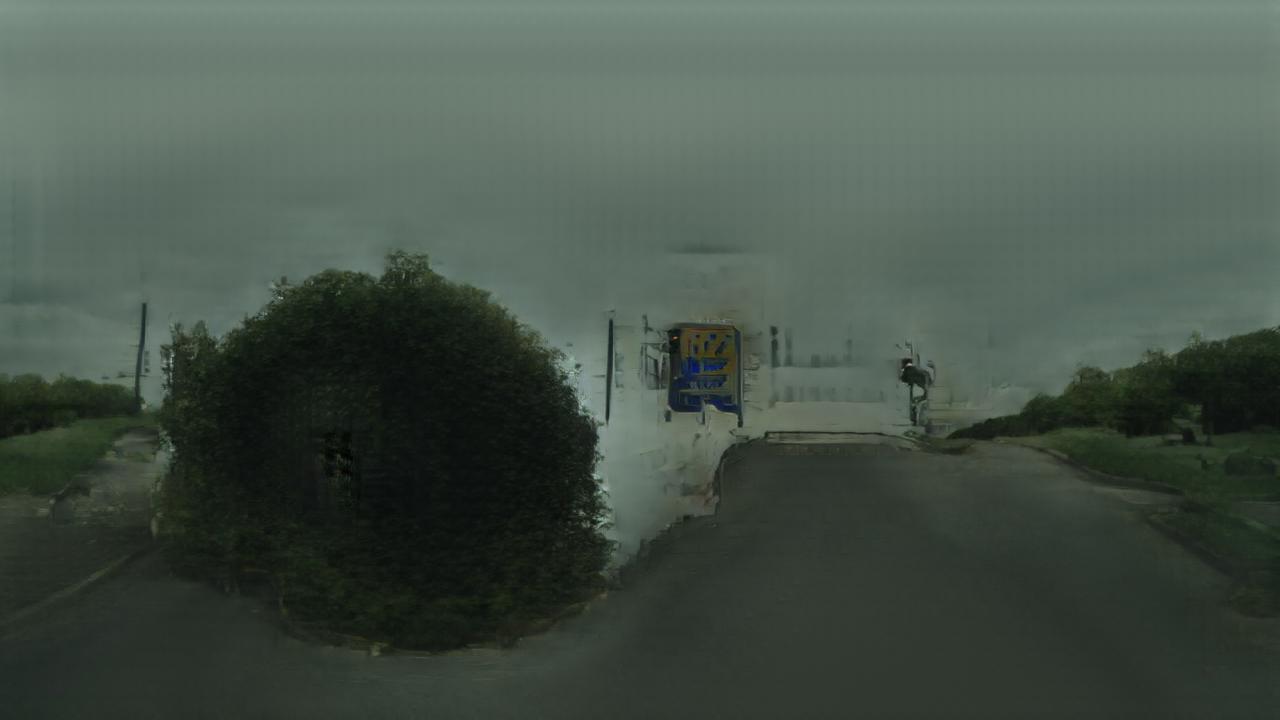}\\[-1mm]
			\small{Input image with anomalies highlighted}&
			\small{Predicted semantic map}&
			\small{Resynthesized image}\\
		\includegraphics[width=\localwidth]{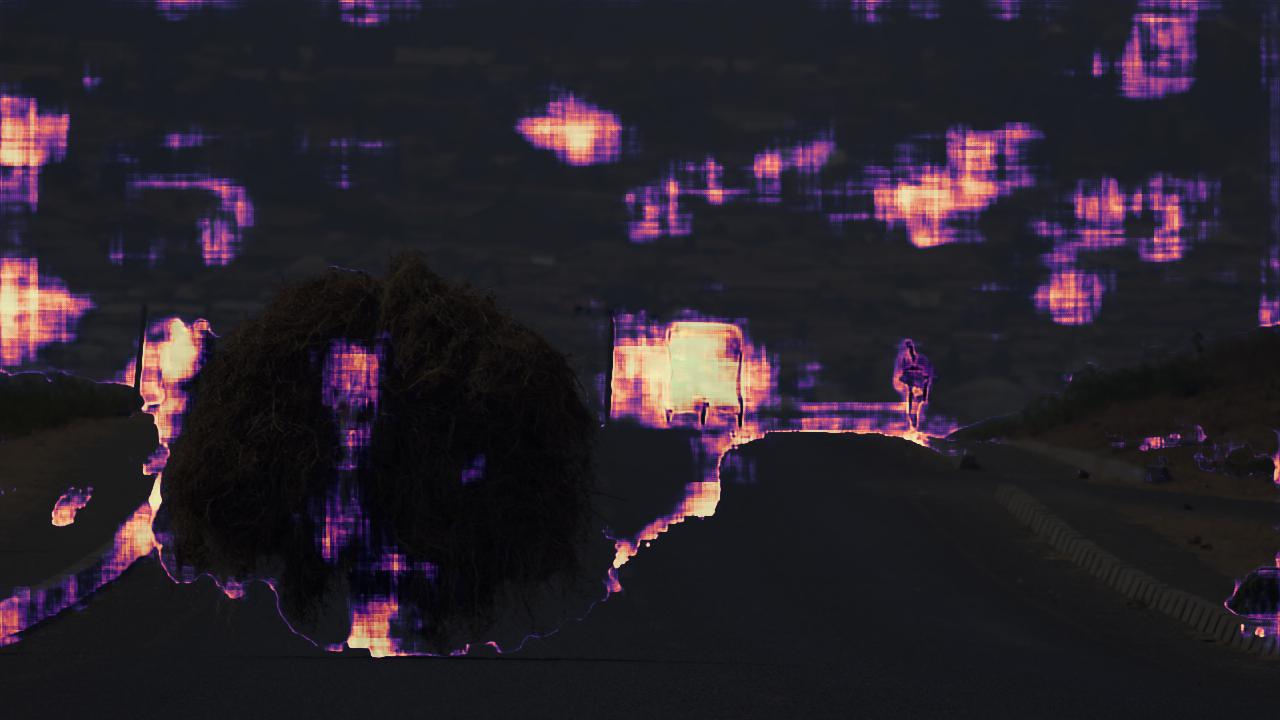}&
		\includegraphics[width=\localwidth]{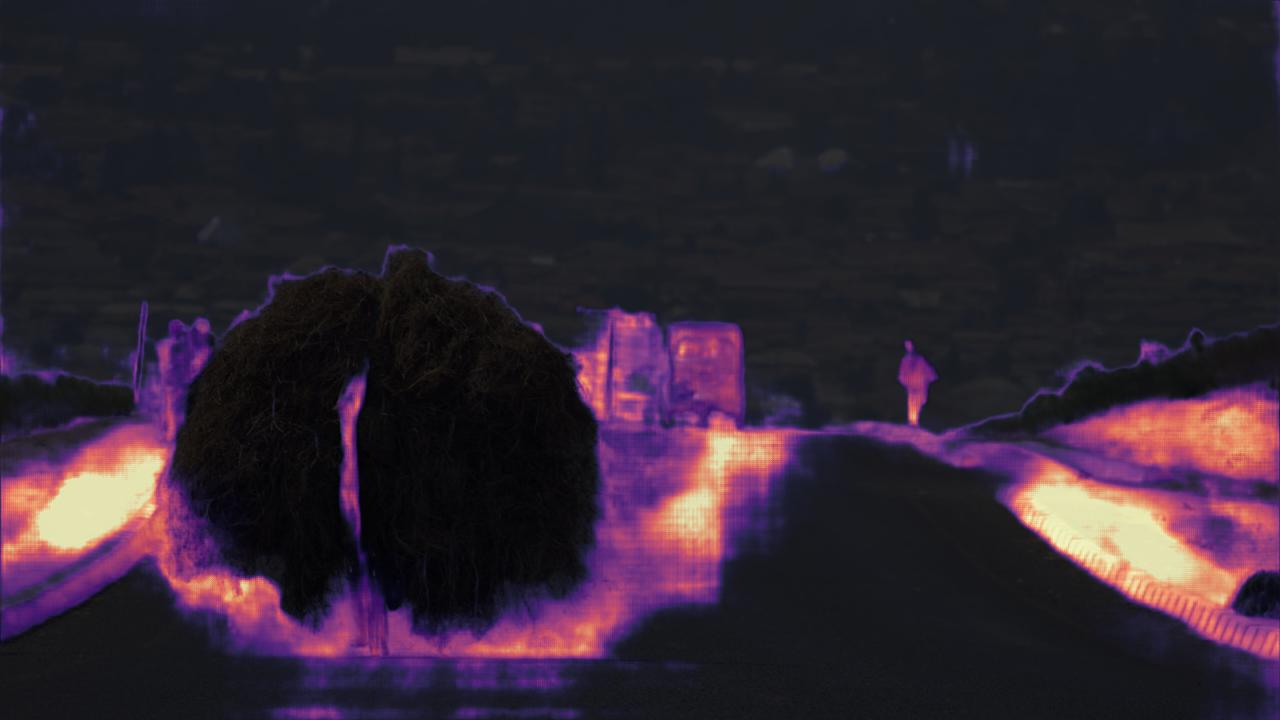}&
		\includegraphics[width=\localwidth]{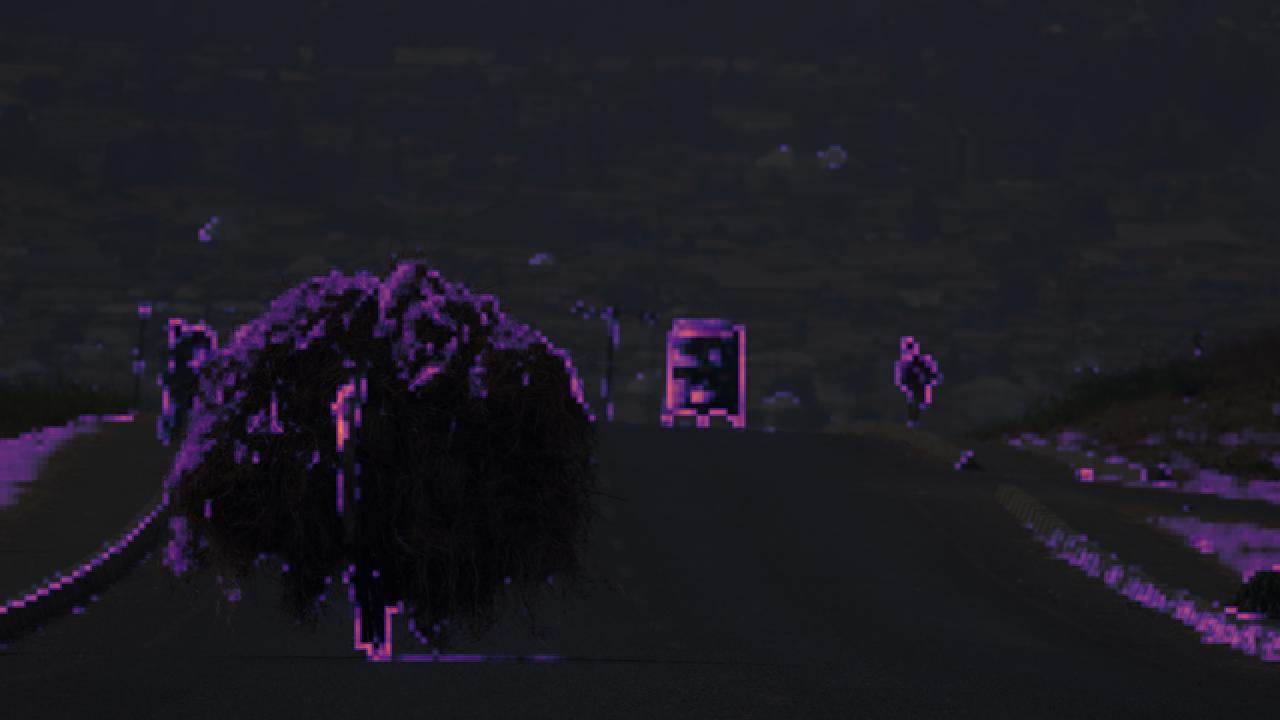}\\[-1mm]
			\small{Anomaly score - \textit{Ours}}&
			\small{Anomaly score - \textit{Uncertainty (Dropout)}}&
			\small{Anomaly score - \textit{RBM}}\\
	\end{tabular}
	\vspace{-3mm}
	\caption{
		{\bf Failure cases.} 
		Our approach sometimes fails when the anomaly bears resemblance to an existing class:
		For example, animals classified as people in the first row or 
		transported hay classified as vegetation in the third row.
		The system as a whole is nonetheless still aware of the obstacle because of its presence in the semantic map.
	}
	\label{fig:supplement_anomaly_failure_case}
\end{figure*}

\providecommand{\localwidth}{}
\renewcommand{\localwidth}{0.30\linewidth}

\providecommand{\localwidth}{}
\renewcommand{\localwidth}{0.19\linewidth}

\begin{figure*}[t!]
	\centering
\begin{subfigure}[t]{\localwidth}
		\begin{subfigure}[t]{\linewidth}
				\centering
			\includegraphics[width=\linewidth]{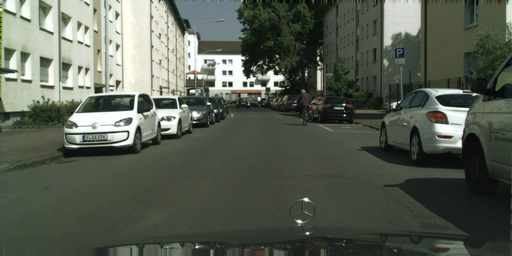}
			\includegraphics[width=\linewidth]{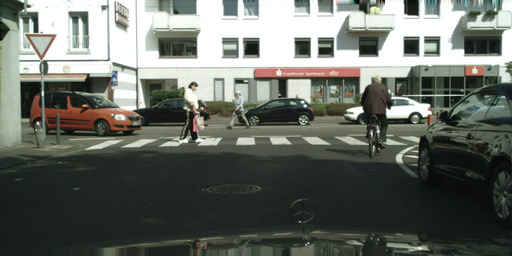}
			\includegraphics[width=\linewidth]{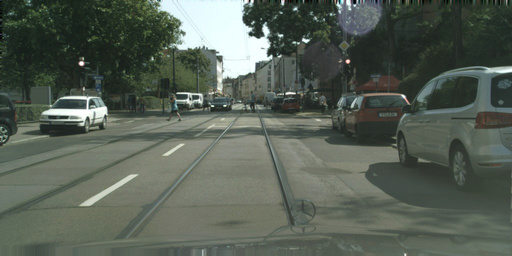}
			\includegraphics[width=\linewidth]{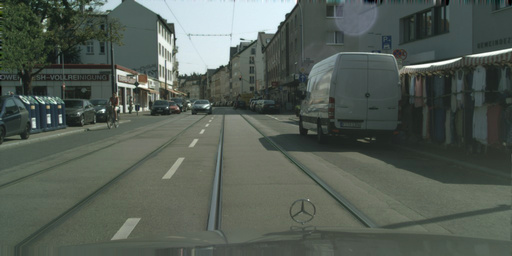}
			\includegraphics[width=\linewidth]{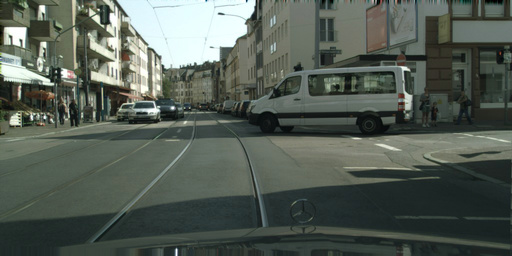}
			\includegraphics[width=\linewidth]{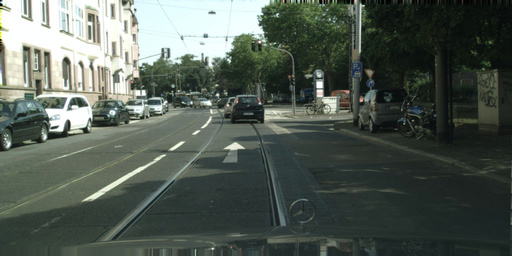}
			\includegraphics[width=\linewidth]{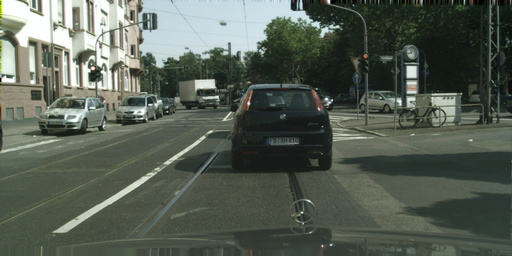}
			\includegraphics[width=\linewidth]{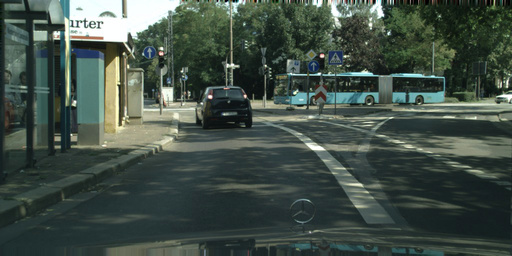}
			\includegraphics[width=\linewidth]{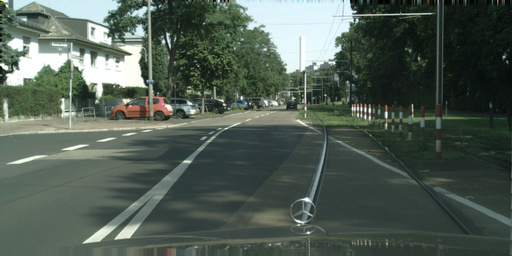}
			\includegraphics[width=\linewidth]{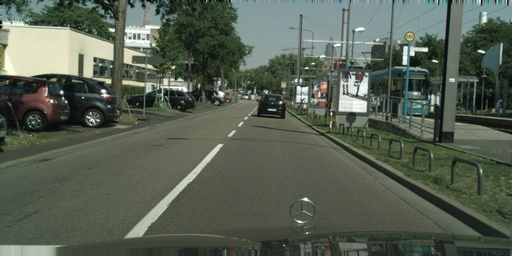}
			\includegraphics[width=\linewidth]{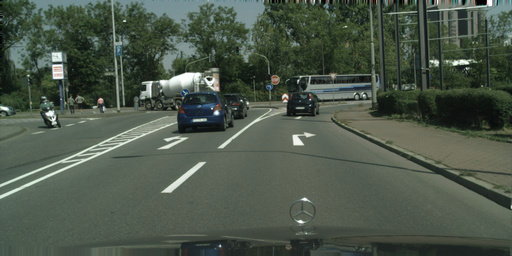}
			\caption{	\centering Input image (normal)}
		\end{subfigure}
	
\end{subfigure}
\begin{subfigure}[t]{\localwidth}
		\begin{subfigure}[t]{\linewidth}
		\includegraphics[width=\linewidth]{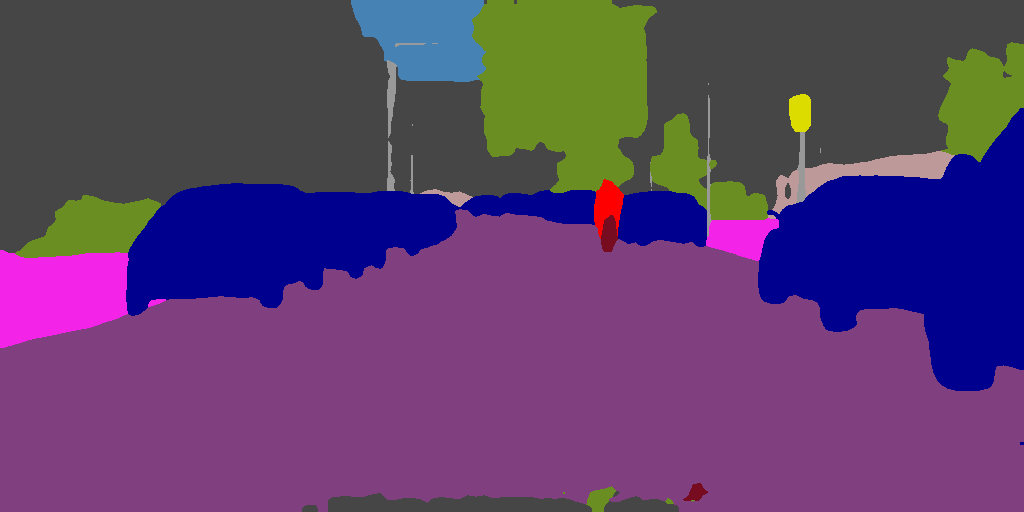}
		\includegraphics[width=\linewidth]{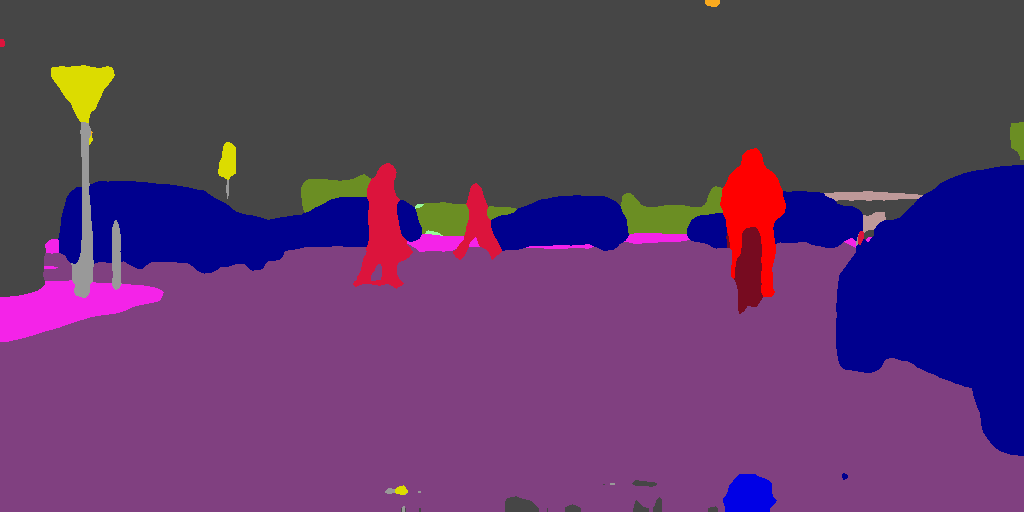}
		\includegraphics[width=\linewidth]{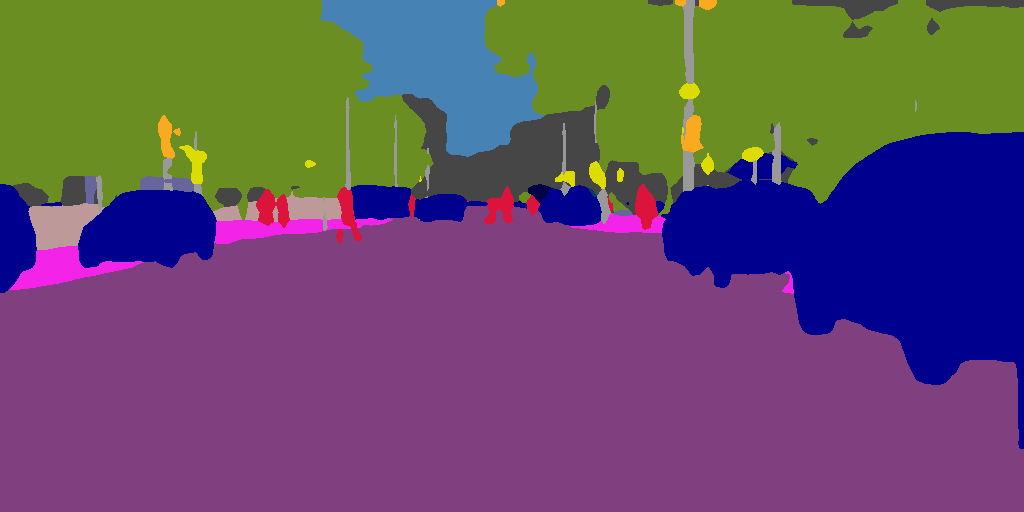}
		\includegraphics[width=\linewidth]{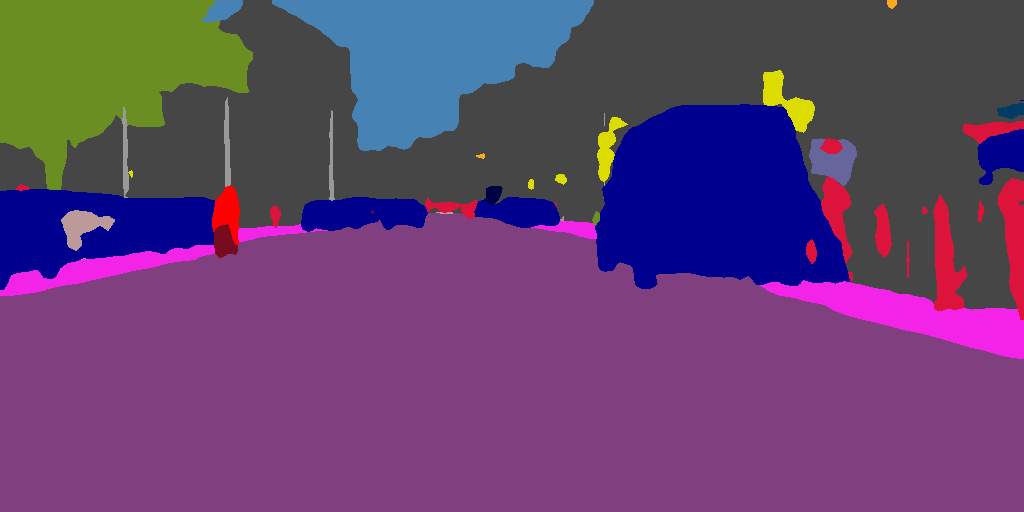}
		\includegraphics[width=\linewidth]{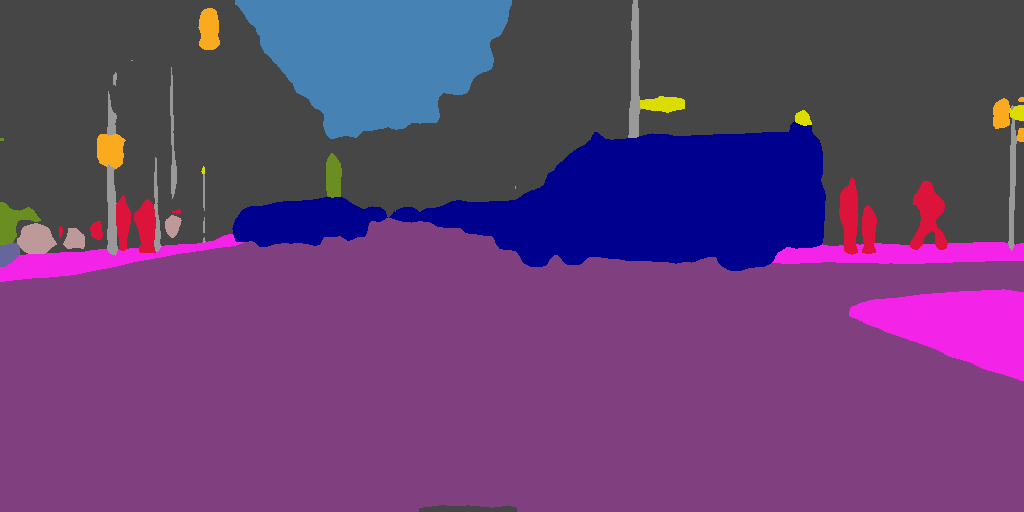}
		\includegraphics[width=\linewidth]{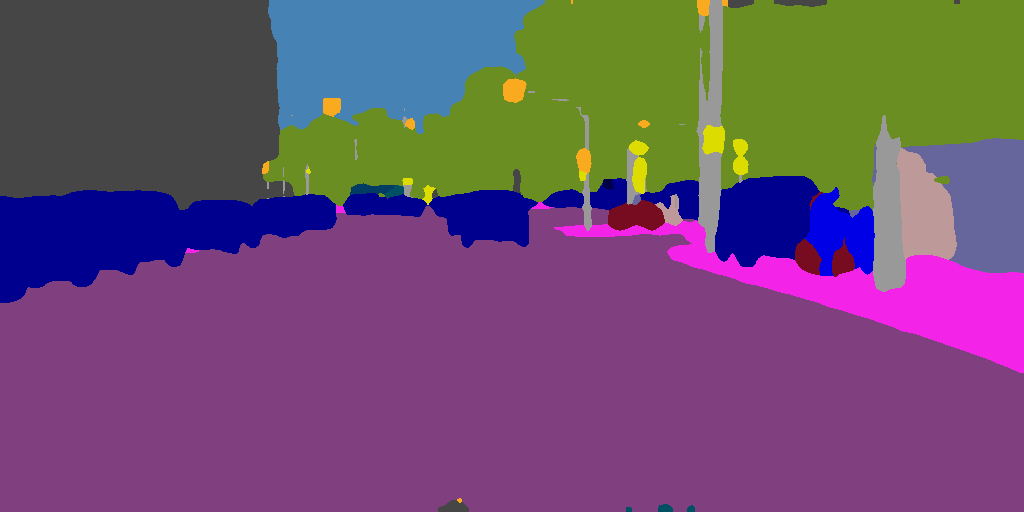}
		\includegraphics[width=\linewidth]{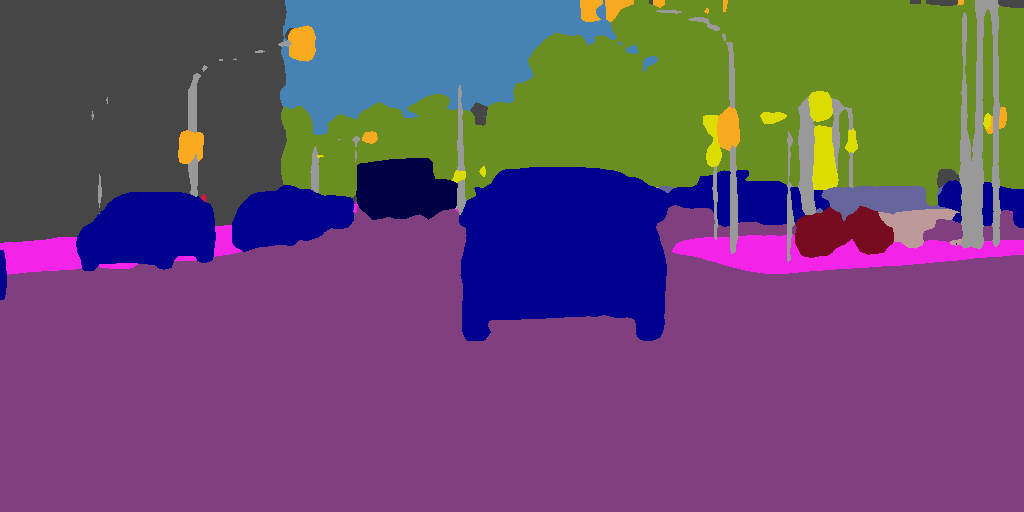}
		\includegraphics[width=\linewidth]{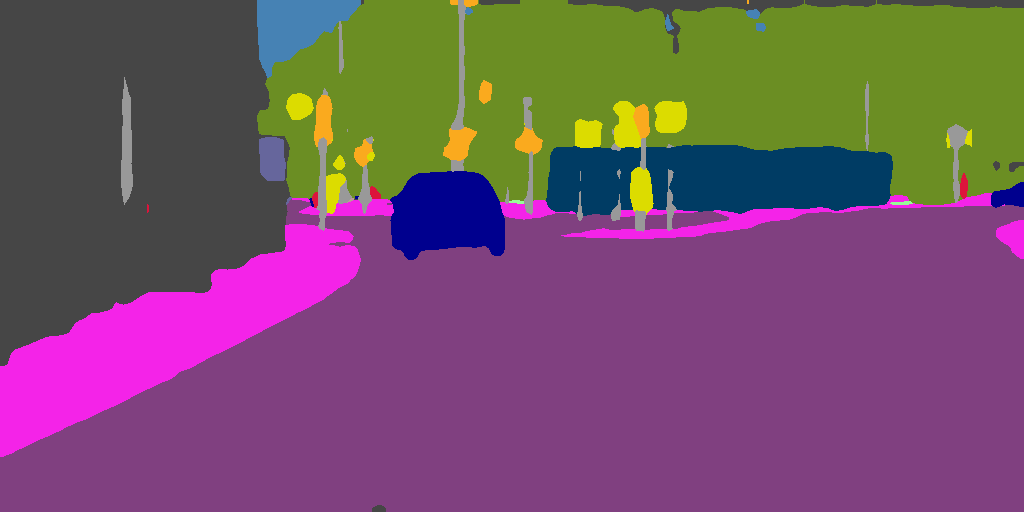}
		\includegraphics[width=\linewidth]{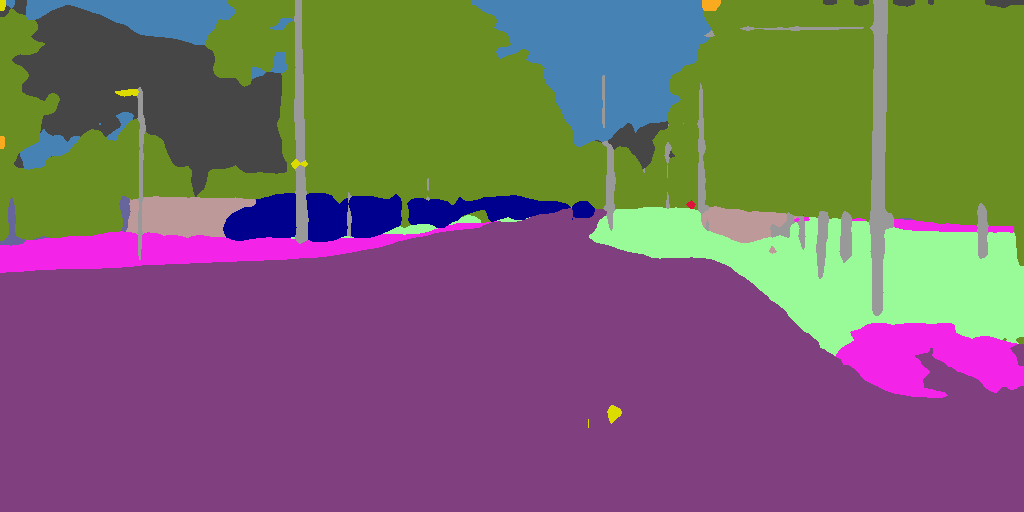}
		\includegraphics[width=\linewidth]{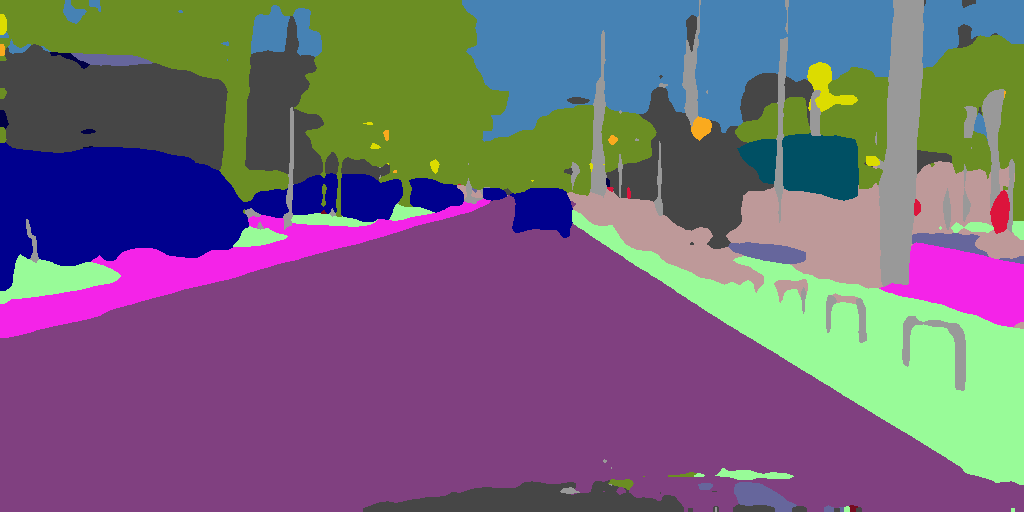}
		\includegraphics[width=\linewidth]{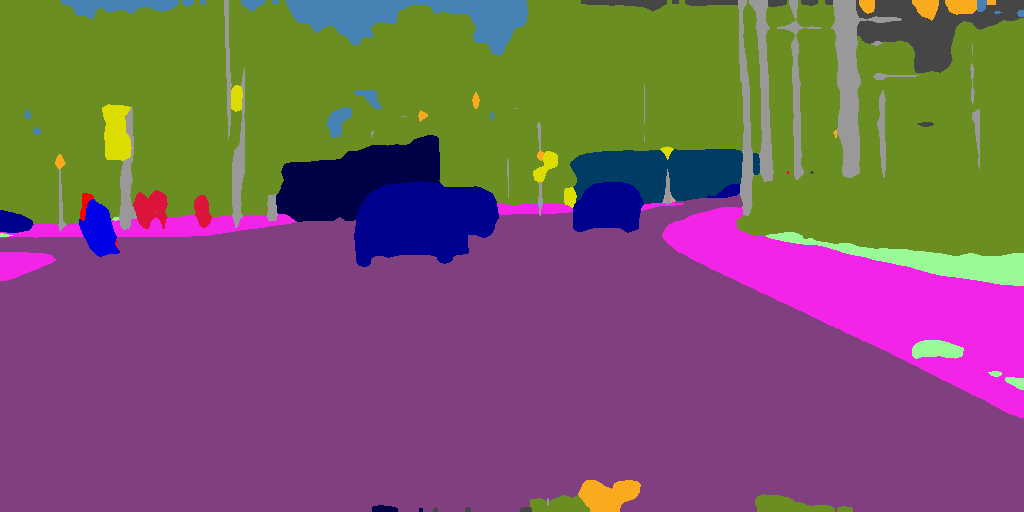}
		\caption{\centering Predicted map (normal) }
	\end{subfigure}
\end{subfigure}
\begin{subfigure}[t]{\localwidth}
	\begin{subfigure}[t]{\linewidth}
	\includegraphics[width=\linewidth]{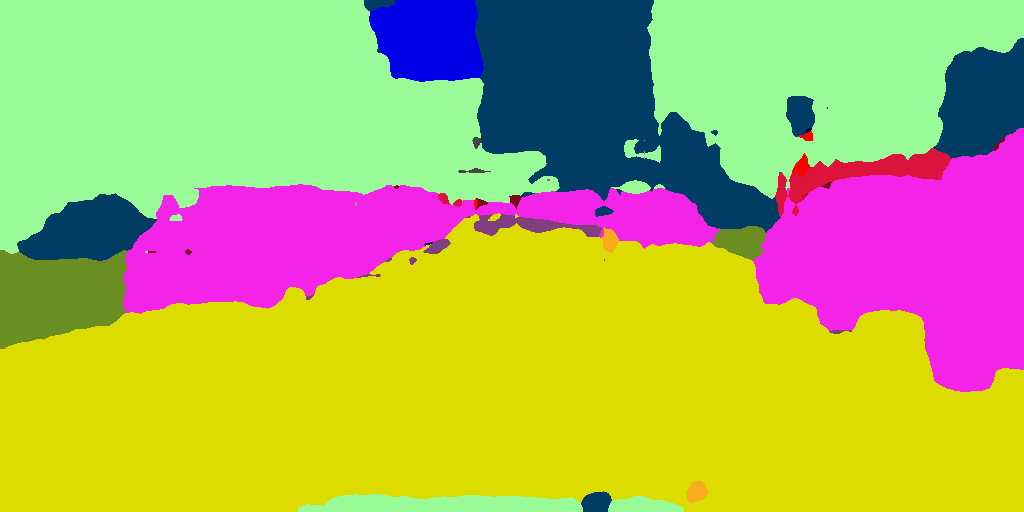}
	\includegraphics[width=\linewidth]{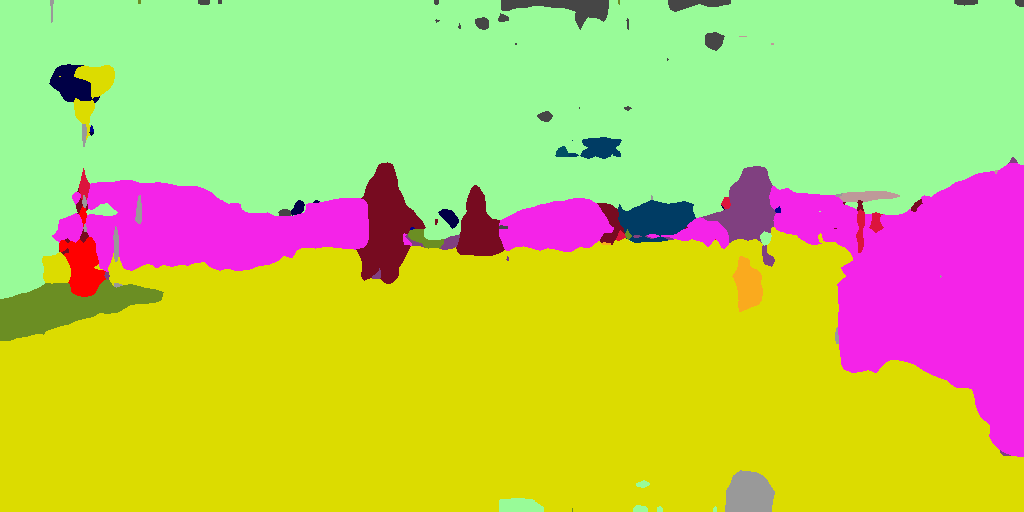}
	\includegraphics[width=\linewidth]{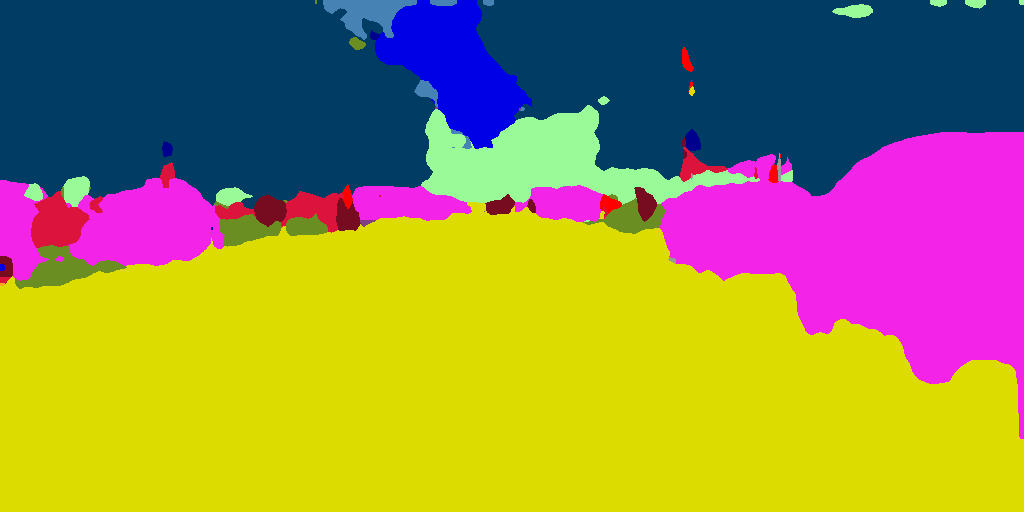}
	\includegraphics[width=\linewidth]{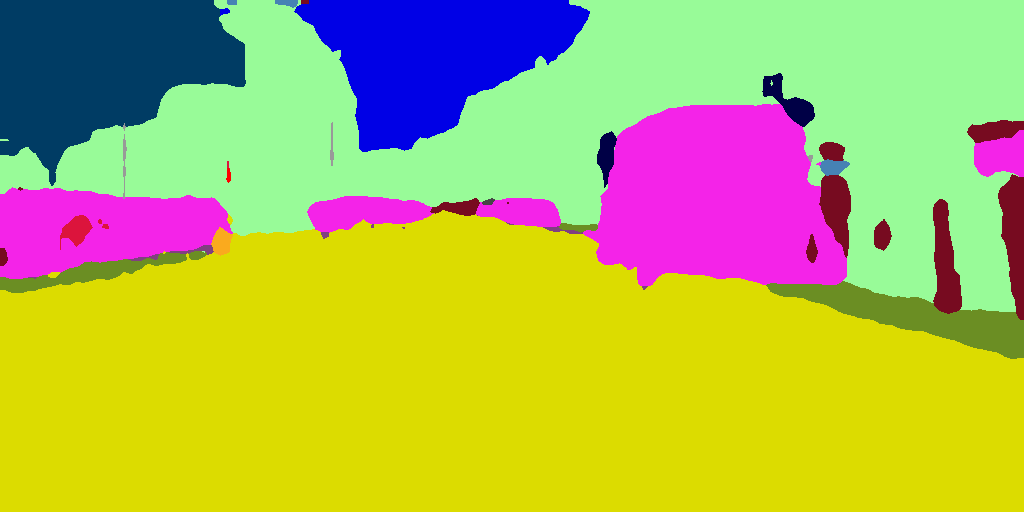}
	\includegraphics[width=\linewidth]{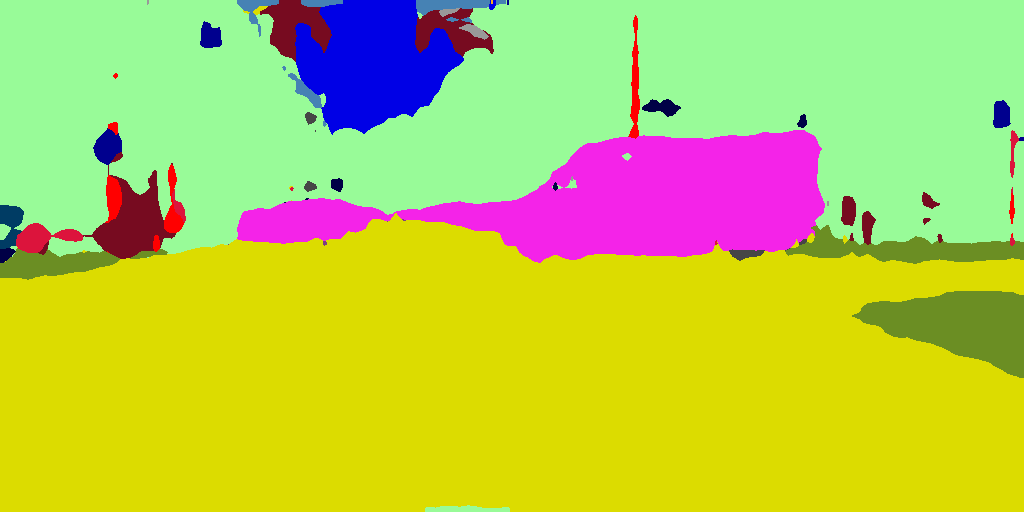}
	\includegraphics[width=\linewidth]{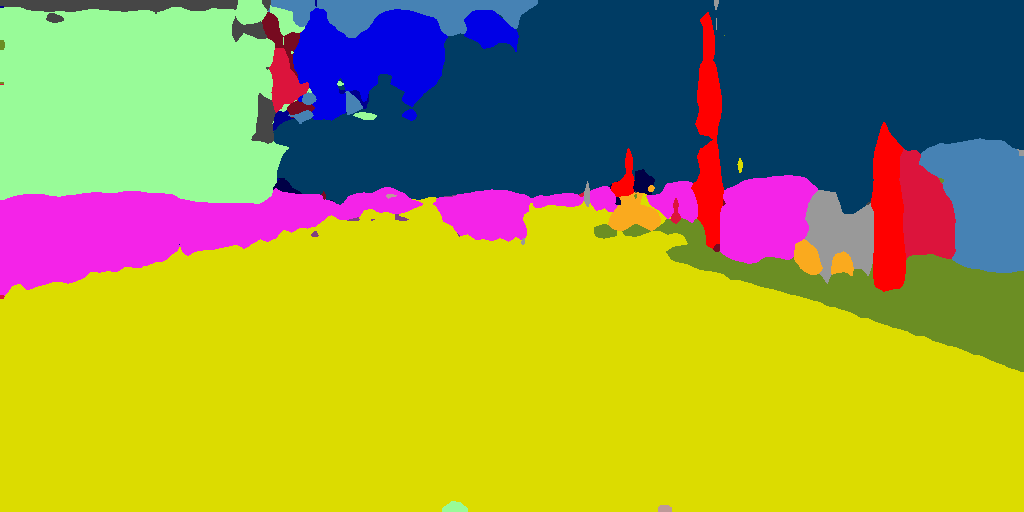}
	\includegraphics[width=\linewidth]{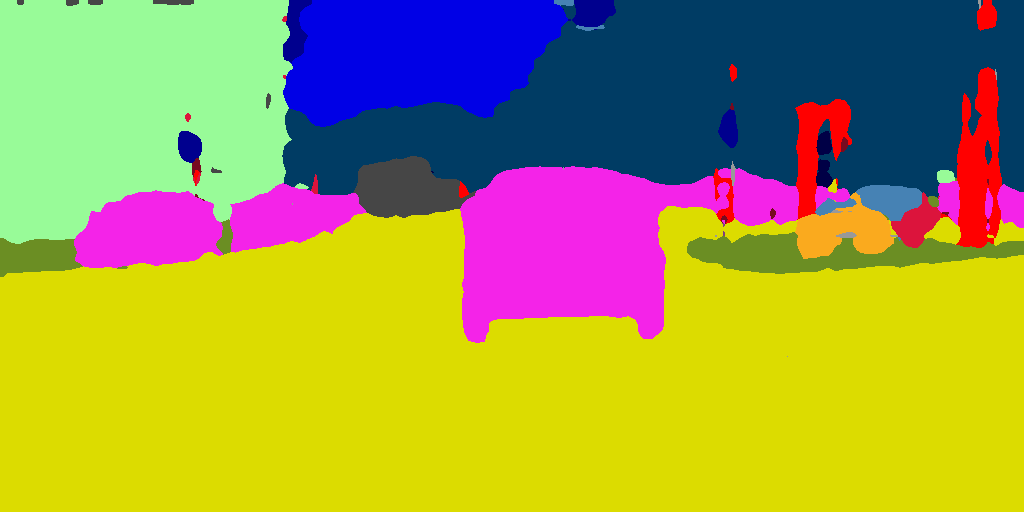}
	\includegraphics[width=\linewidth]{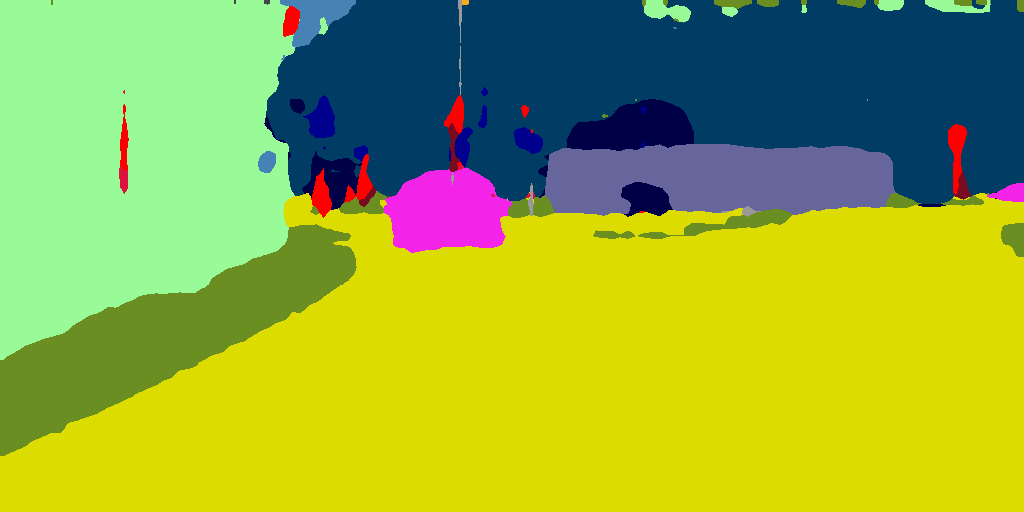}
	\includegraphics[width=\linewidth]{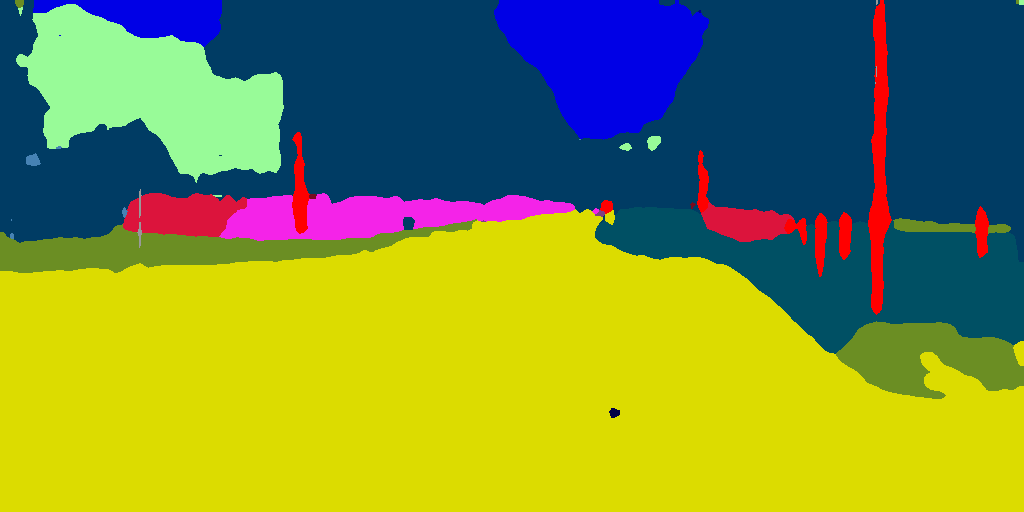}
	\includegraphics[width=\linewidth]{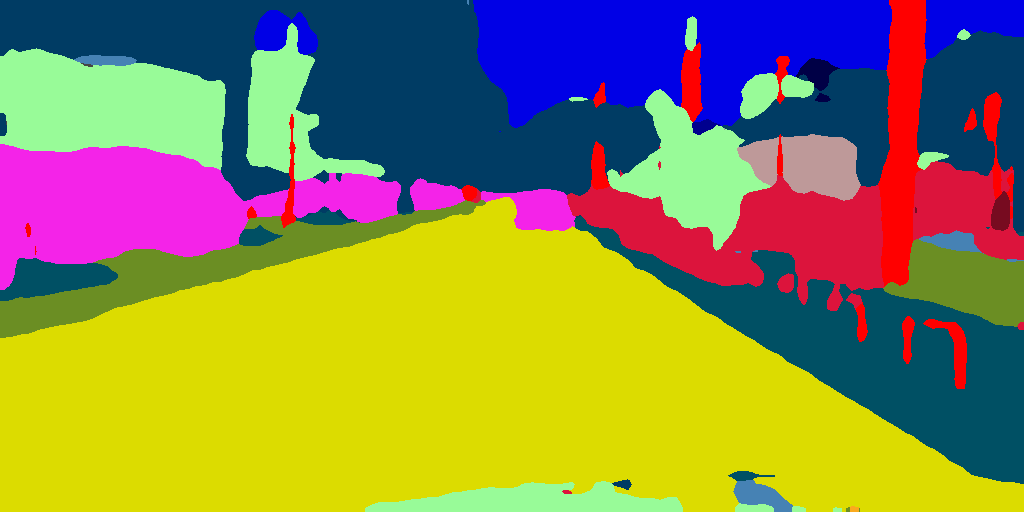}
	\includegraphics[width=\linewidth]{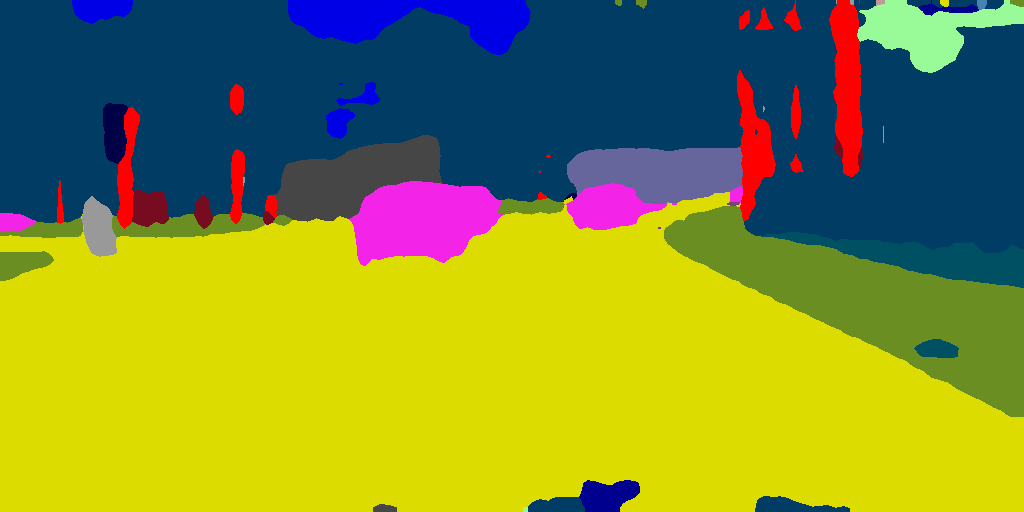}

						\caption{\centering Predicted map (Shift)}
	\end{subfigure}
 \end{subfigure}
\begin{subfigure}[t]{\localwidth}
	\begin{subfigure}[t]{\linewidth}
	\includegraphics[width=\linewidth]{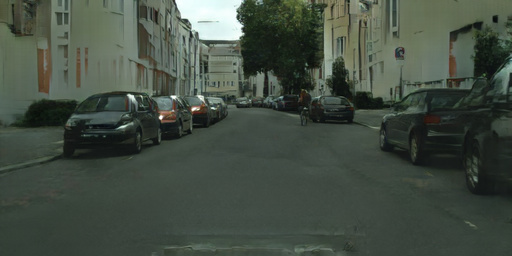}
\includegraphics[width=\linewidth]{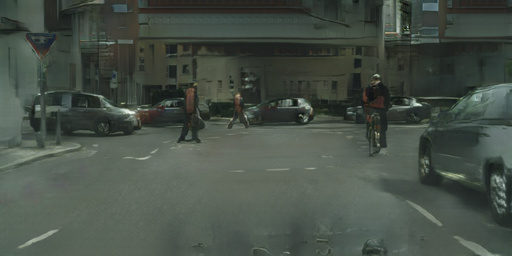}
\includegraphics[width=\linewidth]{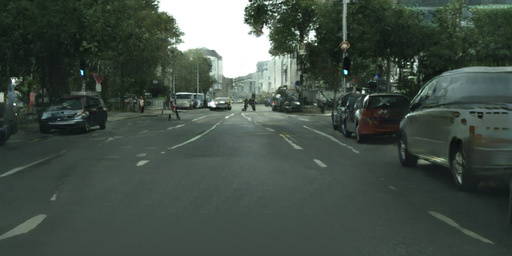}
\includegraphics[width=\linewidth]{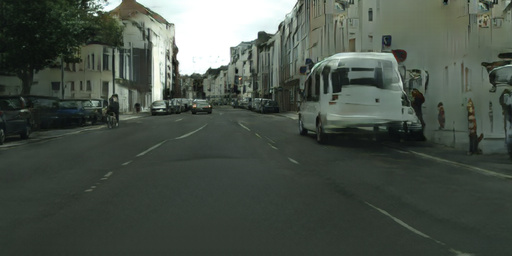}
\includegraphics[width=\linewidth]{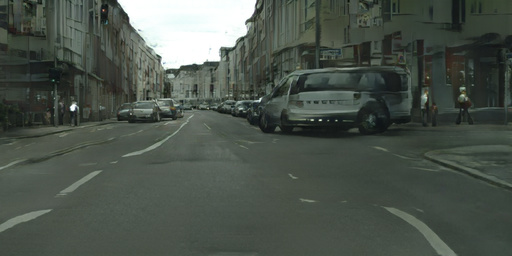}
\includegraphics[width=\linewidth]{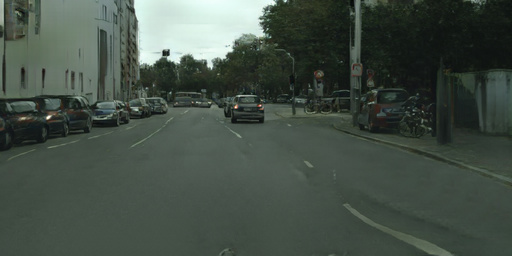}
\includegraphics[width=\linewidth]{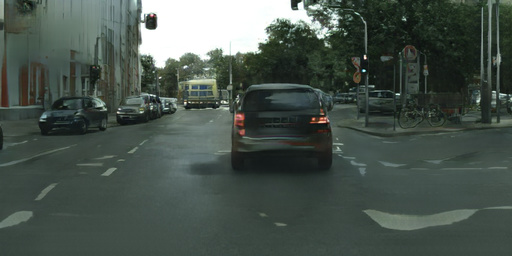}
\includegraphics[width=\linewidth]{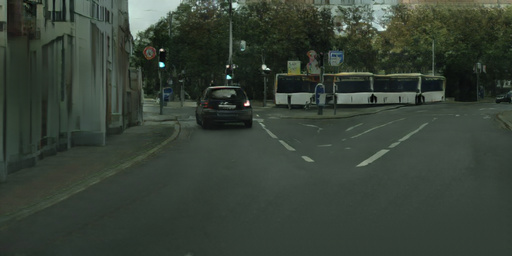}
\includegraphics[width=\linewidth]{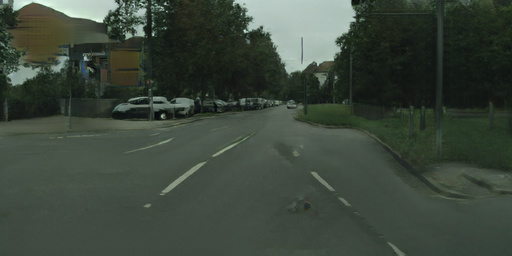}
\includegraphics[width=\linewidth]{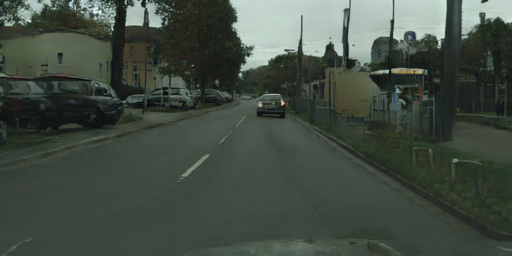}
\includegraphics[width=\linewidth]{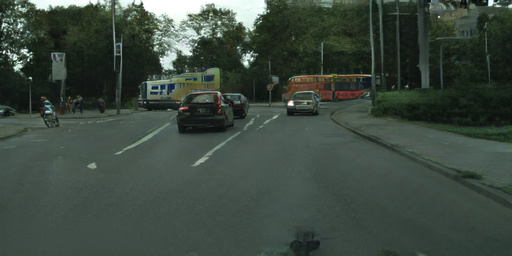}
			\caption{\centering Resynthesized image (normal)}
	\end{subfigure}		
\end{subfigure}
\begin{subfigure}[t]{\localwidth}
	\begin{subfigure}[t]{\linewidth}
	\includegraphics[width=\linewidth]{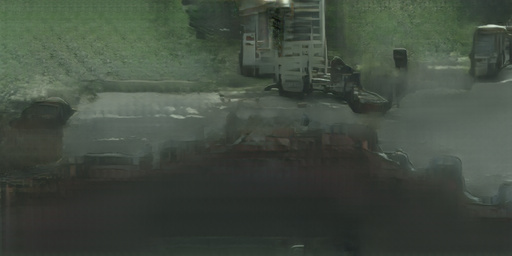}
	\includegraphics[width=\linewidth]{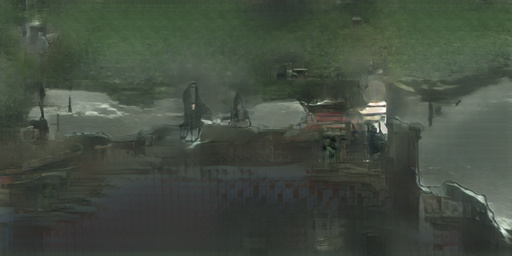}
	\includegraphics[width=\linewidth]{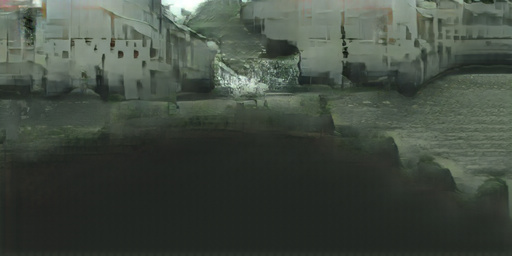}
	\includegraphics[width=\linewidth]{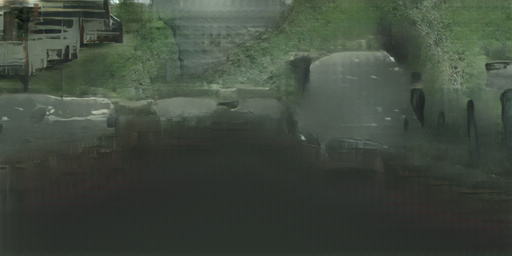}
	\includegraphics[width=\linewidth]{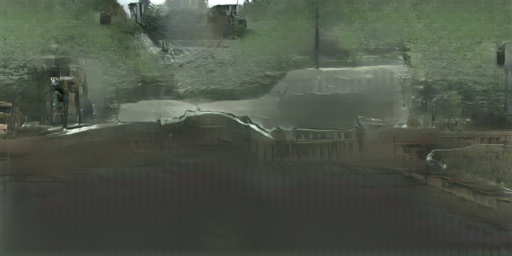}
	\includegraphics[width=\linewidth]{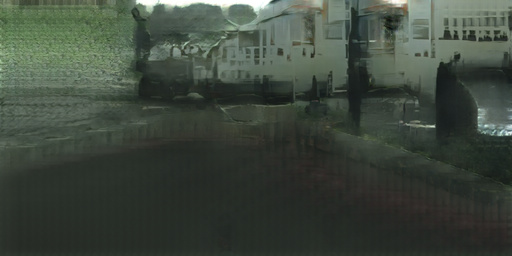}
	\includegraphics[width=\linewidth]{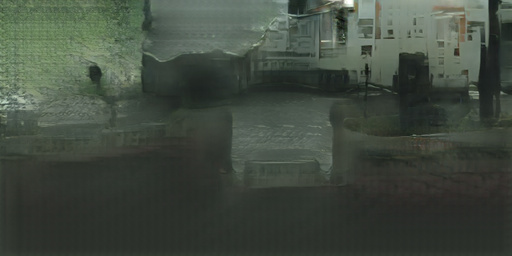}
	\includegraphics[width=\linewidth]{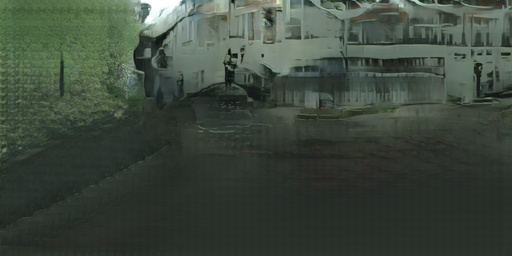}
	\includegraphics[width=\linewidth]{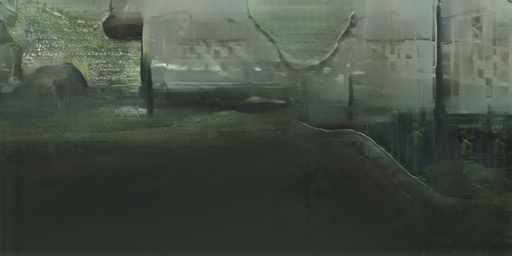}
	\includegraphics[width=\linewidth]{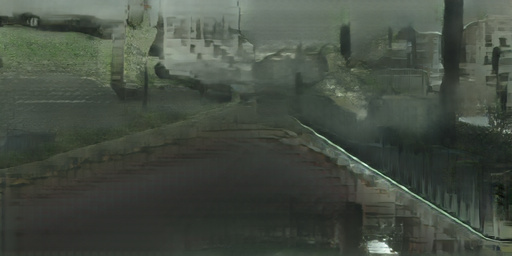}
	\includegraphics[width=\linewidth]{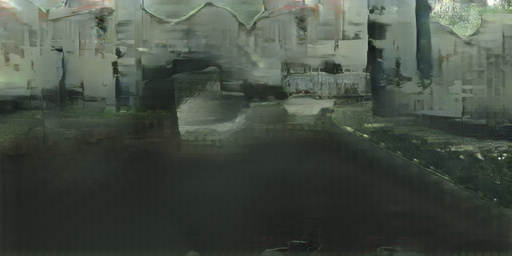}
			\caption{\centering Resynthesized image (Shift)}
	\end{subfigure}		
\end{subfigure}
	\caption{ \textbf{Detecting Houdini  adversarial attacks on Cityscapes.} Without attack, the re-synthesized image {\bf (d)} obtained from {\bf (b)} looks similar to it. By contrast, the resynthesized image {\bf (e)} obtained from the semantic maps {\bf (c)} computed from a Houdini-compromised input differs massively from the original one.}
	\label{fig:advsup1}
\end{figure*}
\providecommand{\localwidth}{}
\renewcommand{\localwidth}{0.19\linewidth}

\begin{figure*}[t!]
	\centering
\begin{subfigure}[t]{\localwidth}
		\begin{subfigure}[t]{\linewidth}
				\centering
			\includegraphics[width=\linewidth]{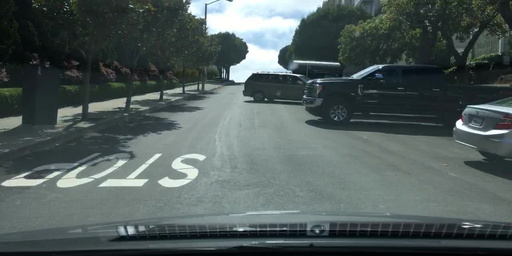}
			\includegraphics[width=\linewidth]{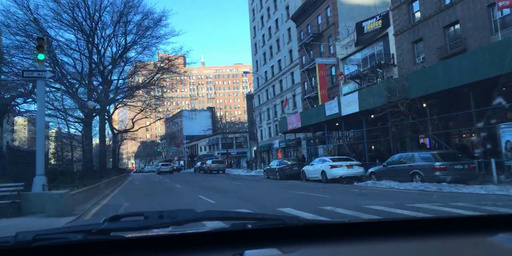}
			\includegraphics[width=\linewidth]{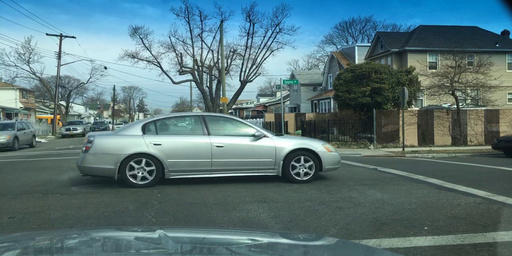}
			\includegraphics[width=\linewidth]{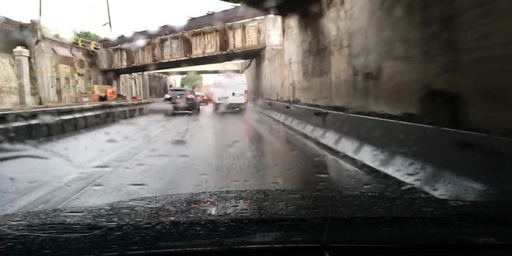}
			\includegraphics[width=\linewidth]{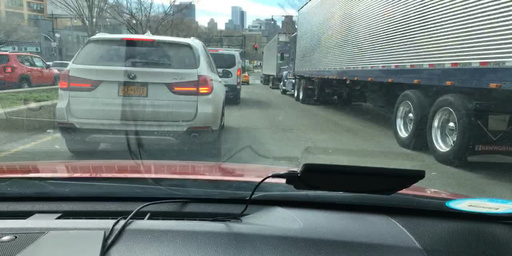}
			\includegraphics[width=\linewidth]{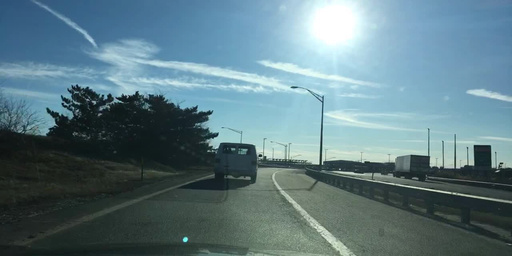}
			\includegraphics[width=\linewidth]{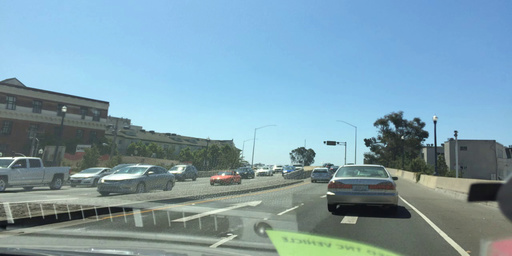}
			\includegraphics[width=\linewidth]{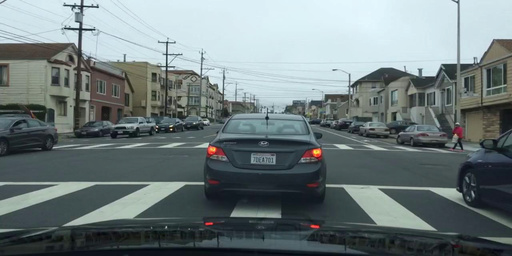}
			\includegraphics[width=\linewidth]{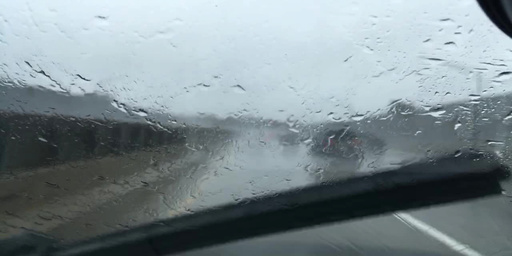}
			\includegraphics[width=\linewidth]{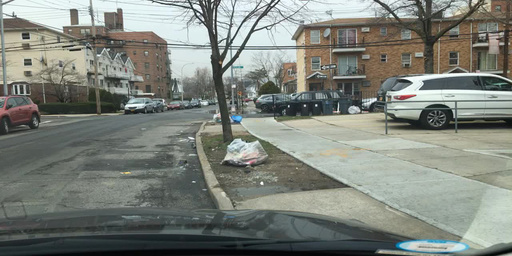}
			\includegraphics[width=\linewidth]{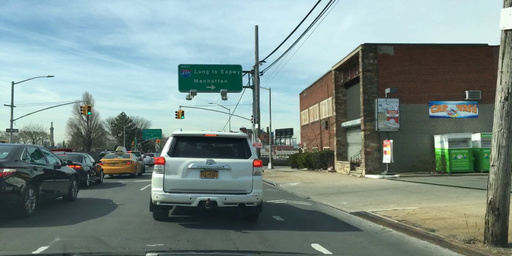}
			\caption{	\centering Input image (normal)}
		\end{subfigure}
	
\end{subfigure}
\begin{subfigure}[t]{\localwidth}
		\begin{subfigure}[t]{\linewidth}
		\includegraphics[width=\linewidth]{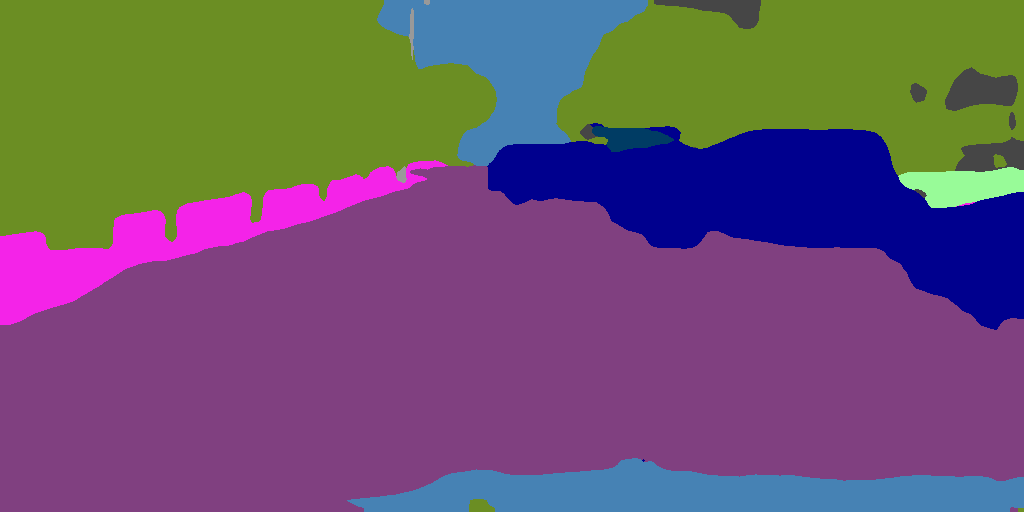}
		\includegraphics[width=\linewidth]{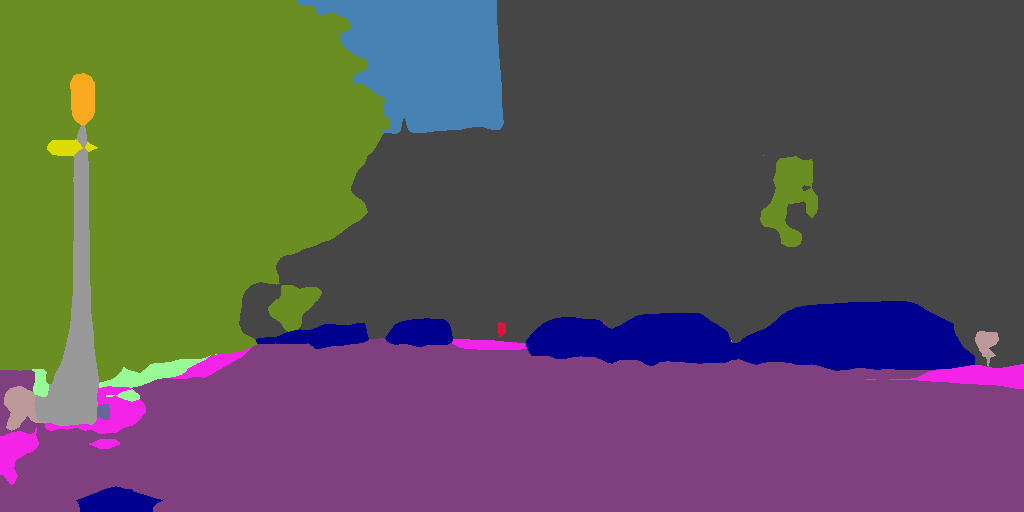}
		\includegraphics[width=\linewidth]{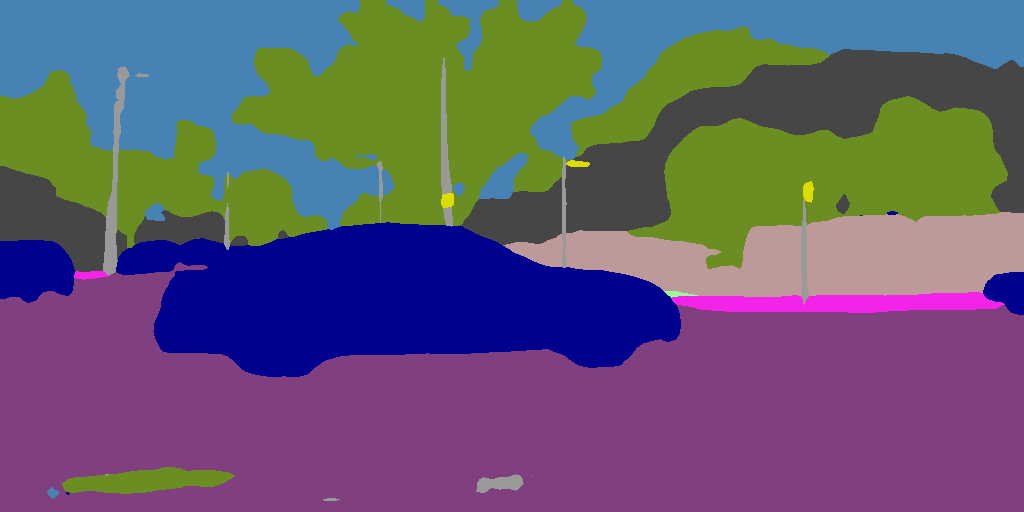}
		\includegraphics[width=\linewidth]{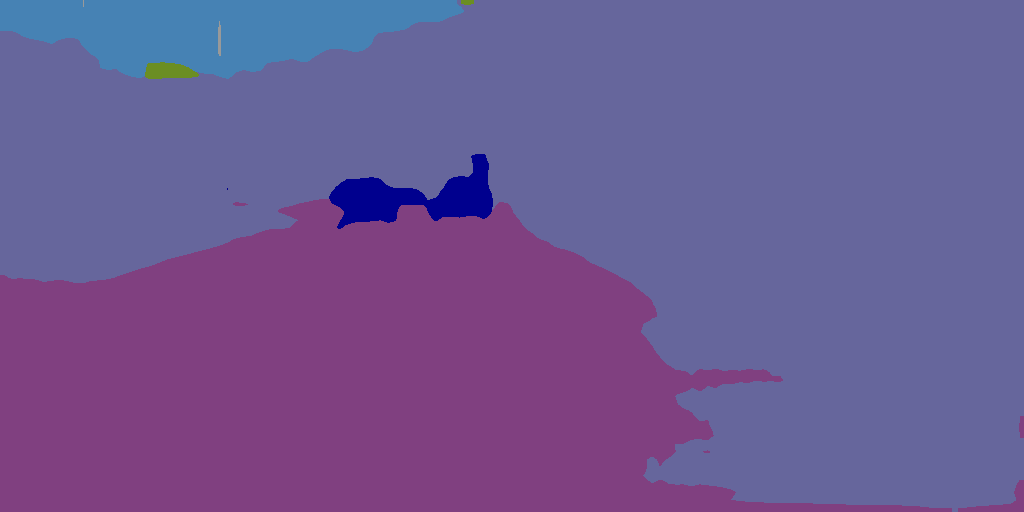}
		\includegraphics[width=\linewidth]{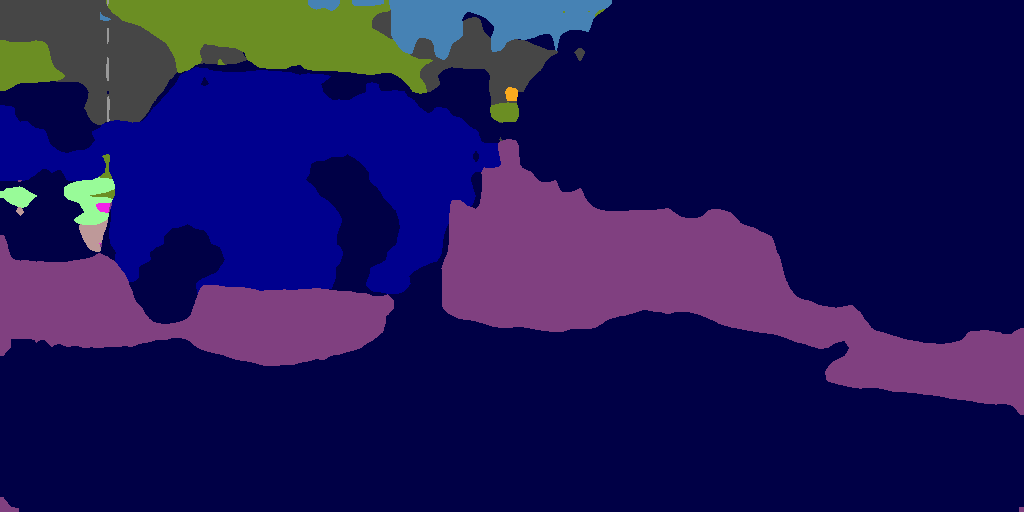}
		\includegraphics[width=\linewidth]{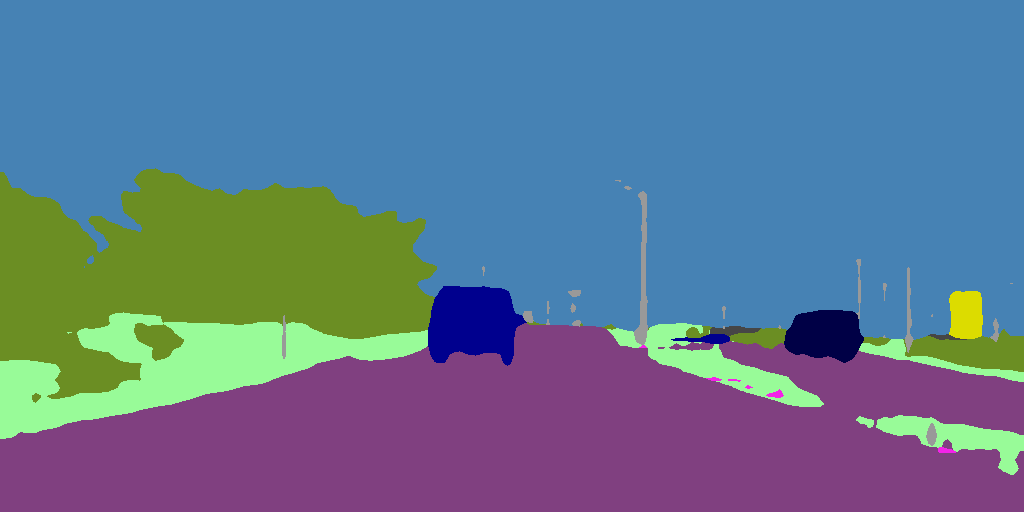}
		\includegraphics[width=\linewidth]{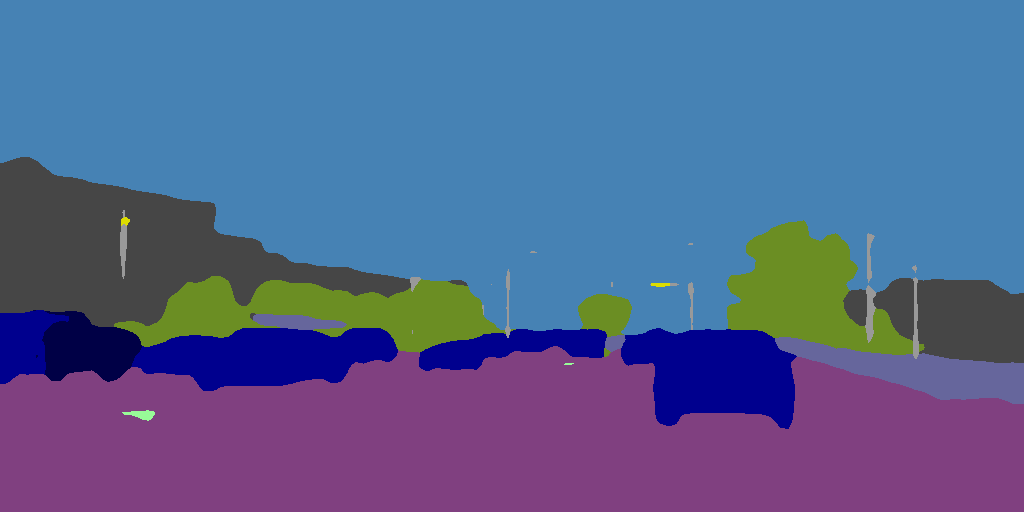}
		\includegraphics[width=\linewidth]{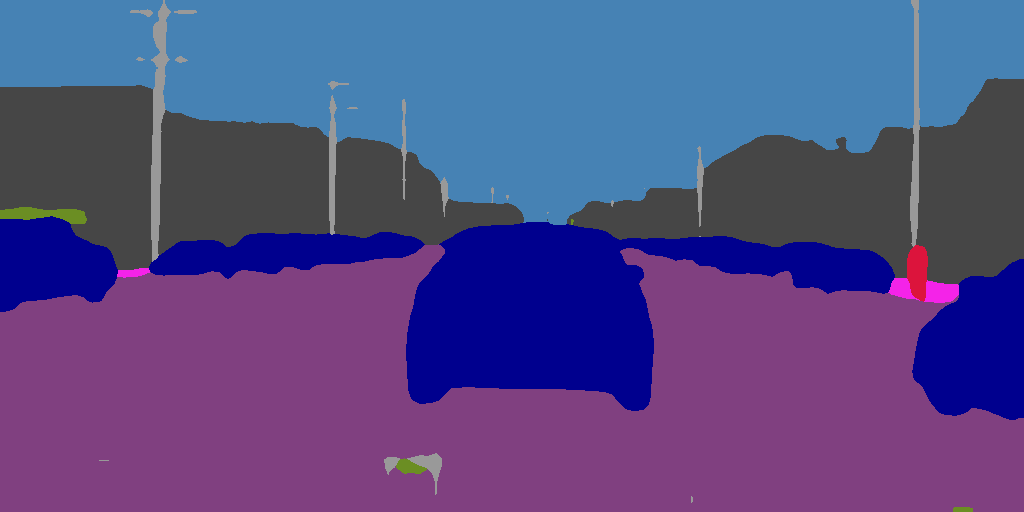}
		\includegraphics[width=\linewidth]{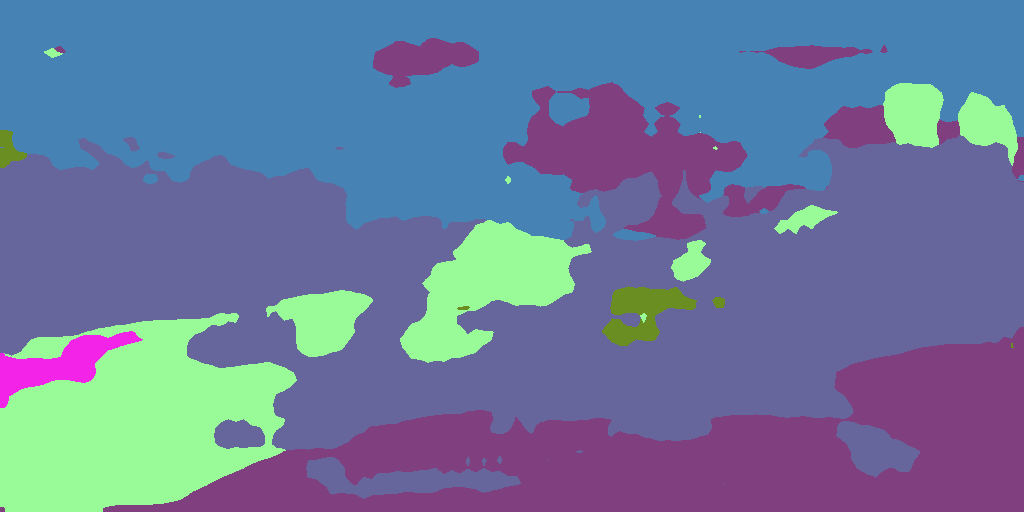}
		\includegraphics[width=\linewidth]{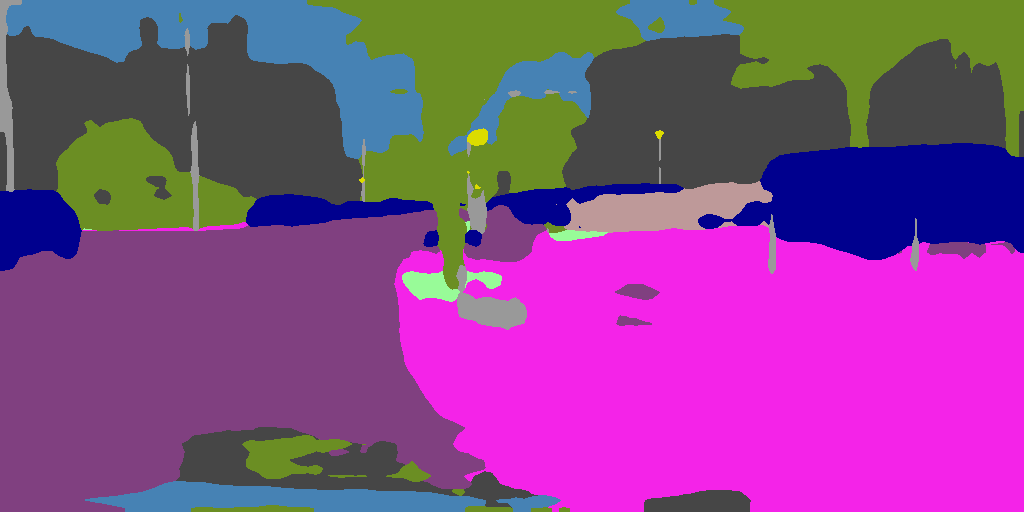}
		\includegraphics[width=\linewidth]{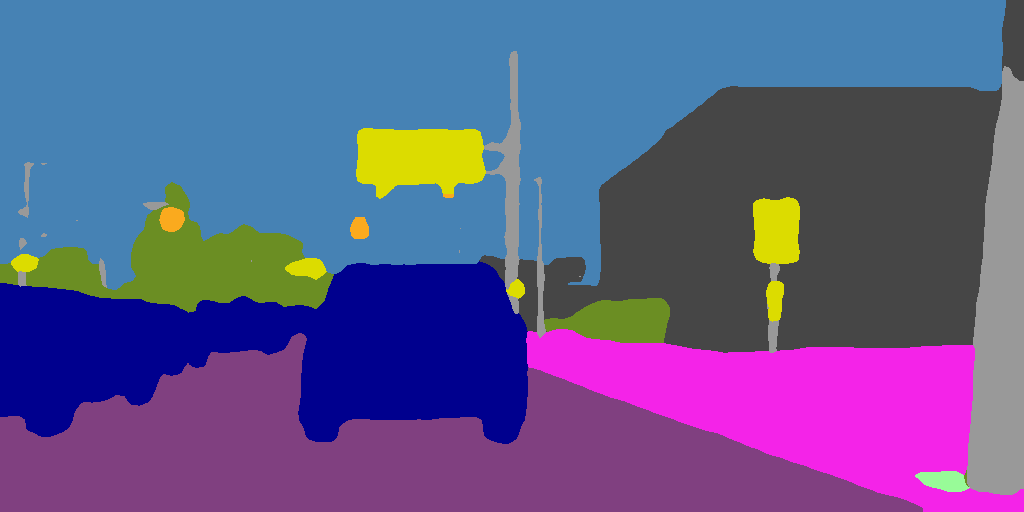}
		\caption{\centering Predicted map (normal) }
	\end{subfigure}
\end{subfigure}
\begin{subfigure}[t]{\localwidth}
	\begin{subfigure}[t]{\linewidth}
	\includegraphics[width=\linewidth]{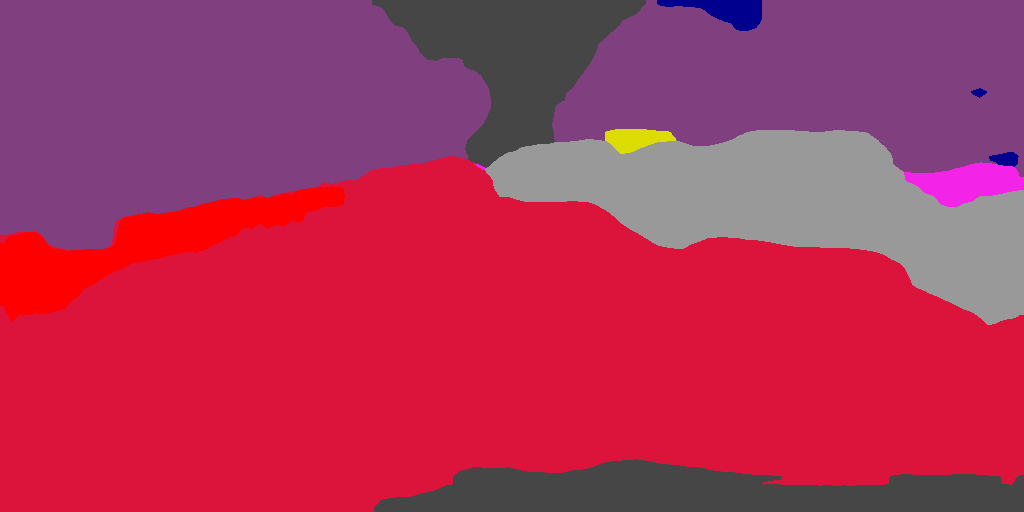}
	\includegraphics[width=\linewidth]{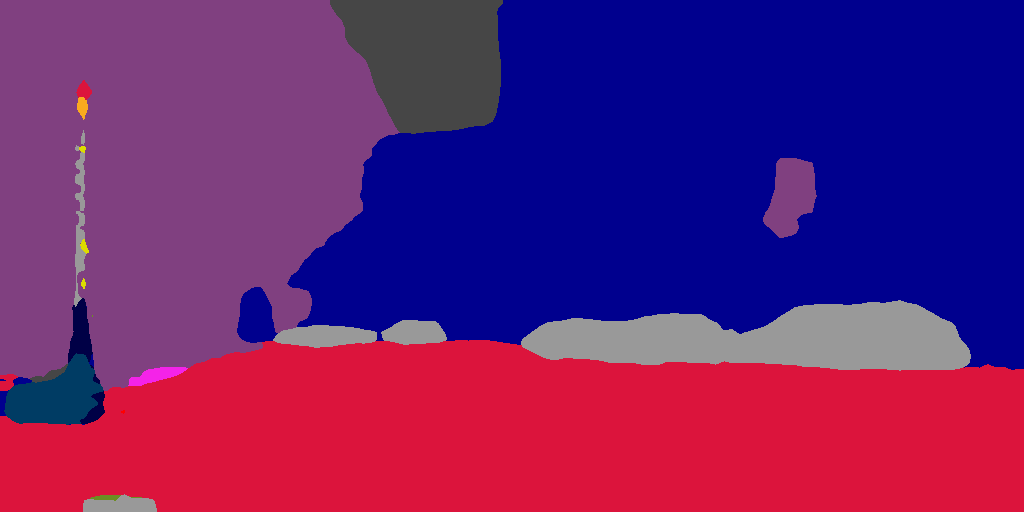}
	\includegraphics[width=\linewidth]{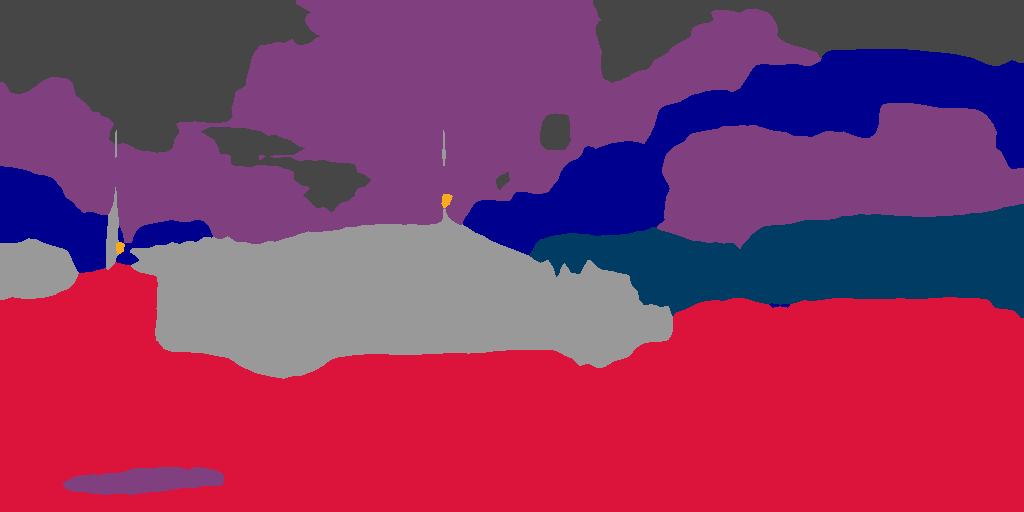}
	\includegraphics[width=\linewidth]{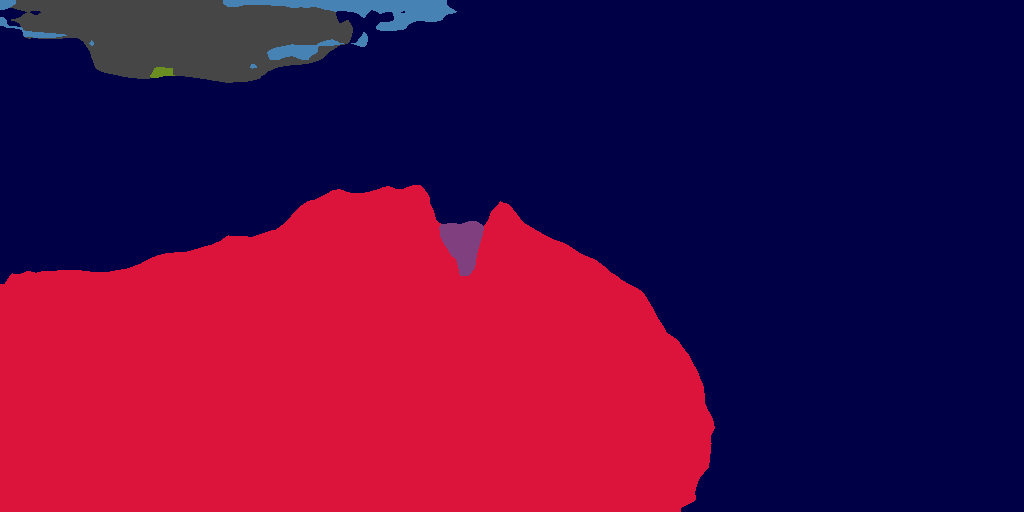}
	\includegraphics[width=\linewidth]{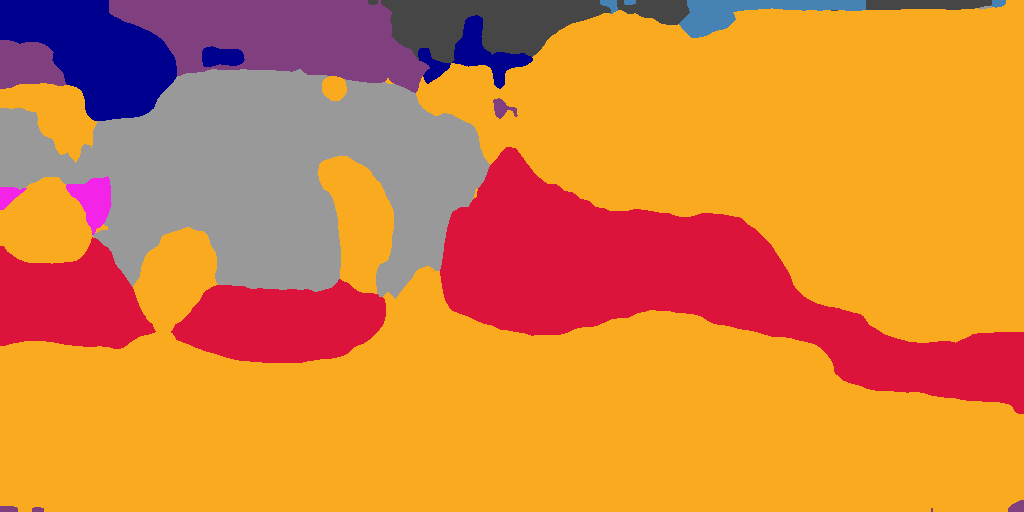}
	\includegraphics[width=\linewidth]{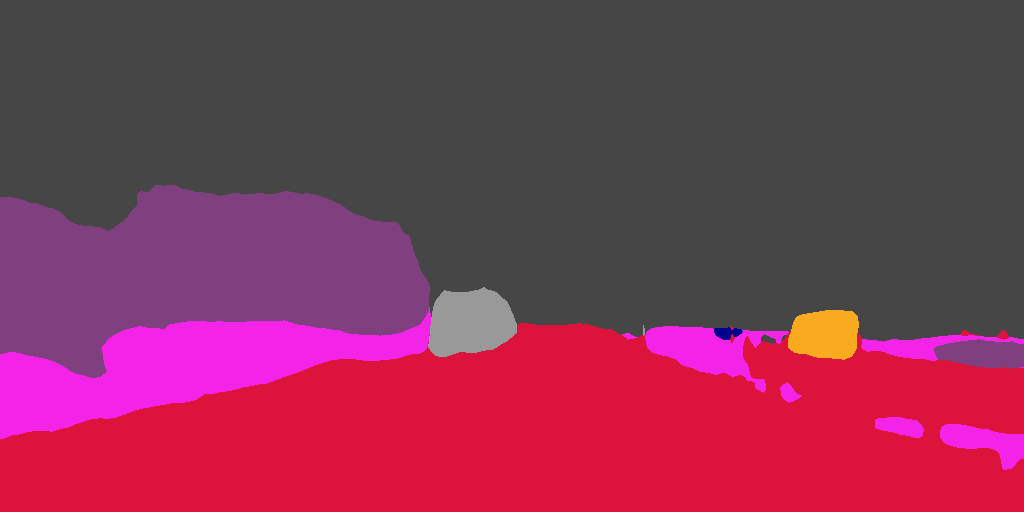}
	\includegraphics[width=\linewidth]{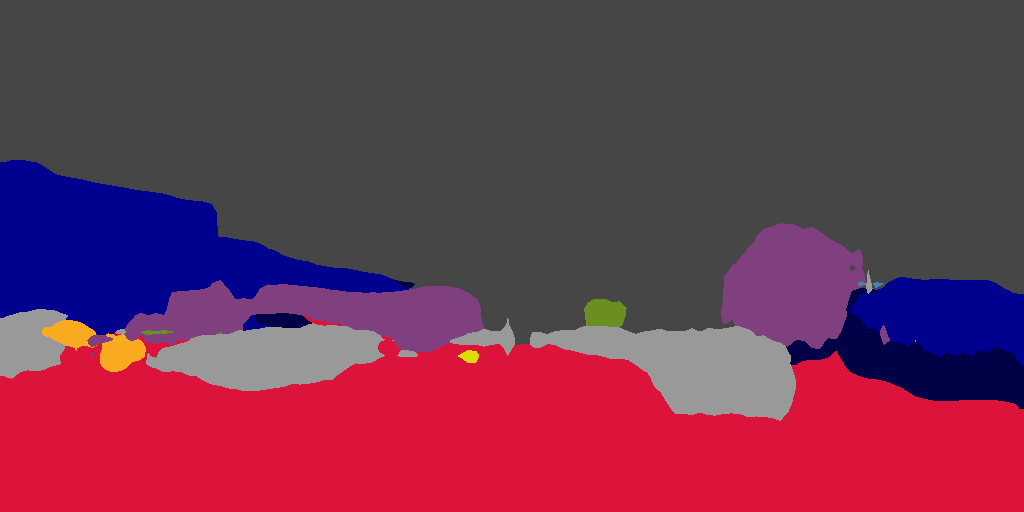}
	\includegraphics[width=\linewidth]{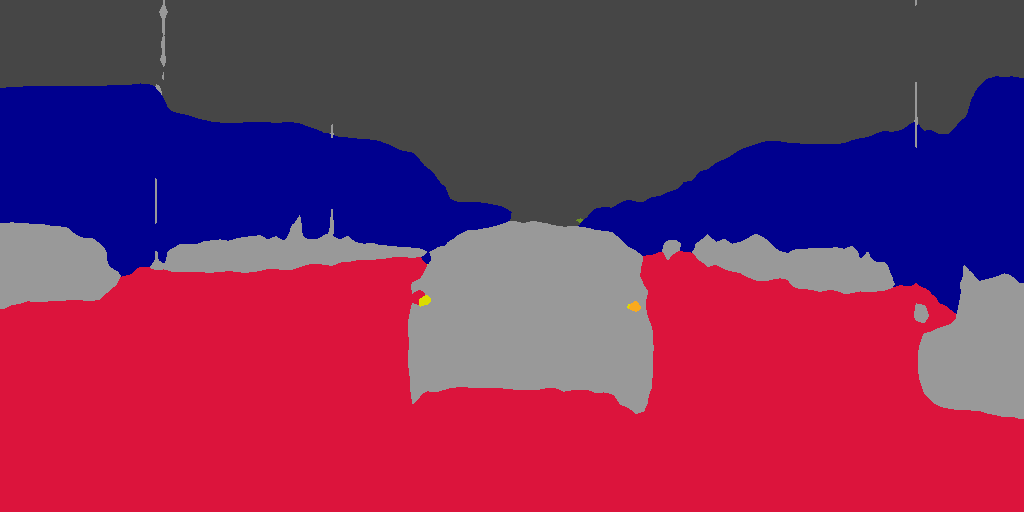}
	\includegraphics[width=\linewidth]{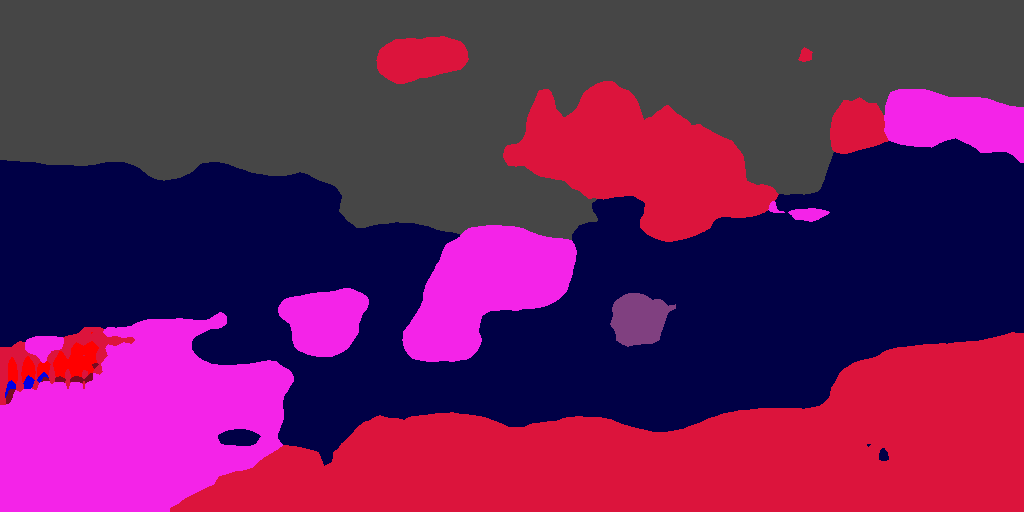}
	\includegraphics[width=\linewidth]{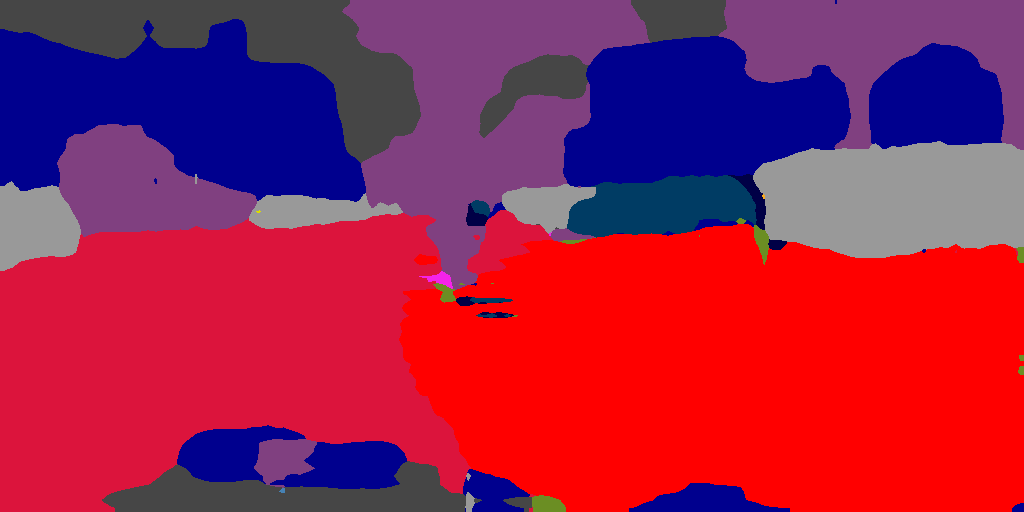}
	\includegraphics[width=\linewidth]{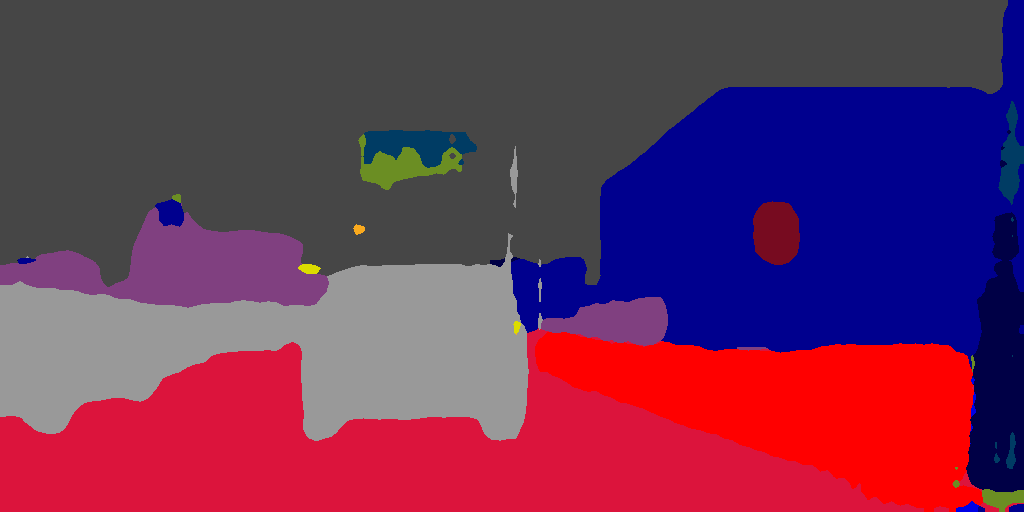}

						\caption{\centering Predicted map (Shift)}
	\end{subfigure}
 \end{subfigure}
\begin{subfigure}[t]{\localwidth}
	\begin{subfigure}[t]{\linewidth}
	\includegraphics[width=\linewidth]{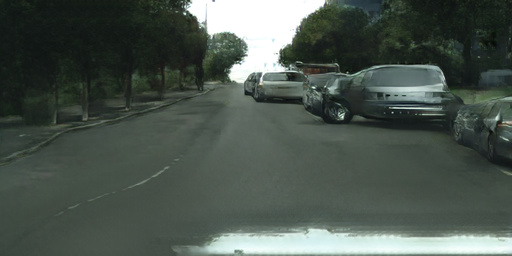}
\includegraphics[width=\linewidth]{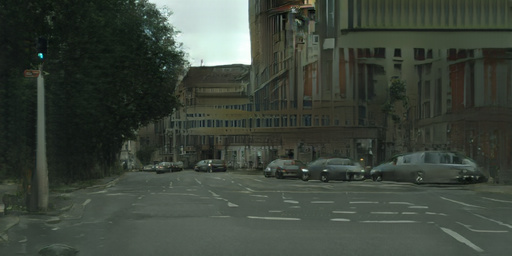}
\includegraphics[width=\linewidth]{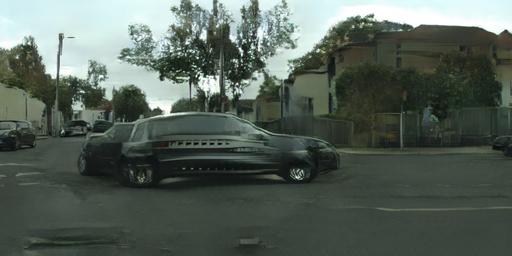}
\includegraphics[width=\linewidth]{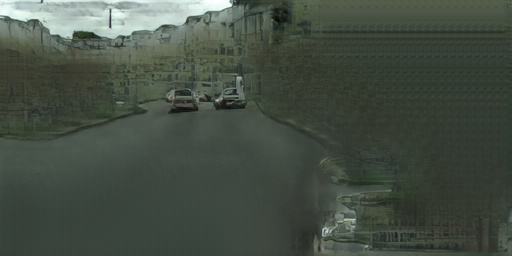}
\includegraphics[width=\linewidth]{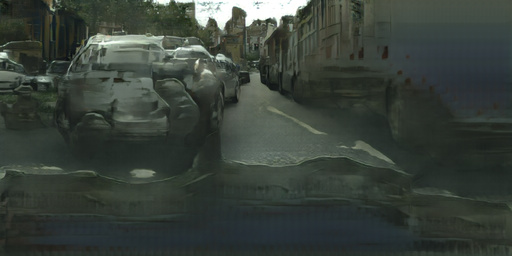}
\includegraphics[width=\linewidth]{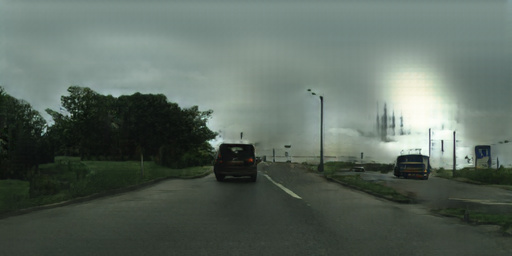}
\includegraphics[width=\linewidth]{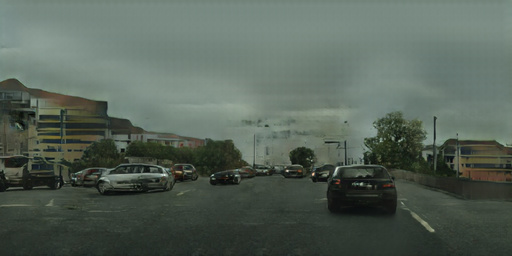}
\includegraphics[width=\linewidth]{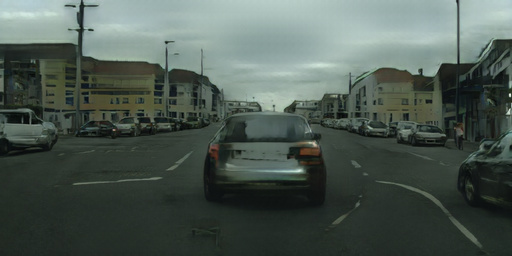}
\includegraphics[width=\linewidth]{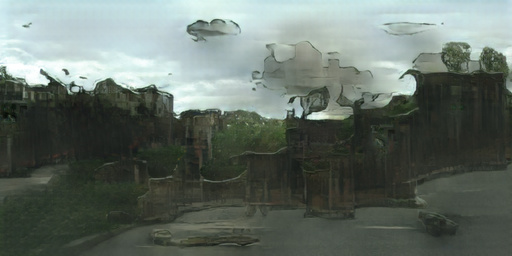}
\includegraphics[width=\linewidth]{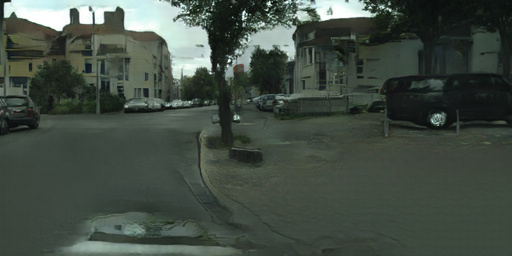}
\includegraphics[width=\linewidth]{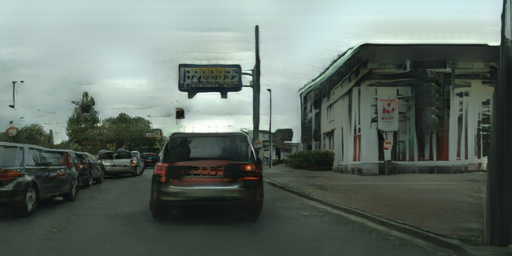}
			\caption{\centering Resynthesized image (normal)}
	\end{subfigure}		
\end{subfigure}
\begin{subfigure}[t]{\localwidth}
	\begin{subfigure}[t]{\linewidth}
	\includegraphics[width=\linewidth]{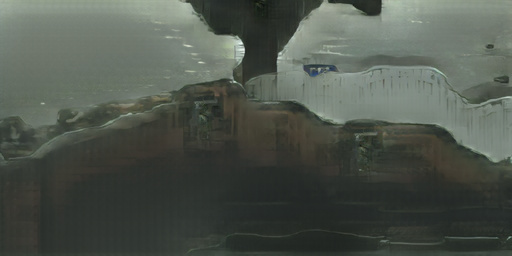}
	\includegraphics[width=\linewidth]{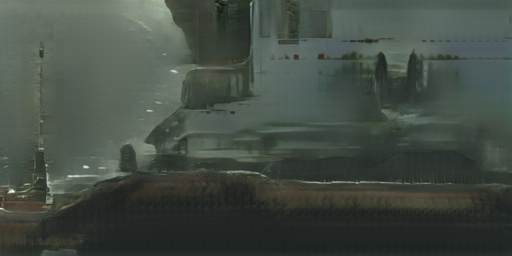}
	\includegraphics[width=\linewidth]{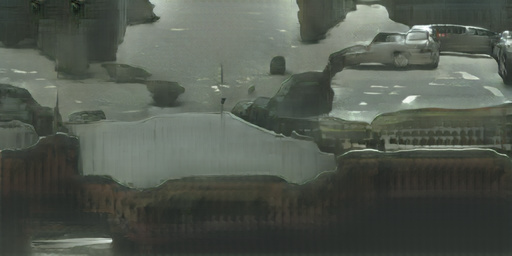}
	\includegraphics[width=\linewidth]{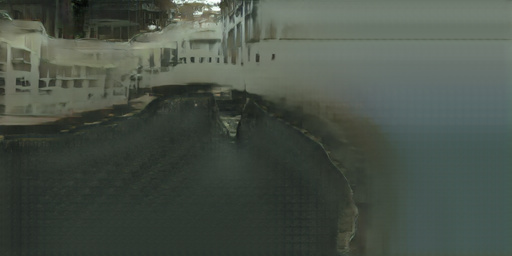}
	\includegraphics[width=\linewidth]{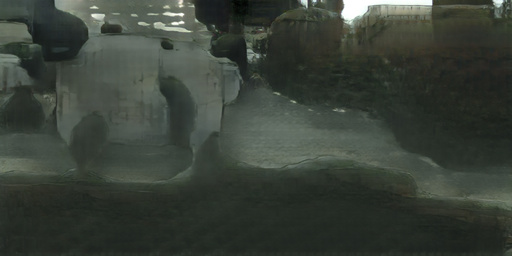}
	\includegraphics[width=\linewidth]{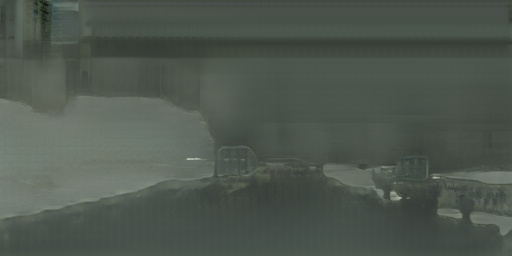}
	\includegraphics[width=\linewidth]{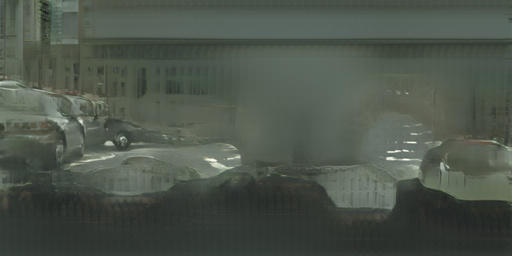}
	\includegraphics[width=\linewidth]{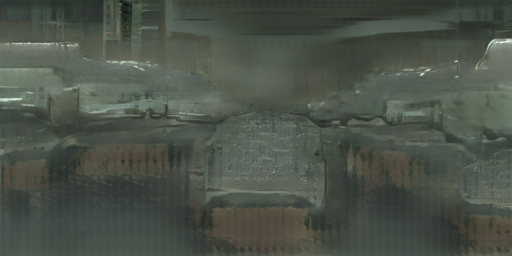}
	\includegraphics[width=\linewidth]{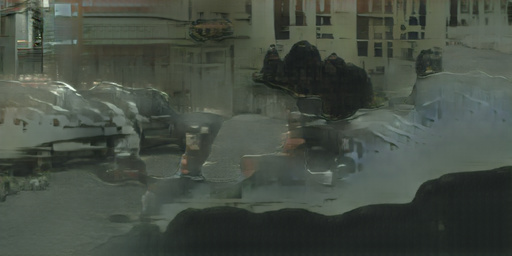}
	\includegraphics[width=\linewidth]{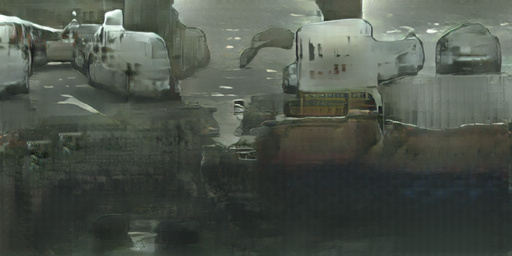}
	\includegraphics[width=\linewidth]{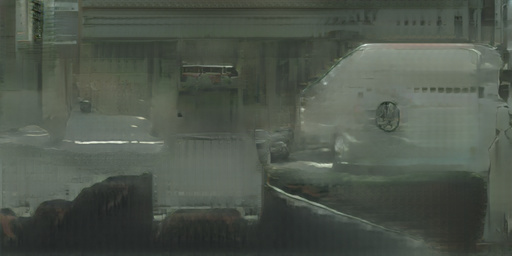}
			\caption{\centering Resynthesized image (Shift)}
	\end{subfigure}		
\end{subfigure}
	\caption{ \textbf{Detecting DAG adversarial attacks on the BDD dataset.} Without attack, the re-synthesized image {\bf (d)} obtained from {\bf (b)} looks similar to it. By contrast, the resynthesized image {\bf (e)} obtained from the semantic maps {\bf (c)} computed from a DAG-compromised input differs massively from the original one.}
	\label{fig:advsup2}
\end{figure*}

\section{ Detecting Adversarial Samples}

We show additional results on adversarial example detection  on the Cityscapes and BDD datasets using the Houdini and DAG attack schemes in Figs.~\ref{fig:advsup1} and~\ref{fig:advsup2}. 
To obtain these results, we set the maximal number of iterations to 200 in all settings and $L_{\infty}$ perturbation of 0.05 across each iteration of the attack.  We randomly choose 80\% of the original validation samples to train the logistic  detectors and the rest of the samples are used for evaluation.  While evaluating the state-of-the-art Scale Consistency method~\cite{Xiao18}, we found by cross-validation that a patch size of $256\times256$ resulted in the best performance for an input image of size $1024\times512$.

\section{Image Attribution}

We used Wikimedia Commons images kindly provided under the Creative Commons Attribution license by the following authors:
Thomas R Machnitzki \href{https://commons.wikimedia.org/wiki/File:Goose_on_the_road_Memphis_TN_2013-03-17_001.jpg}{[link]},
Megan Beckett \href{https://commons.wikimedia.org/wiki/File:Rhino_crossing_road.JPG}{[link]},
Infrogmation \href{https://commons.wikimedia.org/wiki/File:Broadmoor9JanConesSkidloader.jpg}{[link]},
Kyah \href{https://commons.wikimedia.org/wiki/File:Federation_chantier_aout_2006_-_5.JPG}{[link]},
PIXNIO \href{https://commons.wikimedia.org/wiki/File:Bovine_catle_beside_road.jpg}{[link]},
Matt Buck \href{https://commons.wikimedia.org/wiki/File:Beeston_MMB_A6_Middle_Street.jpg}{[link]},
Luca Canepa \href{https://commons.wikimedia.org/wiki/File:Zebra_Crossing_Abbey_Road_Style_(63894353).jpeg}{[link]},
Jonas Buchholz \href{https://commons.wikimedia.org/wiki/File:Aihole-Pattadakal_road.JPG}{[link]}
and
Kelvin JM \href{https://commons.wikimedia.org/wiki/File:A_man_carrying_dry_grass_on_bicycle_for_domestic_animal_like_cows.jpg}{[link]}.

\end{document}